\documentclass[sigconf,natbib=true]{acmart}
\usepackage{subfig}
\AtBeginDocument{%
  }

\setcopyright{acmlicensed}
\copyrightyear{2025}
\acmYear{2025}
\acmDOI{XXXXXXX.XXXXXXX}
\acmISBN{978-1-4503-XXXX-X/2018/06}

\usepackage{graphicx}
\usepackage{multirow}
\usepackage{amsmath}
\usepackage{appendix}
\usepackage{float}
\usepackage{pifont}
\newcommand{\cmark}{\ding{51}}%
\newcommand{\xmark}{\ding{55}}%
\begin{document}

\title{OmniGeo: Towards a Multimodal Large Language Models for Geospatial Artificial Intelligence}


\author{Long Yuan}
\authornote{Equal contribution.}
\affiliation{%
  \institution{Beijing Jiaotong University}
  \city{Beijing}
  \country{China}
}
\email{24120408@bjtu.edu.cn}

\author{Fengran Mo}
\authornotemark[1]
\affiliation{%
  \institution{University of Montreal}
  \city{Montreal}
  \state{Quebec}
  \country{Canada}
}
\email{fengran.mo@umontreal.ca}

\author{Kaiyu Huang}
\authornote{Corresponding author}
\affiliation{%
  \institution{Beijing Jiaotong University}
  \city{Beijing}
  \country{China}
}
\email{kyhuang@bjtu.edu.cn}

\author{Wenjie Wang}
\affiliation{%
  \institution{Beijing Jiaotong University}
  \city{Beijing}
  \country{China}
}

\author{Wangyuxuan Zhai}
\affiliation{%
  \institution{Beijing Jiaotong University}
  \city{Beijing}
  \country{China}
}

\author{Xiaoyu Zhu}
\affiliation{%
  \institution{Beijing Jiaotong University}
  \city{Beijing}
  \country{China}
}

\author{You Li}
\affiliation{%
  \institution{Beijing Jiaotong University}
  \city{Beijing}
  \country{China}
}

\author{Jinan Xu}
\affiliation{%
  \institution{Beijing Jiaotong University}
  \city{Beijing}
  \country{China}
}

\author{Jian-Yun Nie}
\affiliation{%
  \institution{University of Montreal}
  \city{Montreal}
  \state{Quebec}
  \country{Canada}
}

\renewcommand{\shortauthors}{Yuan et al.}
\newcommand{\hky}[1]{\textcolor{magenta}{[#1 -- \textsc{hky}]}}
\begin{abstract}
The rapid advancement of multimodal large language models~(LLMs) has opened new frontiers in artificial intelligence, enabling the integration of diverse large-scale data types such as text, images, and spatial information.
In this paper, we explore the potential of multimodal LLMs~(MLLM) for geospatial artificial intelligence~(GeoAI), a field that leverages spatial data to address challenges in domains including Geospatial Semantics, Health Geography, Urban Geography, Urban Perception, and Remote Sensing.
We propose a MLLM~(OmniGeo) tailored to geospatial applications, capable of processing and analyzing heterogeneous data sources, including satellite imagery, geospatial metadata, and textual descriptions. 
By combining the strengths of natural language understanding and spatial reasoning, our model enhances the ability of instruction following and the accuracy of GeoAI systems.
Results demonstrate that our model outperforms task-specific models and existing LLMs on diverse geospatial tasks, effectively addressing the multimodality nature while achieving competitive results on the zero-shot geospatial tasks. Our code will be released after publication.
\end{abstract}





\maketitle

\section{Introduction}

\begin{figure}[!t]
\centering
\includegraphics[width=1\linewidth]{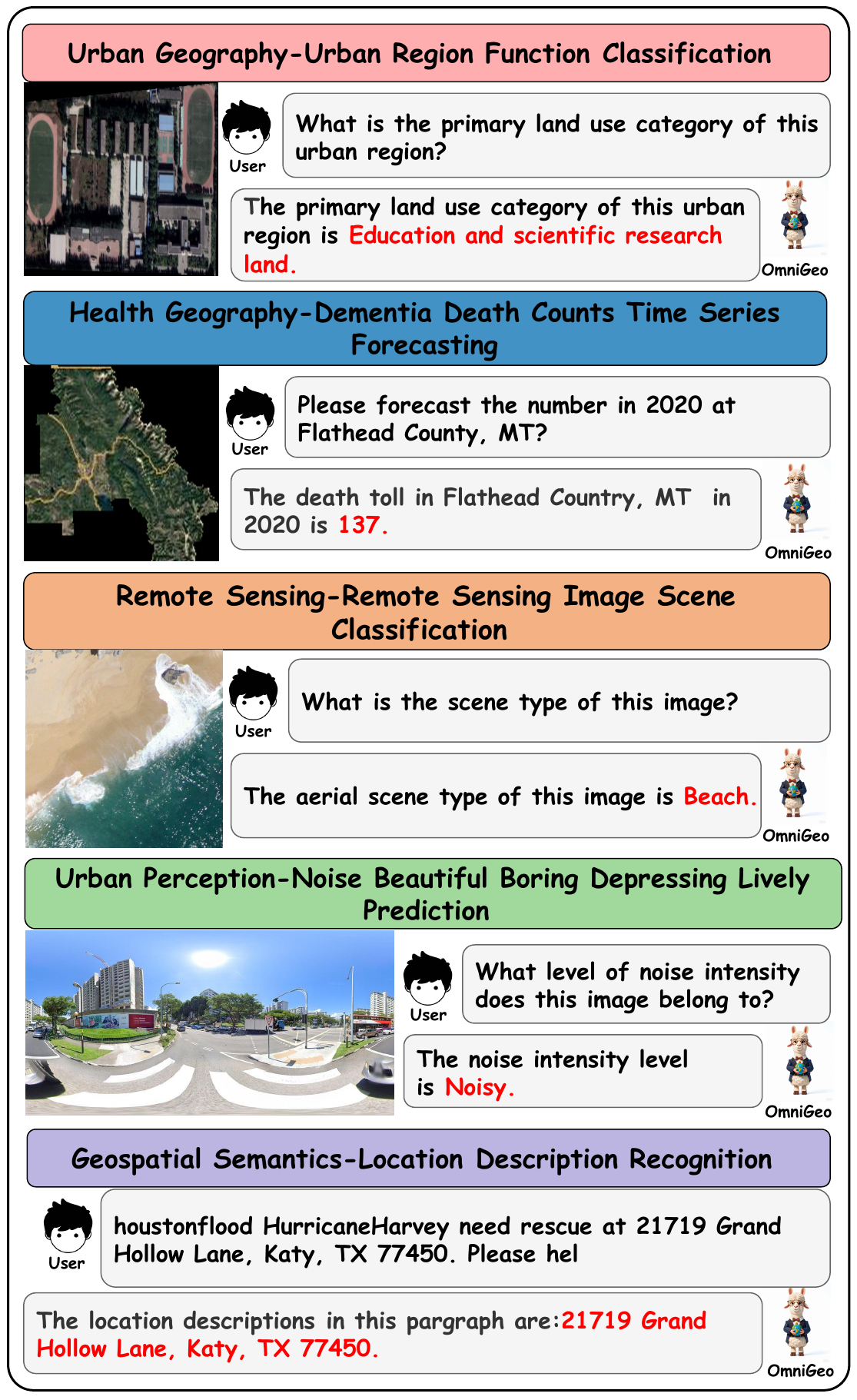}
\vspace{-4ex}
\caption{The Illustration of all the tasks covered by \textbf{OmniGeo} through an engaging dialogue.}
\label{fig:OmniGeo}
\vspace{-4ex}
\end{figure}


Recent advancements in machine learning~(ML) and artificial intelligence~(AI) underscore the immense potential of data and computational power~\cite{Hollmann2025, Pannone2024, Yoo2025}. 
Exceptionally large language models~(LLMs), trained on large-scale datasets, have demonstrated remarkable effectiveness in a wide range of learning tasks~\cite{deepseekai2024deepseekv3technicalreport, openai2024openaio1card, zhu2023minigpt}.
In particular, the unprecedented success of LLMs has catalyzed a paradigm shift in the training of ML and AI models. 
Instead of developing task-specific models from scratch for each individual task, the generality and adaptability of LLMs are leveraged to perform multiple tasks simultaneously with a single model through few-shot/zero-shot learning techniques~\cite{10.1145/3653070}. This enables deployment across a diverse range of domains~\cite{mo2024survey,huang2024survey}, such as healthcare~\cite{NEURIPS2023_5abcdf8e, lee2024llmcxrinstructionfinetunedllmcxr}, education~\cite{gan2023largelanguagemodelseducation}, law~\cite{jiang-etal-2024-leveraging, mahari-etal-2023-law}, and finance~\cite{Xia_Sun_Wang_An_2024, chen-etal-2022-convfinqa}.
LLMs have encapsulated the knowledge embedded in their training corpus, which encompasses billions or even trillions of tokens sourced from the Internet~\cite{10.5555/3692070.3693479}. 
Therefore, we aim to investigate approaches to extract the geospatial knowledge that LLMs possess, enhancing a variety of geospatial machine learning tasks.

However, as shown in Figure~\ref{fig:OmniGeo}, geospatial tasks typically include sequence labeling, time series prediction forecasting, geospatial image classification, and tasks related to urban functions~\cite{10.1145/3653070}. 
Thus, geographic science is an inherently complex discipline that encompasses a wide range of tasks, indicating that addressing geospatial tasks requires the integration of multiple modalities.
The primary technical challenge in Geospatial Artificial Intelligence~(GeoAI) lies in its inherently multimodal nature. 
The data modalities in GeoAI include text, images (e.g., RS images or street view images), trajectory data, knowledge graphs, and geospatial vector data, which encapsulate critical geospatial information~\cite{10.1007/978-3-030-14745-7_2, article,CHOI2021103091, rao2023catsconditionaladversarialtrajectory}. 
Each modality exhibits unique structural characteristics that demand distinct representations. 
As a result, the multimodal nature of GeoAI complicates the straightforward application of existing LLMs across the full spectrum of GeoAI tasks.
Few existing studies integrate these diverse representations while incorporating suitable inductive biases into a single model requires careful and thoughtful design. 

In this paper, we propose a unified MLLM that leverages diverse geographic spatial data to guide the implicit geospatial knowledge learned by the foundation model during unsupervised pre-training, allowing it to better understand geoscience tasks and provide accurate responses.
The data utilized include geographic textual data from tweets and webpages, spatial polygon vector data from CDC Wonder (dementia data at the state and county level), spatial point and polygon vector data from Gaode Maps (points of interest, POI), RS images from WorldView, street view images from Google Maps, and spatial polygon vector data from city government websites (urban planning layer data). 
Initially, because LLMs are unable to directly comprehend the spatial distribution of large-scale POI data of the region, we convert the geographic spatial vector data (POI and dementia data) into text paragraphs, crop RS images by region, and generate captions for both RS and street view images. 
Subsequently, the aforementioned geographic spatial data are aligned with geographic entities to create a multimodal instruction fine-tuning dataset. 
Finally, we obtain OmniGeo using these instruction data in LoRA and full-supervision fine-tuning manner. 
This model achieves commendable performance across multiple geospatial tasks, proving that the model have successfully extracted the implicit geospatial knowledge from the base model, thereby enabling more accurate geospatial task inference.

Our contributions are summarized as (1) We propose OmniGeo, a MLLM to facilitate the integration of geospatial information from different modalities, mitigating interference between tasks, and enabling a single large model to simultaneously handle multiple heterogeneous geospatial tasks. (2) OmniGeo has constructed 12 multimodal instruction fine-tuning data for GeoAI and achieves competitive performance on geospatial tasks. (3) To the best of our knowledge, OmniGeo is the first large-scale multimodal language model that covers the five core tasks in the GeoAI field (health geography, urban geography, RS images, urban perception, and semantic analysis), taking the step towards the multimodal development of GeoAI.

\section{Related Work}

\textbf{Geospatial Artificial Intelligence.} For toponym recognition and location description recognition, it is regarded as a subtask of Named Entity Recognition (NER), with the goal of identifying named places or location descriptions from text. early methods utilized general NER tools such as Stanford NER~\cite{finkel2005incorporating} and spaCy NER~\cite{honnibal2017spacy} to uniformly identify geospatial semantics. Wang et al.~\cite{wang2020neurotpr} were the first to design the NeuroTPR, an RNN-based model which can extract location information from text.

For urban region function classification, it aims to classify the primary functional categories of urban region based on various geospatial data within the target region, which is beneficial for city planning and resource allocation. early methods~\cite{yuan2016analyzing, askarizad2022perception} primarily relied on researchers' subjective analysis of urban planning maps and human mobility data to determine urban region function categories. With the acquisition and application of multi-source heterogeneous geographic data, Jing et al.~\cite{jing2008remote} developed an RS image semantic interpretation model to analyze the characteristics of specific urban functional regions, whereas Qi et al.~\cite{qi2011measuring} performed a qualitative analysis of local urban region functions using GPS-based taxi mobility data. Existing research, including Place2Vec~\cite{yao2017sensing} and HGI~\cite{huang2023learning}, has developed region-specific semantic embeddings, 
to effectively carry out various downstream tasks.

For RS image scene classification, Yao et al.~\cite{yao2016semantic} leverage stacked sparse autoencoders to classify scenes via learning a large number of discriminative image features. 
To transfer the pre-trained CNN to RS image classification tasks, He et al.~\cite{he2018remote} designed the MSCP algorithm to automatically select and combine multi-level feature maps extracted from the pre-trained CNN.

\noindent \textbf{Multimodal Large Language Models for Domain Tasks.} MLLM research in GeoAI is primarily focused on handling RS images. To understand objects with arbitrary sizes and orientations, Wang et al.~\cite{wang2022advancing} designed the Remote-Sensing-RVSA with a new rotation-invariant scalable attention mechanism, achieving SOTA performance in several RS image visual tasks. Pang et al.~\cite{pang2024vhmversatilehonestvision} developed a large-scale RS image-text dataset, HnstD, to perform a variety of common RS image tasks and show clear improvements over previous VLMs. Kartik Kuckreja et al.~\cite{kuckreja2024geochat} constructed a new RS multimodal instruction-following dataset to train models capable of performing region-level reasoning, and named the model GeoChat.

Different from the above work, we hope to achieve a new paradigm in which a single model can cover all core tasks of GeoAI, in an attempt to alleviate the inherent challenges brought by multimodal complex data.

\begin{figure*}[!t]
    \centering
    \includegraphics[width=1.0\textwidth]{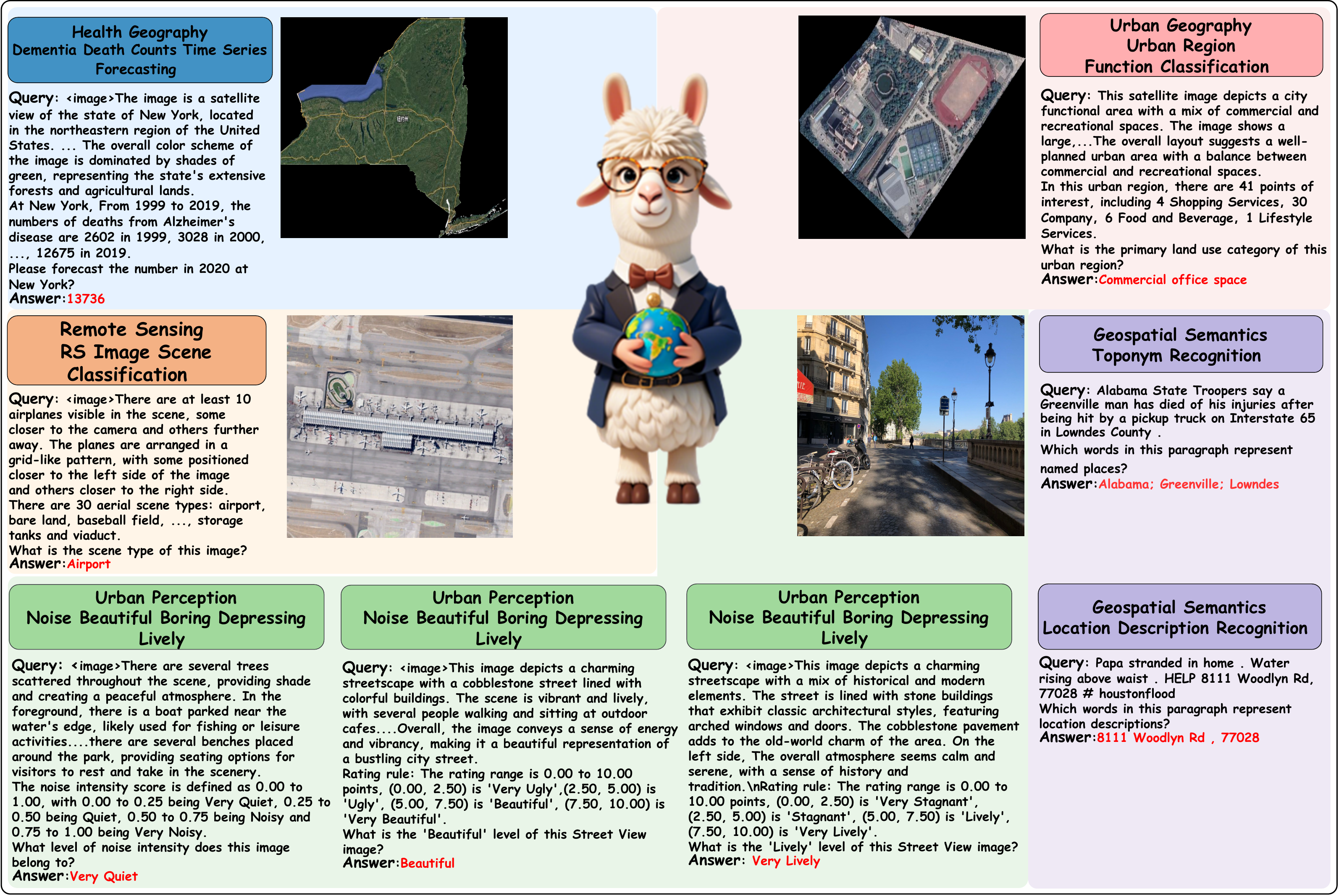}
    \caption{A detailed illustration of the image-text instruction data and geospatial tasks covered by OmniGeo.}
    \label{fig:instructions}

\end{figure*}
\section{Preliminaries}

\subsection{Task Definition}

\noindent\textbf{Toponym Recognition and Location Description Recognition}: Toponym Recognition is a subtask of NER. The challenge of this task lies in correctly distinguishing geographically similar entities that appear as names of coarse-grained locations (such as cities, states, and countries). Formally, given a text $T = \{t_1, t_2, ..., t_n\}$, the goal is to assign each token $t_i$ a label $y_i \in \{\text{B-LOC}, \text{I-LOC}, \text{O}\}$, where
\begin{equation}
y_i \in \left\{
\begin{array}{ll}
\text{B-LOC} & \text{if } t_i \text{ is the beginning of a location description,} \\
\text{I-LOC} & \text{if } t_i \text{ is inside a location description,} \\
\text{O} & \text{if } t_i \text{ is outside a location description.}
\end{array}
\right.
\end{equation}

Moreover, the goal of Location Description Recognition is to identify the location description that appears within a given text. The challenge of this task lies in whether it can fully output more fine-grained location descriptions, such as home addresses, highway exits, and road intersections, rather than large-scale geographic entities like cities, states, and countries.
Formally, given a text $T = \{t_1, t_2, ..., t_n\}$, the task is to assign each token $t_i$ a label $y_i \in \{\text{B-LDR}, \text{I-LDR}, \text{O}\}$, where
\begin{equation}
y_i \in \left\{
\begin{array}{ll}
\text{B-LDR} & \text{if } t_i \text{ is the beginning of a location description,} \\
\text{I-LDR} & \text{if } t_i \text{ is inside a location description,} \\
\text{O} & \text{if } t_i \text{ is outside a location description.}
\end{array}
\right.
\end{equation}
\raggedbottom
\noindent\textbf{Dementia Death Counts Time Series Forecasting}: 
Given historical dementia mortality time series data \( \{Y_t^r,Y_{t-1}^r, ..., Y_1\} \) for a specific region \( r \) (e.g., state, county) and the corresponding RS image \( RS_r \) for the region at a fixed time \( t \), the goal is to predict the mortality at the next time point, \( Y_{t+1}^r \). 
The challenge of this task lies in whether it can eliminate "geographic bias"~\cite{faisal2022geographic} and reasonably estimate the death count, considering the geographical distribution differences~\cite{akushevich2021geographic} in death rate growth. 
The task can be formally defined as
\begin{equation}
Y_{t+1}^r = f\left(\{Y_t^r,Y_{t-1}^r, ..., Y_1\}, RS_r\right)
\end{equation}
\raggedbottom
where \( Y_t^r \) represents the number of deaths in region \( r \) at time \( t \),\( RS_r \) is the RS image of the region \( r \).

\noindent\textbf{Urban Region Function Classification}: The goal of this task is to determine the human activity patterns within the region based on the distribution of POI data and regional RS images, and correctly estimate the primary urban function category. Formally, this can be expressed as:
\begin{equation}
\hat{F}_r = f(\mathbf{P}_r, RS_r),
\end{equation}
\raggedbottom
where $\hat{F}_r$ is the predicted primary urban function of region $r$ and $\mathbf{P}_r$ is a set representing the counts of different types of POIs in region $r$, i.e., $\mathbf{P}_r = (p_{r1}, p_{r2}, \dots, p_{rn})$. The $p_{ri}$ and $RS_r$ denote the number of POIs of type $i$ and the RS image of the region $r$.

\noindent\textbf{Remote Sensing Image Scene Classification}: The goal of this task is to correctly distinguish highly similar scene types based on information such as the types, quantities, and layouts of geographic entities in the RS images. Formally, it can be expressed as
$\hat{C}_i = f(RS_i),
$
The $\hat{C}_i$ and $RS_i$ are the predicted scene and the original type of the RS image $i$.

\noindent\textbf{Urban Perception Prediction}: Based on street view images, this task involves judging the fine-grained perceptual features of urban neighborhoods from seven perceptual indicators (Noise, Beautiful, Boring, Depressing, Lively, Safe, Wealthy), with each indicator subdivided into four perceptual features. The challenge of this task lies in whether the model possesses advanced human perception knowledge~\cite{zhang2021perception, zhang2018measuring}. Formally, this can be expressed as:
\begin{equation}
\hat{P}_i = f(SVI_i),
\end{equation}
\raggedbottom
where $\hat{P}_i$ is the predicted perception feature vector of the urban region in image $i$. Specifically, $\hat{P}_i$ consists of seven perception indicators (\textbf{Noise}, \textbf{Beautiful}, \textbf{Boring}, \textbf{Depressing}, \textbf{Lively}, \textbf{Safe}, \textbf{Wealthy}), and the prediction for each indicator is the maximum value among its four different levels of perceptual features:
\begin{equation}
    \hat{P}_i = \max(P_i^1, P_i^2, P_i^3, P_i^4), \quad \text{for each } i \in \{1, 2, \dots, 7\},
\end{equation}
\raggedbottom
where $P_i^k$ ($k = 1, 2, 3, 4$) represents the $k$-th level feature of the $i$-th perception indicator.$SVI_i$ is the street view image $i$.

\subsection{Exploration of LLMs for GeoAI}

As a starting point for this study, we empirically demonstrate the potential of leveraging LLMs for addressing geospatial tasks. 
We aim to show that our investigation not only highlight the effectiveness of general-purpose, few-shot learners in the geospatial domain, but also challenge the prevailing paradigm of training task-specific models as a standard practice in GeoAI research. 
To this end, we compare the performance of specific ML models and LLMs. 
These models are evaluated against multiple supervised, task-specific baselines on two representative geospatial tasks: (1) toponym recognition and (2) RS image scene classification.

\begin{table}[h]
  \begin{centering}
  \small
  \caption{Accuracy of the baseline models on two representative geospatial tasks.}
  \label{table:validResult}
  \vspace{-3ex}
  \begin{tabular}{lcc}
  \toprule
  \multirow{2}{*}{Model} & \multirow{2}{*}{Toponym Recognition} & Remote Sensing\\
     &  &  Image Classification \\
   \midrule
   Stanford NER & 0.757 & - \\
   AlexNet & - & 0.812 \\
   LLaVA & 0.708 & 0.648 \\
   GPT-4 & 0.731 & 0.794 \\
  \bottomrule
  \end{tabular}
    \vspace{-3ex}   
  \end{centering}
  \end{table}
As shown in Table~\ref{table:validResult}, GPT-4 achieves promising results, demonstrating that a significant amount of geospatial knowledge embedded in LLMs.
This suggests that using instruction data to guide foundation models in constructing specialized geospatial LLMs holds significant potential.
However, extracting this knowledge from LLMs is a non-trivial task.
LLMs are unable to perform spatial reasoning in a way that is grounded in the real world.
Therefore, it is critical to align geospatial information from different modalities to uncover accurate geospatial knowledge during inference.


\section{Methodology}
In this section, we first describe the sources of geospatial data and the process of constructing multimodal instruction fine-tuning datasets.
We then propose the use of multi-task joint training to activate the inherent geospatial knowledge within MLLMs, thereby enhancing their instruction-following capabilities and improving accuracy in geospatial tasks

\begin{figure*}[!t]
    \centering
    \includegraphics[width=0.8\textwidth]{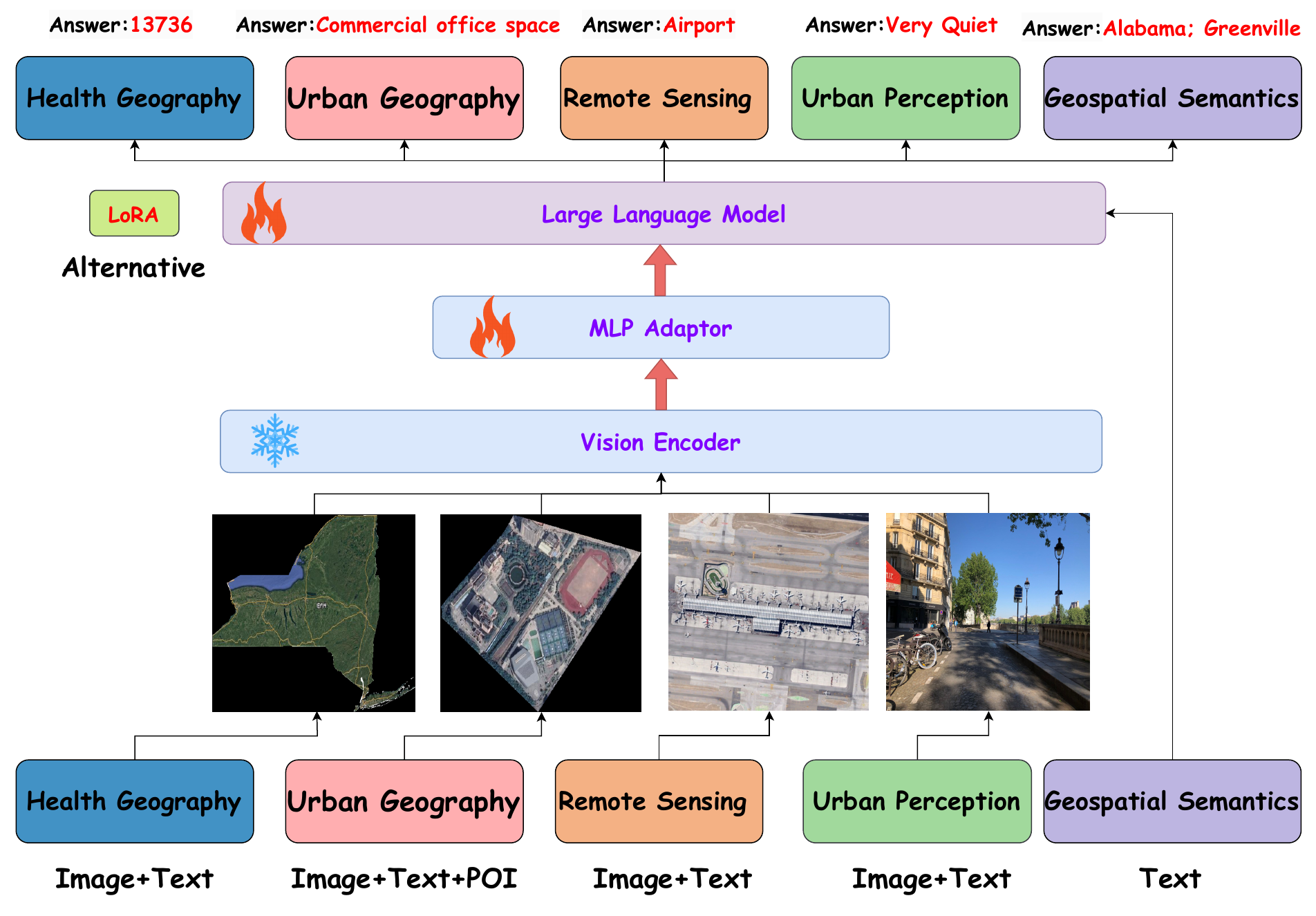}
        \vspace{-3ex}   

    \caption{The training framework overview of the OmniGeo, which integrates five GeoAI tasks with various modalities.}
    \label{fig:OmniGeo framework}

\end{figure*}

\subsection{Multimodal Instruction Data Construction}

In GeoAI, RS images provide a macro perspective of urban spatial structure or geographical locations, while SVI depict a more granular view of urban street blocks. Various tasks in GeoAI are closely associated with different geographic entities, and captions for remote sensing and street view images will further enhance this collaborative information. Therefore, the core motivation behind constructing a fine-tuning dataset is to equip each task with images (either RS or SVI) and complement them with detailed captions. This will assist in spatial reasoning and the 3D perception of geographic entities by the model.

\noindent\textbf{Data sources.} 
In GeoAI, the primary data structures encompass text, images (e.g., remote sensing and street view images), geographic knowledge graphs, geospatial vector data, videos, and audio.
This paper includes geographic text data from tweets and web pages, geospatial polygon vector data from CDC Wonder (dementia data at the state and county levels in the US), geospatial point and polygon vector data from Amap (POI), remote sensing images from WorldView, street view images from Google Maps, and geospatial polygon vector data from city government websites (urban planning layer data).

\noindent\textbf{Data construction.} 
Based on the aforementioned data sources, we process all the data to construct instruction datasets that can be utilized by LLMs. 
The specific construction methods for different tasks and data modalities are outlined as follows:

For \textbf{Health Geography}, historical death statistics are first retrieved from CDC Wonder and converted into text. 
Data from 1999 to 2019 are used for model training, while data from 2020 are reserved for evaluation.
Moreover, we add the image data~(RS) as supplementary information for the Health Geography to provide modal alignment.
In particular, the WorldView series of RS images are cropped region by region, and captions are generated using InstructBLIP~\cite{NEURIPS2023_9a6a435e}, which are subsequently concatenated with the statistical data. 
This process results in two datasets: one at the US state level and the other at the US county level.

For \textbf{Urban Geography}, POI data for Beijing and Shenzhen are initially retrieved from Gaode Maps and classified into 11 categories. 
Functional region partition layer data for these cities are then obtained from government websites, and the POI data are converted into frequency-based text. 
Similarly, RS images are copped based on layer data, and captions are generated and concatenated with the POI text. 
This process results in two datasets: one for Beijing and one for Shenzhen.

For \textbf{Remote Sensing}, the RS image classification is a common task in geographic space and has abundant resources in the Internet.
We collect the geospatial benchmark datasets AID and UC-Merced. 
AID is a large-scale dataset for RS image scene classification, comprising 30 scene types with a total of 10,000 images, while UC-Merced includes 21 land use types, with 100 images per scene.

For \textbf{Urban Perception}, the Noise indicator dataset is constructed by discretizing the original average labels in the Visual-Soundscapes~\cite{2023_ceus_soundscapes} dataset into four categories (Very Quiet, Quiet, Noisy, and Very Noisy), resulting in two datasets: Noise-SG and Noise-SZ.
The remaining six indicators (Beautiful, Boring, Depressing, Lively, Safe, Wealthy) are similarly processed within the Global Streetscapes~\cite{HOU2024216} dataset, yielding six subsets: GS-Beautiful, GS-Boring, GS-Depressing, GS-Lively, GS-Safe, and GS-Wealthy.
Moreover, we utilize the caption of street view images that are generated with InstructBLIP in this subset to enhance the alignment between modals.

For \textbf{Geospatial Semantics}, textual resources containing geospatial information are abundant.
Thus, existing geospatial benchmark datasets are directly employed. 
LGL, GeoVirus, and NEEL are used for toponym recognition, while HaveyTweet2017 and GeoCorpora are focused on location description recognition. 
The contents is mainly gathered from web pages, tweets, and other sources.

\subsection{Image-Text Instruction Fine-Tuning}
As shown in Figure~\ref{fig:instructions}, to guide the inherent geospatial knowledge of MLLMs, we propose to conduct image-text fine-tuning, which provides an exploration of the potential of MLLMs in GeoAI.

\noindent\textbf{Dementia Death Counts Time Series Forecasting}: As shown in the blue block in Figure~\ref{fig:instructions}, basic LLMs are capable of performing certain time-series prediction tasks.
However, these tasks often exhibit pronounced geographical distribution disparities. 
Factors such as the distribution of urban industrial areas, neighborhood environments, high latitudes, high altitudes, and hilly terrain are all implicated in cognitive decline among patients~\cite{Zhang2024GeospatialAI, FLETCHER2024101708, CLARKE2015849, Wu2015}.
Thus, different regions (states or countries) inherently exhibit distinct patterns in their temporal properties. 
Specifically, we pair each time-series data with corresponding remote sensing images from the relevant regions and further concatenate captions of these images to enhance the correlation between temporal data and geographical characteristics.


\noindent\textbf{Urban Region Function Classification}: 
As shown in the red block in Figure~\ref{fig:instructions}, the goal of this task is to model potential human activity patterns based on POI data from the target region. 
In particular, we integrate RS images and their associated captions to capture high-level semantic information about specific regions, such as the spatial arrangement of urban geographical entities, including shopping malls, schools, and residential districts. 
This approach enables OmniGeo to perceive the distribution of POIs within the region from a broader, more comprehensive perspective. 
The output of the model represents the primary functional category of the region.


\noindent\textbf{RS Image Scene Classification}: As shown in the orange block in Figure~\ref{fig:instructions}, the model processes Rs images and their corresponding captions for specific scenes, with the goal of capturing unified high-level semantics across scenes exhibiting the same geographic context but varying spatial and spectral resolutions. Additionally, the model is designed to differentiate challenging samples, where geographic entities remain consistent but scene types diverge.

\begin{table*}[!t]
  \begin{centering}
  \caption{Results of OmniGeo and the baselines on \textit{Health Geography}, the dementia death counts time series forecasting task.}
  \label{table:Health_eography}
  \vspace{-3ex}
  \begin{tabular}{lcccc|cccc}
  \toprule
  {\multirow{2}[2]{*}{Model}} & \multicolumn{4}{c|}{US state-level} & \multicolumn{4}{c}{US country-level}\\
  \cmidrule(r){2-9} 
  & MSE$\downarrow$ & MAE$\downarrow$ & MAPE$\downarrow$ & $R^2\uparrow$ & MSE$\downarrow$ & MAE$\downarrow$ & MAPE$\downarrow$ & $R^2\uparrow$  \\
   \midrule
   ARIMA & 562768.0000 & 462.0000 & 6.70\% & 0.9840 & 708935.4100 & 374.1000 & 4.71\% & 0.9800 \\
   \midrule
   GPT-4o & 746173.8182 & 420.3636 & 0.07\% & 0.9792 & 1361.8132 & 17.0774 & 0.18\% & 0.9841 \\
   LLaVA1.5-7B & 935866.3636 & 451.8182 & 0.06\% & 0.9739 & 1284.8832 & 16.0810 & 0.16\% & 0.9849 \\
   Qwen2-VL-7B & 1234893.4444 & 688.5556 & 23.45\% & 0.9704 & 263919.3817 & 148.5340 & 383.21\% & - \\
    \midrule
   OmniGeo~(LLaVA) & \textbf{25329.8182} & \textbf{117.6364} & \textbf{0.03\%} & \textbf{0.9993} & \textbf{462.3494} & \textbf{11.6622} & \textbf{0.15\%} & 0.9947 \\
   OmniGeo~(Qwen2) & 192489.0000 & 290.8182 & 0.06\% & 0.9946 & 482.4748 & 12.3655 & 0.16\% & \textbf{0.9948} \\
  \bottomrule
  \end{tabular}
  \end{centering}
\end{table*}

\begin{table}[!t]
  \begin{centering}
  \caption{Results of OmniGeo and the baselines on \textit{Urban Geography} task with urban region function classification.}
  \label{table:Urban geography}
    \vspace{-3ex}
  \begin{tabular}{lcccc}
  \toprule  
  {\multirow{2}[2]{*}{Model}} & \multicolumn{4}{c}{UG-Shenzhen} \\
  \cmidrule(r){2-5}
  & Acc$\uparrow$ & P$\uparrow$ & R$\uparrow$ & F1$\uparrow$  \\
   \midrule
   Place2Vec & 0.5674 & 0.4737 & 0.5674 & 0.5164 \\
   HGI & 0.6620 & 0.6893 & 0.6620 & 0.6754 \\
   \midrule
   BLIP-2 & 0.3567 & 0.1992 & 0.3567 & 0.2556 \\
   GPT-4o & \textbf{0.8277} & 0.8798 & \textbf{0.8277} & 0.8530 \\
   LLaVA1.5-7B & 0.3914 & 0.2755 & 0.3914 & 0.3234 \\
   Qwen2-VL-7B & 0.3670 & 0.2314 & 0.3670 & 0.2839 \\
   \midrule
   OmniGeo~(LLaVA) & 0.8268 & \textbf{0.8872} & 0.8268 & \textbf{0.8559} \\
   OmniGeo~(Qwen2) & \textbf{0.8277} & 0.8823 & \textbf{0.8277} & 0.8541 \\
  \bottomrule
  \end{tabular}
  \end{centering}
\end{table}

\noindent\textbf{Urban Perception Prediction}: As shown in the green block in Figure~\ref{fig:instructions}, the urban perception prediction is different from typical image classification, which typically classifies images based solely on visual instances. 
Instead, the model is required to simulate high-level human perception abilities by analyzing street view images and their associated captions. 
It estimates various perceptual dimensions, including Noise, Beauty, Boredom, Depression, Livelihood, Safety, and Wealth, where high-level perceptual knowledge is more complex and challenging to assess. 
The output of OmniGeo consists of one of four distinct levels for each of the seven perceptual indicators.


\noindent\textbf{Toponym Recognition}: The language capabilities of advanced LLMs are sufficiently powerful, yet extracting frequently occurring and easily confused place names within the same text remains difficult. 
The output of OmniGeo is a semicolon-separated list of toponyms.


\noindent\textbf{Location Description Recognition}: 
This task presents greater challenges than toponym recognition, as real-world location descriptions exhibit considerable disarray due to factors such as cultural differences, varying educational levels, and diverse lifestyle habits, particularly in relation to the handling of special symbols. 
The output is a list of location descriptions separated by semicolons.


\subsection{Training Objective}
To further align and internalize geospatial knowledge with the intrinsic knowledge of MLLMs, 
we select LLaVA1.5 and Qwen2-VL from top open-source MLLMs as the base for OmniGeo. Multimodal instruction fine-tuning data from multiple tasks will be proportionally mixed for joint training, using both LoRA and full supervision fine-tuning approaches. The base models accept default systerm instructions and task-specific queries, with the training goal being to generate task-specific answers in an autoregressive manner.

Specifically, for tokenized training texts, $[x_1, ..., x_{n_x}, y_1, ..., y_{n_y}]$ where $x$ and $y$ denote queries and ground-truth answer, respectively. The training loss $\mathcal{L}_Q$ is defined as:
\begin{equation}
\mathcal{L}_Q = -\log p(y|x)=\sum_{i=1}^{n_y} -\log p(y_i|y_{i-1},...,y_1,x_{n_x}, ..., x_1)
\end{equation}
\raggedbottom
In terms of LoRA (Low-Rank Adaptation) fine-tuning, given the pre-trained weight matrix $W_0 \in \mathbb{R}^{d_{\text{in}} \times d_{\text{out}}}$, the fine-tuning introduces low-rank matrices $A \in \mathbb{R}^{d_{\text{in}} \times r}$ and $B \in \mathbb{R}^{r \times d_{\text{out}}}$, where $r$ is the rank of the adaptation. The objective function is defined as
\begin{equation}
\mathcal{L}_{\text{LoRA}} = \mathbb{E}_{(x, y)} \left[ \mathcal{L}(f(W_0 + A B, x), y) \right]
\end{equation}
where $A \in \mathbb{R}^{d_{\text{in}} \times r}$ and $B \in \mathbb{R}^{r \times d_{\text{out}}}$ are the low-rank matrices learned during fine-tuning, $W_0$ is the pre-trained model weight matrix and $f(W_0 + AB, x)$ is the model output after LoRA fine-tuning. 
LoRA fine-tuning adjusts the low-rank matrices $A$ and $B$ to adapt the model to various geospatial tasks, without requiring the retraining of the entire weight matrix $W_0$.

\section{Experiments}
\subsection{Datasets}
We performed experiments on 19 datasets across 5 subtasks, including 14 multimodal datasets (1-4) and 5 text-only datasets (5), with a total of 128,060 samples, including
(1) \textbf{Health Geography}: Self-constructed Dataset. US state-level and US country-level; (2) \textbf{Urban Geography}: Self-constructed Dataset. UG-Beijing and UG-Shenzhen; (3) \textbf{Remote Sensing}: geographic benchmark dataset. AID~\cite{7907303} and UC-Merced~\cite{Nilsback08}; (4) \textbf{Urban Perception}: self-constructed dataset. For noise indicators: Noise-SG and Noise-SZ. For other indicators: GS-Beautiful, GS-Boring, GS-Depressing, GS-Lively, GS-Safe and GS-Wealthy; and (5) \textbf{Geospatial Semantics}: geographic benchmark dataset. LGL~\cite{5447903}, GeoVirus~\cite{gritta-etal-2018-melbourne}, and NEEL~\cite{HU2023103191} are toponym recognition datasets, whereas HaveyTweet2017~\cite{unknown} and GeoCorpora~\cite{Wallgrün02012018} are location description recognition datasets.

The geospatial semantic analysis datasets is randomly divided into an 80\% training set and a 20\% test set. For the dementia death prediction datasets, it is randomly split into 80\% training set and 20\% test set at both the state and county levels. The urban region functional classification datasets is split using stratified random sampling into 80\% for training and 20\% for testing. For the RS image scene classification datasets, the AID is divided by stratified random sampling into 80\% training and 20\% testing, while UC-Merced uses the official test set. For the urban perception datasets, the 7 urban perception datasets are split using stratified random sampling into 80\% training set and 20\% test set.

\begin{table*}[!t]
  \begin{centering}
  \caption{Results of OmniGeo and the baseline models on the seven \textit{Urban Perception} prediction tasks and \textit{Remote Sensing} (RS) image scene classification task with Weighted-F1 scores.}
  \label{table:Urban Perception}
    \vspace{-3ex}   
  \begin{tabular}{lcccccccc}
  \toprule
  {Model}  & {Noise} & {Beautiful} & {Boring} & {Depressing} & {Lively} & {Safe} & Wealthy & {RS}\\
   \midrule
   AlexNet  & 0.3642  & 0.4343  & 0.4998  & 0.4547 & 0.5144 & 0.4550  & 0.4581 & 0.7410 \\
   ResNet18  & 0.3802  & 0.4146  & 0.4464  & 0.4218  & 0.4819  & 0.4333  & 0.4278 & 0.7564 \\
   ResNet50 & 0.3359 & 0.4332 & 0.4565 & 0.4440 & 0.4977 & 0.4253 & 0.4428 & 0.7252 \\
   DenseNet161 & 0.2940 & 0.4501 & 0.4593 & 0.4587 & 0.4975 & 0.4833 & 0.4727 & 0.7882 \\
   \midrule
   OpenCLIP-B-9B & 0.3063 & 0.2576 & 0.3426 & 0.1976 & 0.3428 & 0.2906 & 0.2554 & 0.5766 \\
   BLIP2 & 0.2506 & 0.2865 & 0.1797 & 0.1151 & 0.2818 & 0.1271 & 0.1538 & 0.3722 \\
   GPT-4o & 0.3133 & 0.2686 & 0.3262 & 0.3485 & 0.1934 & 0.3313 & 0.2862 & 0.7125 \\
   MiniCPM & 0.1927 & 0.1336 & 0.2874 & 0.2418 & 0.2049 & 0.1102 & 0.2287 & 0.0141 \\
   InternVL2-8B & 0.2811 & 0.3051 & 0.1025 & 0.2832 & 0.2671 & 0.2413 & 0.3135 & 0.6792 \\
   LLaVA1.5-7B & 0.2412 & 0.4728 & 0.0747 & 0.1153 & 0.2135 & 0.2342 & 0.2412 & 0.2394 \\
   Qwen2-VL-7B & 0.1209 & 0.2365 & 0.2884 & 0.1604 & 0.2917 & 0.1853 & 0.2993 & 0.5634 \\
    \midrule
   OmniGeo~(LLaVA) & \textbf{0.4407} & 0.5438 & 0.5413 & 0.5159 & 0.6174 & 0.5362 & \textbf{0.5047} & \textbf{0.9303} \\
  
   OmniGeo~(Qwen2) & 0.4397 & \textbf{0.5645} & \textbf{0.5578} & \textbf{0.5740} & \textbf{0.6188} & \textbf{0.6062} & 0.4830 & 0.9104 \\
  \bottomrule
  \end{tabular}
  \label{table: Urban Perception}
  \end{centering}
\end{table*}

\begin{table*}[!t]
  \begin{centering}
  \caption{Results of OmniGeo and the baseline models on \textit{Geospatial Semantics} tasks in terms of the toponym recognition and location description recognition.}
  \label{table:Geospatial Semantics}
  \vspace{-3ex}
  \begin{tabular}{lccccccccc}
  \toprule
  {\multirow{3}{*}{Model}}  & \multicolumn{6}{c}{Toponym Recognition} & \multicolumn{3}{c}{Location Description Recognition}  \\
    \cmidrule(r){2-10}    
  & \multicolumn{3}{c}{LGL} & \multicolumn{3}{c}{GeoVirus} & \multicolumn{3}{c}{HaveyTweet2017} \\
    \cmidrule(r){2-10}
    & P$\uparrow$ & R$\uparrow$ & F1-score$\uparrow$ & P$\uparrow$ & R$\uparrow$ & F1-score$\uparrow$ & P$\uparrow$ & R$\uparrow$ & F1-score$\uparrow$ \\
   \midrule
   Stanford NER & 0.5872 & 0.5329 & 0.5588 & 0.9043 & 0.8351 & 0.8683 & 0.5140 & 0.3816 & 0.4380 \\
   spaCy NER & 0.2390 & 0.5371 & 0.3308 & 0.5248 & 0.8028 & 0.6349 & 0.4261 & 0.4145 & 0.4203 \\
   \midrule
   NeuroTPR & 0.3701 & 0.6578 & 0.4737 & 0.7059 & 0.8929 & 0.7884 & 0.5428 & 0.5686 & 0.5554 \\
   BERT & 0.5248 & 0.6647 & 0.5865 & 0.5670 & 0.7520 & 0.6483 & 0.6376 & 0.7549 & 0.6913 \\
   \midrule
   BLIP-2 & 0.6667 & 0.0042 & 0.0083 & 0.5000 & 0.0039 & 0.0078 & 0.4500 & 0.0501 & 0.0902 \\
   GPT-4o & 0.5285 & 0.8204 & 0.6429 & 0.8841 & 0.8207 & 0.8512 & 0.4434 & 0.6657 & 0.5323 \\
   LLaVA1.5-7B & 0.4160 & 0.5710 & 0.4814 & 0.8081 & 0.5494 & 0.6541 & 0.3705 & 0.5181 & 0.4361 \\
   Qwen2-VL-7B & 0.2673 & 0.3512 & 0.3036 & 0.6164 & 0.3889 & 0.4769 & 0.4237 & 0.0696 & 0.1196 \\
    \midrule
   OmniGeo~(LLaVA) & 0.7140 & 0.6722 & 0.6925 & 0.8917 & 0.8425 & 0.8664 & 0.7366 & \textbf{0.8022} & 0.7680 \\
   OmniGeo~(Qwen2) & \textbf{0.7806} & \textbf{0.8204} & \textbf{0.8000} & \textbf{0.9240} & \textbf{0.9167} & \textbf{0.9203} & \textbf{0.7884} & 0.7577 & \textbf{0.7727} \\
  \bottomrule
  \end{tabular}
  \end{centering}
\end{table*}

\subsection{Evaluation Metrics}
For \textbf{Health Geography}: time series forecasting task. The evaluation metrics include: Mean Absolute Error (MAE), Mean Squared Error (MSE), Mean Absolute Percentage Error (MAPE), and R2-score. For \textbf{Urban Geography}: multimodal multi-class classification task. Considering the issue of class imbalance, the main evaluation metrics are: Precision (P), Recall (R), and Weighted-F1. For \textbf{Remote Sensing}: multimodal multi-class classification task. The evaluation metrics include: P, R, Weighted-F1. For \textbf{Urban Perception}: multimodal multi-class classification task. Considering the issue of class imbalance, the main evaluation metrics are: P, R, Weighted-F1. For \textbf{Geospatial Semantics}: Sequence labeling task. The evaluation metrics are: P, R, F1-score.

\subsection{Baseline Methods}
We use four groups of models as baselines: task-specific ML models, deep neural network models in general domains, top open-source LLMs, MLLMs, and a closed-source model GPT-4o. They are detailed as: \textbf{(1) ARIMA ~\cite{box2015time}}: An advanced time series forecasting ML model, for the temporal relationship of dementia mortality modeling; \textbf{(2) Place2Vec and HGI}: learn region-specific semantic embeddings to effectively perform various downstream tasks; \textbf{(3) AlexNet~\cite{krizhevsky2012imagenet}, ResNet18~\cite{he2016deep}, ResNet50~\cite{he2016deep}, DenseNet161~\cite{huang2017densely}}: Four deep neural network models that have demonstrated strong classification and generalization abilities on non-RS image datasets;
\textbf{(4) Stanford NER, spaCy NER, and NeuroTPR}: General NER tools. NeuroTPR is a recurrent neural network-based model that can extract location information from text; \textbf{(5) BERT~\cite{devlin-etal-2019-bert}, OpenCLIP-B~\cite{cherti2023reproducible}, BLIP2~\cite{10.5555/3618408.3619222}, LLaVA1.5~\cite{NEURIPS2023_6dcf277e}, Qwen2-VL~\cite{Wang2024Qwen2VLEV}, MiniCPM~\cite{yao2024minicpmvgpt4vlevelmllm}, InternVL2~\cite{chen2024internvl, chen2024far}}: Vanilla BERT and six top open-source vision-language models; and \textbf{(6) GPT-4o}: A typical closed-source visual model that outperforms most open-source models in the general domain.


\subsection{Implementation Details}
The base models of OmniGeo are LLaVA1.5-7B and Qwen2-VL-7B, with model parameters and pre-trained weights set according to the official configurations. OmniGeo~(LLaVA) is fine-tuned with either full fine-tuning or LoRA fine-tuning on 8 A6000 GPUs with 48GB memory, training for 3 to 5 epochs, with a learning rate of 2e-6 and a warmup ratio of 0.03. OmniGeo~(Qwen2) is fine-tuned with either full fine-tuning or LoRA fine-tuning on 4 L20 GPUs with 48GB memory, training for 3 to 5 epochs, with a learning rate of 1e-5 and a warmup ratio of 0.1.

\begin{table*}[!t]
\centering
\caption{Results of the modal ablation experiments on the dementia death counts time series forecasting and urban region function classification tasks.}
\label{table:Ablation Study}
    \vspace{-3ex}
\resizebox{\textwidth}{!}{
\begin{tabular}{lccccccccccccc}
\toprule
\multirow{2}{*}{Model} & \multirow{2}{*}{Config.} & \multirow{2}{*}{w/ image} & \multicolumn{4}{c}{US state-level} & \multicolumn{4}{c}{US country-level} & \multicolumn{2}{c}{Urban Geography(UG-Shenzhen)} \\
\cmidrule(lr){4-14}
 & & & MSE$\downarrow$ & MAE$\downarrow$ & MAPE$\downarrow$ & $R^2$$\uparrow$ & MSE$\downarrow$ & MAE$\downarrow$ & MAPE$\downarrow$ & $R^2$$\uparrow$ & P$\uparrow$ & R$\uparrow$ & Weighted-F1$\uparrow$ \\
\midrule
\multirow{4}{*}{{Qwen2-VL-7B}} & LoRA & \cmark & \textbf{198724.8182} & \textbf{250.4545} & \textbf{0.05\%} & \textbf{0.9945} & 509.0063 & 12.3903 & 0.17\% & 0.9946 & \textbf{0.8823} & \textbf{0.8277} & \textbf{0.8541} \\
~ & LoRA & \xmark & 275918.5455 & 281.4545 & 0.05\% & 0.9923 & \textbf{482.4748} & \textbf{12.3655} & \textbf{0.16\%} & \textbf{0.9948} & 0.3509 & 0.4316 & 0.3871 \\
~ & Full & \cmark & \textbf{192489.0000} & \textbf{290.8182} & \textbf{0.06\%} & \textbf{0.9946} & \textbf{3391.0498} & \textbf{15.9542} & \textbf{0.17\%} & \textbf{0.9620} & \textbf{0.8821} & \textbf{0.8212} & \textbf{0.8506} \\
~ & Full & \xmark & 1043293.1429 & 757.1429 & 0.22\% & 0.9793 & 5355.6794 & 28.9660 & 0.21\% & 0.9431 & 0.6049 & 0.4682 & 0.5278 \\
\midrule
\multirow{4}{*}{{LLaVA1.5-7B}} & LoRA & \cmark & \textbf{25329.8182} & \textbf{117.6364} & \textbf{0.03\%} & \textbf{0.9993} & 706.5149 & 13.2911 & 0.16\% & 0.9920 & \textbf{0.8827} & \textbf{0.8305} & \textbf{0.8558} \\
~ & LoRA & \xmark & 135622.0000 & 196.5455 & 0.04\% & 0.9962 & \textbf{668.5937} & \textbf{13.3393} & \textbf{0.16\%} & \textbf{0.9925} & 0.6786 & 0.4026 & 0.5054 \\
~ & Full & \cmark & \textbf{143825.5455} & \textbf{225.3636} & \textbf{0.05\%} & \textbf{0.9960} & \textbf{462.3494} & \textbf{11.6622} & \textbf{0.15\%} & \textbf{0.9947} & \textbf{0.8872} & \textbf{0.8268} & \textbf{0.8559} \\
~ & Full & \xmark & 175533.9091 & 275.1818 & 0.06\% & 0.9951 & 498. 8000 & 12.3462 & 0.17\% & 0.9943 & 0.3868 & 0.4345 & 0.4093 \\
\bottomrule
\end{tabular}}
\end{table*}

\begin{table*}[!t]
  \begin{centering}
    \caption{Results of task ablation on the base model LLaVA1.5-7B.}
    \label{table:task ablation}
    \vspace{-3ex}
    \resizebox{\textwidth}{!}{
  \begin{tabular}{lcccccccccccccc}
  \toprule
  {\multirow{3}{*}{Model}}  & \multicolumn{8}{c}{Health Geography } & \multicolumn{2}{c}{Urban Geography} & \multicolumn{2}{c}{Urban Perception(Noise)} & \multicolumn{2}{c}{RS} \\
    \cmidrule(r){2-15}    
  & \multicolumn{4}{c}{US state-level} & \multicolumn{4}{c}{US country-level} & \multicolumn{1}{c}{UG-Shenzhen} & \multicolumn{1}{c}{UG-Beijing(zs)} & \multicolumn{1}{c}{UP-Singapore} & \multicolumn{1}{c}{UP-Shenzhen(zs)} & \multicolumn{1}{c}{AID} & \multicolumn{1}{c}{UC-Merced(zs)} \\
    \cmidrule(r){2-15}
    & MSE$\downarrow$ & MAE$\downarrow$& MAPE$\downarrow$& $R^2\uparrow$& MSE$\downarrow$& MAE$\downarrow$& MAPE$\downarrow$& $R^2\uparrow$ & F1$\uparrow$ & F1$\uparrow$ & F1$\uparrow$ & F1$\uparrow$ & F1$\uparrow$ & F1$\uparrow$\\
   \midrule
   w/. HG-s \& HG-c & 424105.4545 & 302.3636 & 0.04\% & 0.9881 & 519.6601 & 12.1513 & 0.15\% & 0.9941 & - & - & - & - & - & - \\
   w/. UG-SZ & - & - & - & - & - & - & - & - & \textbf{0.8637} & 0.3460 & - & - & - & - \\
   w/. UP-SG & - & - & - & - & - & - & - & - & - & - & 0.2632 & 0.2410 & - & - \\
   w/. RS-AID & - & - & - & - & - & - & - & - & - & - & - & - & 0.8012 & 0.5832\\
      \midrule
   OmniGeo (Ours) & \textbf{25329.8182} & \textbf{117.6364} & \textbf{0.03\%} & \textbf{0.9993} & \textbf{462.3494} & \textbf{11.6622} & \textbf{0.15\%} & \textbf{0.9947} & 0.8559 & \textbf{0.3557} & \textbf{0.4407} & \textbf{0.4683} & \textbf{0.9303} & \textbf{0.6025} \\
  \bottomrule
  \end{tabular}}
  \end{centering}
\end{table*}

\section{Results}

\subsection{Main Results}
\noindent\textbf{Health Geography} task performance is shown in
Table~\ref{table:Health_eography} where LLaVA1.5 is on par with GPT-4o and both of them surpass the task-specific model ARIMA and the vision LLM Qwen2-VL. 
However, our OmniGeo, leveraging rich geospatial knowledge and multimodal advantages, outperforms all these baselines in both regression metrics and model interpretability scores.

\noindent\textbf{Urban Geography} task results are reported in Table~\ref{table:Urban geography}.
After multi-modal instruction fine-tuning, OmniGeo achieves absolute improvements of 0.3473 and 0.2227 in Weighted-F1 compared to Place2Vec and HGI, respectively.
Besides, it also performs on par with the best closed-source model GPT-4o, which suggests that OmniGeo can utilize RS images and POI data to model regional-level semantic information and human activity patterns. 
Notably, most open-source MLLMs show very low classification performance, which implies that knowledge from the general domain cannot be directly transferred to this task due to the significant domain gap. Nevertheless, our designed strategies address such issues and thus enhance the classification accuracy of OmniGeo.


\noindent\textbf{Remote Sensing and Urban Perception} tasks comparison are presented in Table~\ref{table: Urban Perception}. 
Benefiting from the geospatial knowledge gained during fine-tuning, OmniGeo can understand high-level semantic information of each scene type and achieve the best classification performance.
The typical MLLMs perform even worse than the four DL models, due to the lack of explicit geographic spatial knowledge and its application capabilities. 
The traditional DL models exhibit good classification attributing to the specific fine-tuning.
These observations indicate the importance of grasping or arousing the domain knowledge parametrically.


\noindent\textbf{Geospatial Semantics} task results are shown in Table~\ref{table:Geospatial Semantics}, where our OmniGeo still outperforms the other systems based on various types of backbone models.
It even outperforms the strong baseline GPT-4o, which indicates that with proper optimization, open-source models can be viable alternatives for geographic knowledge enhancement, offering better performance than the best commercial models in the GeoAI task.

\subsection{Ablation Studies}
To investigate the effectiveness of leveraging multimodal information and the synergy between geospatial tasks for model training, we conduct ablation studies from two perspectives: i) integrating multimodal information via different fine-tuned configurations (e.g., LoRA or full parameters) and ii) the affect of jointly fine-tuning various tasks simultaneously. The results are shown in Table~\ref{table:Ablation Study} and Table~\ref{table:task ablation}, respectively.

\noindent \textbf{Impact of multimodal information.}
As shown in Table~\ref{table:Ablation Study}, fusing multimodal information (image in our cases) significantly improves the performance on top of two backbone models compared to using single modal information (text) only.
The improvements are across various datasets and fine-tuning settings (both LoRA and full parameters).
The slight performance decrease is only observed on the US country-level dataset, which might be because of the low resolution of RS images and noise from atypical country data.



\noindent \textbf{Impact of the jointly fine-tuning.}
From Table~\ref{table:task ablation}, we observe that our OmniGeo~(LLaVA) fine-tuned with multiple tasks jointly shows more robust results compared to the baselines where each task was fine-tuned individually based on LLaVA-1.5-7B.
This is mainly attributed to the cross-modal and cross-task knowledge sharing and transfer via multiple task resources jointly fine-tuning. 
A performance slightly decreased is observed on the Shenzhen dataset, which might imply the inherent noise in multi-task datasets and minor information loss due to domain transfer.

\section{Conclusion and Future Work}

This study contributes a large-scale, high-quality visual instruction-following dataset to the GeoAI community. Besides, we explore the potential of MLLMs in GeoAI, facilitating cross-modal geographic knowledge sharing and transfer, and develop the first MLLM, OmniGeo, covering all core tasks in GeoAI. It achieves performance surpassing or on par with GPT-4o, effectively addressing the inherent multimodal nature of data in GeoAI and taking the first step towards the deployment of MLLMs in GeoAI. In the future, we can further develop more robust and effective systems for GeoAI, which provide flexible interaction~\cite{mo2023convgqr,mo2023learning} with the domain users~\cite{wang2024user} and support more geographic tasks~\cite{mai2025towards}.


\bibliographystyle{ACM-Reference-Format}
\bibliography{sample-base}


\begin{thebibliography}{69}


\ifx \showCODEN    \undefined \def \showCODEN     #1{\unskip}     \fi
\ifx \showISBNx    \undefined \def \showISBNx     #1{\unskip}     \fi
\ifx \showISBNxiii \undefined \def \showISBNxiii  #1{\unskip}     \fi
\ifx \showISSN     \undefined \def \showISSN      #1{\unskip}     \fi
\ifx \showLCCN     \undefined \def \showLCCN      #1{\unskip}     \fi
\ifx \shownote     \undefined \def \shownote      #1{#1}          \fi
\ifx \showarticletitle \undefined \def \showarticletitle #1{#1}   \fi
\ifx \showURL      \undefined \def \showURL       {\relax}        \fi
\providecommand\bibfield[2]{#2}
\providecommand\bibinfo[2]{#2}
\providecommand\natexlab[1]{#1}
\providecommand\showeprint[2][]{arXiv:#2}

\bibitem[Akushevich et~al\mbox{.}(2021)]%
        {akushevich2021geographic}
\bibfield{author}{\bibinfo{person}{Igor Akushevich}, \bibinfo{person}{Arseniy~P Yashkin}, \bibinfo{person}{Anatoliy~I Yashin}, {and} \bibinfo{person}{Julia Kravchenko}.} \bibinfo{year}{2021}\natexlab{}.
\newblock \showarticletitle{Geographic disparities in mortality from Alzheimer's disease and related dementias}.
\newblock \bibinfo{journal}{\emph{Journal of the American Geriatrics Society}} \bibinfo{volume}{69}, \bibinfo{number}{8} (\bibinfo{year}{2021}), \bibinfo{pages}{2306--2315}.
\newblock


\bibitem[Askarizad and He(2022)]%
        {askarizad2022perception}
\bibfield{author}{\bibinfo{person}{Reza Askarizad} {and} \bibinfo{person}{Jinliao He}.} \bibinfo{year}{2022}\natexlab{}.
\newblock \showarticletitle{Perception of spatial legibility and its association with human mobility patterns: An empirical assessment of the historical districts in rasht, iran}.
\newblock \bibinfo{journal}{\emph{International Journal of Environmental Research and Public Health}} \bibinfo{volume}{19}, \bibinfo{number}{22} (\bibinfo{year}{2022}), \bibinfo{pages}{15258}.
\newblock


\bibitem[Box et~al\mbox{.}(2015)]%
        {box2015time}
\bibfield{author}{\bibinfo{person}{George~EP Box}, \bibinfo{person}{Gwilym~M Jenkins}, \bibinfo{person}{Gregory~C Reinsel}, {and} \bibinfo{person}{Greta~M Ljung}.} \bibinfo{year}{2015}\natexlab{}.
\newblock \bibinfo{booktitle}{\emph{Time series analysis: forecasting and control}}.
\newblock \bibinfo{publisher}{John Wiley \& Sons}.
\newblock


\bibitem[Chen et~al\mbox{.}(2022)]%
        {chen-etal-2022-convfinqa}
\bibfield{author}{\bibinfo{person}{Zhiyu Chen}, \bibinfo{person}{Shiyang Li}, \bibinfo{person}{Charese Smiley}, \bibinfo{person}{Zhiqiang Ma}, \bibinfo{person}{Sameena Shah}, {and} \bibinfo{person}{William~Yang Wang}.} \bibinfo{year}{2022}\natexlab{}.
\newblock \showarticletitle{{C}onv{F}in{QA}: Exploring the Chain of Numerical Reasoning in Conversational Finance Question Answering}. In \bibinfo{booktitle}{\emph{Proceedings of the 2022 Conference on Empirical Methods in Natural Language Processing}}, \bibfield{editor}{\bibinfo{person}{Yoav Goldberg}, \bibinfo{person}{Zornitsa Kozareva}, {and} \bibinfo{person}{Yue Zhang}} (Eds.). \bibinfo{publisher}{Association for Computational Linguistics}, \bibinfo{address}{Abu Dhabi, United Arab Emirates}, \bibinfo{pages}{6279--6292}.
\newblock
\href{https://doi.org/10.18653/v1/2022.emnlp-main.421}{doi:\nolinkurl{10.18653/v1/2022.emnlp-main.421}}


\bibitem[Chen et~al\mbox{.}(2024a)]%
        {chen2024far}
\bibfield{author}{\bibinfo{person}{Zhe Chen}, \bibinfo{person}{Weiyun Wang}, \bibinfo{person}{Hao Tian}, \bibinfo{person}{Shenglong Ye}, \bibinfo{person}{Zhangwei Gao}, \bibinfo{person}{Erfei Cui}, \bibinfo{person}{Wenwen Tong}, \bibinfo{person}{Kongzhi Hu}, \bibinfo{person}{Jiapeng Luo}, \bibinfo{person}{Zheng Ma}, {et~al\mbox{.}}} \bibinfo{year}{2024}\natexlab{a}.
\newblock \showarticletitle{How Far Are We to GPT-4V? Closing the Gap to Commercial Multimodal Models with Open-Source Suites}.
\newblock \bibinfo{journal}{\emph{arXiv preprint arXiv:2404.16821}} (\bibinfo{year}{2024}).
\newblock


\bibitem[Chen et~al\mbox{.}(2024b)]%
        {chen2024internvl}
\bibfield{author}{\bibinfo{person}{Zhe Chen}, \bibinfo{person}{Jiannan Wu}, \bibinfo{person}{Wenhai Wang}, \bibinfo{person}{Weijie Su}, \bibinfo{person}{Guo Chen}, \bibinfo{person}{Sen Xing}, \bibinfo{person}{Muyan Zhong}, \bibinfo{person}{Qinglong Zhang}, \bibinfo{person}{Xizhou Zhu}, \bibinfo{person}{Lewei Lu}, {et~al\mbox{.}}} \bibinfo{year}{2024}\natexlab{b}.
\newblock \showarticletitle{Internvl: Scaling up vision foundation models and aligning for generic visual-linguistic tasks}. In \bibinfo{booktitle}{\emph{Proceedings of the IEEE/CVF Conference on Computer Vision and Pattern Recognition}}. \bibinfo{pages}{24185--24198}.
\newblock


\bibitem[Cherti et~al\mbox{.}(2023)]%
        {cherti2023reproducible}
\bibfield{author}{\bibinfo{person}{Mehdi Cherti}, \bibinfo{person}{Romain Beaumont}, \bibinfo{person}{Ross Wightman}, \bibinfo{person}{Mitchell Wortsman}, \bibinfo{person}{Gabriel Ilharco}, \bibinfo{person}{Cade Gordon}, \bibinfo{person}{Christoph Schuhmann}, \bibinfo{person}{Ludwig Schmidt}, {and} \bibinfo{person}{Jenia Jitsev}.} \bibinfo{year}{2023}\natexlab{}.
\newblock \showarticletitle{Reproducible scaling laws for contrastive language-image learning}. In \bibinfo{booktitle}{\emph{Proceedings of the IEEE/CVF Conference on Computer Vision and Pattern Recognition}}. \bibinfo{pages}{2818--2829}.
\newblock


\bibitem[Choi et~al\mbox{.}(2021)]%
        {CHOI2021103091}
\bibfield{author}{\bibinfo{person}{Seongjin Choi}, \bibinfo{person}{Jiwon Kim}, {and} \bibinfo{person}{Hwasoo Yeo}.} \bibinfo{year}{2021}\natexlab{}.
\newblock \showarticletitle{TrajGAIL: Generating urban vehicle trajectories using generative adversarial imitation learning}.
\newblock \bibinfo{journal}{\emph{Transportation Research Part C: Emerging Technologies}}  \bibinfo{volume}{128} (\bibinfo{year}{2021}), \bibinfo{pages}{103091}.
\newblock
\showISSN{0968-090X}
\href{https://doi.org/10.1016/j.trc.2021.103091}{doi:\nolinkurl{10.1016/j.trc.2021.103091}}


\bibitem[Clarke et~al\mbox{.}(2015)]%
        {CLARKE2015849}
\bibfield{author}{\bibinfo{person}{Philippa~J. Clarke}, \bibinfo{person}{Jennifer Weuve}, \bibinfo{person}{Lisa Barnes}, \bibinfo{person}{Denis~A. Evans}, {and} \bibinfo{person}{Carlos~F. {Mendes de Leon}}.} \bibinfo{year}{2015}\natexlab{}.
\newblock \showarticletitle{Cognitive decline and the neighborhood environment}.
\newblock \bibinfo{journal}{\emph{Annals of Epidemiology}} \bibinfo{volume}{25}, \bibinfo{number}{11} (\bibinfo{year}{2015}), \bibinfo{pages}{849--854}.
\newblock
\showISSN{1047-2797}
\href{https://doi.org/10.1016/j.annepidem.2015.07.001}{doi:\nolinkurl{10.1016/j.annepidem.2015.07.001}}


\bibitem[Dai et~al\mbox{.}(2023)]%
        {NEURIPS2023_9a6a435e}
\bibfield{author}{\bibinfo{person}{Wenliang Dai}, \bibinfo{person}{Junnan Li}, \bibinfo{person}{DONGXU LI}, \bibinfo{person}{Anthony Tiong}, \bibinfo{person}{Junqi Zhao}, \bibinfo{person}{Weisheng Wang}, \bibinfo{person}{Boyang Li}, \bibinfo{person}{Pascale~N Fung}, {and} \bibinfo{person}{Steven Hoi}.} \bibinfo{year}{2023}\natexlab{}.
\newblock \showarticletitle{InstructBLIP: Towards General-purpose Vision-Language Models with Instruction Tuning}. In \bibinfo{booktitle}{\emph{Advances in Neural Information Processing Systems}}, \bibfield{editor}{\bibinfo{person}{A.~Oh}, \bibinfo{person}{T.~Naumann}, \bibinfo{person}{A.~Globerson}, \bibinfo{person}{K.~Saenko}, \bibinfo{person}{M.~Hardt}, {and} \bibinfo{person}{S.~Levine}} (Eds.), Vol.~\bibinfo{volume}{36}. \bibinfo{publisher}{Curran Associates, Inc.}, \bibinfo{pages}{49250--49267}.
\newblock
\urldef\tempurl%
\url{https://proceedings.neurips.cc/paper_files/paper/2023/file/9a6a435e75419a836fe47ab6793623e6-Paper-Conference.pdf}
\showURL{%
\tempurl}


\bibitem[DeepSeek-AI et~al\mbox{.}(2024)]%
        {deepseekai2024deepseekv3technicalreport}
\bibfield{author}{\bibinfo{person}{DeepSeek-AI}, \bibinfo{person}{Aixin Liu}, \bibinfo{person}{Bei Feng}, \bibinfo{person}{Bing Xue}, \bibinfo{person}{Bingxuan Wang}, \bibinfo{person}{Bochao Wu}, \bibinfo{person}{Chengda Lu}, \bibinfo{person}{Chenggang Zhao}, \bibinfo{person}{Chengqi Deng}, \bibinfo{person}{Chenyu Zhang}, \bibinfo{person}{Chong Ruan}, \bibinfo{person}{Damai Dai}, \bibinfo{person}{Daya Guo}, \bibinfo{person}{Dejian Yang}, \bibinfo{person}{Deli Chen}, \bibinfo{person}{Dongjie Ji}, \bibinfo{person}{Erhang Li}, \bibinfo{person}{Fangyun Lin}, \bibinfo{person}{Fucong Dai}, \bibinfo{person}{Fuli Luo}, \bibinfo{person}{Guangbo Hao}, \bibinfo{person}{Guanting Chen}, \bibinfo{person}{Guowei Li}, \bibinfo{person}{H. Zhang}, \bibinfo{person}{Han Bao}, \bibinfo{person}{Hanwei Xu}, \bibinfo{person}{Haocheng Wang}, \bibinfo{person}{Haowei Zhang}, \bibinfo{person}{Honghui Ding}, \bibinfo{person}{Huajian Xin}, \bibinfo{person}{Huazuo Gao}, \bibinfo{person}{Hui Li}, \bibinfo{person}{Hui Qu},
  \bibinfo{person}{J.~L. Cai}, \bibinfo{person}{Jian Liang}, \bibinfo{person}{Jianzhong Guo}, \bibinfo{person}{Jiaqi Ni}, \bibinfo{person}{Jiashi Li}, \bibinfo{person}{Jiawei Wang}, \bibinfo{person}{Jin Chen}, \bibinfo{person}{Jingchang Chen}, \bibinfo{person}{Jingyang Yuan}, \bibinfo{person}{Junjie Qiu}, \bibinfo{person}{Junlong Li}, \bibinfo{person}{Junxiao Song}, \bibinfo{person}{Kai Dong}, \bibinfo{person}{Kai Hu}, \bibinfo{person}{Kaige Gao}, \bibinfo{person}{Kang Guan}, \bibinfo{person}{Kexin Huang}, \bibinfo{person}{Kuai Yu}, \bibinfo{person}{Lean Wang}, \bibinfo{person}{Lecong Zhang}, \bibinfo{person}{Lei Xu}, \bibinfo{person}{Leyi Xia}, \bibinfo{person}{Liang Zhao}, \bibinfo{person}{Litong Wang}, \bibinfo{person}{Liyue Zhang}, \bibinfo{person}{Meng Li}, \bibinfo{person}{Miaojun Wang}, \bibinfo{person}{Mingchuan Zhang}, \bibinfo{person}{Minghua Zhang}, \bibinfo{person}{Minghui Tang}, \bibinfo{person}{Mingming Li}, \bibinfo{person}{Ning Tian}, \bibinfo{person}{Panpan Huang}, \bibinfo{person}{Peiyi
  Wang}, \bibinfo{person}{Peng Zhang}, \bibinfo{person}{Qiancheng Wang}, \bibinfo{person}{Qihao Zhu}, \bibinfo{person}{Qinyu Chen}, \bibinfo{person}{Qiushi Du}, \bibinfo{person}{R.~J. Chen}, \bibinfo{person}{R.~L. Jin}, \bibinfo{person}{Ruiqi Ge}, \bibinfo{person}{Ruisong Zhang}, \bibinfo{person}{Ruizhe Pan}, \bibinfo{person}{Runji Wang}, \bibinfo{person}{Runxin Xu}, \bibinfo{person}{Ruoyu Zhang}, \bibinfo{person}{Ruyi Chen}, \bibinfo{person}{S.~S. Li}, \bibinfo{person}{Shanghao Lu}, \bibinfo{person}{Shangyan Zhou}, \bibinfo{person}{Shanhuang Chen}, \bibinfo{person}{Shaoqing Wu}, \bibinfo{person}{Shengfeng Ye}, \bibinfo{person}{Shengfeng Ye}, \bibinfo{person}{Shirong Ma}, \bibinfo{person}{Shiyu Wang}, \bibinfo{person}{Shuang Zhou}, \bibinfo{person}{Shuiping Yu}, \bibinfo{person}{Shunfeng Zhou}, \bibinfo{person}{Shuting Pan}, \bibinfo{person}{T. Wang}, \bibinfo{person}{Tao Yun}, \bibinfo{person}{Tian Pei}, \bibinfo{person}{Tianyu Sun}, \bibinfo{person}{W.~L. Xiao}, \bibinfo{person}{Wangding Zeng},
  \bibinfo{person}{Wanjia Zhao}, \bibinfo{person}{Wei An}, \bibinfo{person}{Wen Liu}, \bibinfo{person}{Wenfeng Liang}, \bibinfo{person}{Wenjun Gao}, \bibinfo{person}{Wenqin Yu}, \bibinfo{person}{Wentao Zhang}, \bibinfo{person}{X.~Q. Li}, \bibinfo{person}{Xiangyue Jin}, \bibinfo{person}{Xianzu Wang}, \bibinfo{person}{Xiao Bi}, \bibinfo{person}{Xiaodong Liu}, \bibinfo{person}{Xiaohan Wang}, \bibinfo{person}{Xiaojin Shen}, \bibinfo{person}{Xiaokang Chen}, \bibinfo{person}{Xiaokang Zhang}, \bibinfo{person}{Xiaosha Chen}, \bibinfo{person}{Xiaotao Nie}, \bibinfo{person}{Xiaowen Sun}, \bibinfo{person}{Xiaoxiang Wang}, \bibinfo{person}{Xin Cheng}, \bibinfo{person}{Xin Liu}, \bibinfo{person}{Xin Xie}, \bibinfo{person}{Xingchao Liu}, \bibinfo{person}{Xingkai Yu}, \bibinfo{person}{Xinnan Song}, \bibinfo{person}{Xinxia Shan}, \bibinfo{person}{Xinyi Zhou}, \bibinfo{person}{Xinyu Yang}, \bibinfo{person}{Xinyuan Li}, \bibinfo{person}{Xuecheng Su}, \bibinfo{person}{Xuheng Lin}, \bibinfo{person}{Y.~K. Li},
  \bibinfo{person}{Y.~Q. Wang}, \bibinfo{person}{Y.~X. Wei}, \bibinfo{person}{Y.~X. Zhu}, \bibinfo{person}{Yang Zhang}, \bibinfo{person}{Yanhong Xu}, \bibinfo{person}{Yanhong Xu}, \bibinfo{person}{Yanping Huang}, \bibinfo{person}{Yao Li}, \bibinfo{person}{Yao Zhao}, \bibinfo{person}{Yaofeng Sun}, \bibinfo{person}{Yaohui Li}, \bibinfo{person}{Yaohui Wang}, \bibinfo{person}{Yi Yu}, \bibinfo{person}{Yi Zheng}, \bibinfo{person}{Yichao Zhang}, \bibinfo{person}{Yifan Shi}, \bibinfo{person}{Yiliang Xiong}, \bibinfo{person}{Ying He}, \bibinfo{person}{Ying Tang}, \bibinfo{person}{Yishi Piao}, \bibinfo{person}{Yisong Wang}, \bibinfo{person}{Yixuan Tan}, \bibinfo{person}{Yiyang Ma}, \bibinfo{person}{Yiyuan Liu}, \bibinfo{person}{Yongqiang Guo}, \bibinfo{person}{Yu Wu}, \bibinfo{person}{Yuan Ou}, \bibinfo{person}{Yuchen Zhu}, \bibinfo{person}{Yuduan Wang}, \bibinfo{person}{Yue Gong}, \bibinfo{person}{Yuheng Zou}, \bibinfo{person}{Yujia He}, \bibinfo{person}{Yukun Zha}, \bibinfo{person}{Yunfan Xiong},
  \bibinfo{person}{Yunxian Ma}, \bibinfo{person}{Yuting Yan}, \bibinfo{person}{Yuxiang Luo}, \bibinfo{person}{Yuxiang You}, \bibinfo{person}{Yuxuan Liu}, \bibinfo{person}{Yuyang Zhou}, \bibinfo{person}{Z.~F. Wu}, \bibinfo{person}{Z.~Z. Ren}, \bibinfo{person}{Zehui Ren}, \bibinfo{person}{Zhangli Sha}, \bibinfo{person}{Zhe Fu}, \bibinfo{person}{Zhean Xu}, \bibinfo{person}{Zhen Huang}, \bibinfo{person}{Zhen Zhang}, \bibinfo{person}{Zhenda Xie}, \bibinfo{person}{Zhengyan Zhang}, \bibinfo{person}{Zhewen Hao}, \bibinfo{person}{Zhibin Gou}, \bibinfo{person}{Zhicheng Ma}, \bibinfo{person}{Zhigang Yan}, \bibinfo{person}{Zhihong Shao}, \bibinfo{person}{Zhipeng Xu}, \bibinfo{person}{Zhiyu Wu}, \bibinfo{person}{Zhongyu Zhang}, \bibinfo{person}{Zhuoshu Li}, \bibinfo{person}{Zihui Gu}, \bibinfo{person}{Zijia Zhu}, \bibinfo{person}{Zijun Liu}, \bibinfo{person}{Zilin Li}, \bibinfo{person}{Ziwei Xie}, \bibinfo{person}{Ziyang Song}, \bibinfo{person}{Ziyi Gao}, {and} \bibinfo{person}{Zizheng Pan}.}
  \bibinfo{year}{2024}\natexlab{}.
\newblock \bibinfo{title}{DeepSeek-V3 Technical Report}.
\newblock
\showeprint[arxiv]{2412.19437}~[cs.CL]
\urldef\tempurl%
\url{https://arxiv.org/abs/2412.19437}
\showURL{%
\tempurl}


\bibitem[Devlin et~al\mbox{.}(2019)]%
        {devlin-etal-2019-bert}
\bibfield{author}{\bibinfo{person}{Jacob Devlin}, \bibinfo{person}{Ming-Wei Chang}, \bibinfo{person}{Kenton Lee}, {and} \bibinfo{person}{Kristina Toutanova}.} \bibinfo{year}{2019}\natexlab{}.
\newblock \showarticletitle{{BERT}: Pre-training of Deep Bidirectional Transformers for Language Understanding}. In \bibinfo{booktitle}{\emph{Proceedings of the 2019 Conference of the North {A}merican Chapter of the Association for Computational Linguistics: Human Language Technologies, Volume 1 (Long and Short Papers)}}, \bibfield{editor}{\bibinfo{person}{Jill Burstein}, \bibinfo{person}{Christy Doran}, {and} \bibinfo{person}{Thamar Solorio}} (Eds.). \bibinfo{publisher}{Association for Computational Linguistics}, \bibinfo{address}{Minneapolis, Minnesota}, \bibinfo{pages}{4171--4186}.
\newblock
\href{https://doi.org/10.18653/v1/N19-1423}{doi:\nolinkurl{10.18653/v1/N19-1423}}


\bibitem[Faisal and Anastasopoulos(2022)]%
        {faisal2022geographic}
\bibfield{author}{\bibinfo{person}{Fahim Faisal} {and} \bibinfo{person}{Antonios Anastasopoulos}.} \bibinfo{year}{2022}\natexlab{}.
\newblock \showarticletitle{Geographic and geopolitical biases of language models}.
\newblock \bibinfo{journal}{\emph{arXiv preprint arXiv:2212.10408}} (\bibinfo{year}{2022}).
\newblock


\bibitem[Finkel et~al\mbox{.}(2005)]%
        {finkel2005incorporating}
\bibfield{author}{\bibinfo{person}{Jenny~Rose Finkel}, \bibinfo{person}{Trond Grenager}, {and} \bibinfo{person}{Christopher~D Manning}.} \bibinfo{year}{2005}\natexlab{}.
\newblock \showarticletitle{Incorporating non-local information into information extraction systems by gibbs sampling}. In \bibinfo{booktitle}{\emph{Proceedings of the 43rd annual meeting of the association for computational linguistics (ACL’05)}}. \bibinfo{pages}{363--370}.
\newblock


\bibitem[Fletcher et~al\mbox{.}(2024)]%
        {FLETCHER2024101708}
\bibfield{author}{\bibinfo{person}{Jason Fletcher}, \bibinfo{person}{Katie Jajtner}, {and} \bibinfo{person}{Jinho Kim}.} \bibinfo{year}{2024}\natexlab{}.
\newblock \showarticletitle{Geographic disparities in Alzheimer's disease and related dementia mortality in the US: Comparing impacts of place of birth and place of residence}.
\newblock \bibinfo{journal}{\emph{SSM - Population Health}}  \bibinfo{volume}{27} (\bibinfo{year}{2024}), \bibinfo{pages}{101708}.
\newblock
\showISSN{2352-8273}
\href{https://doi.org/10.1016/j.ssmph.2024.101708}{doi:\nolinkurl{10.1016/j.ssmph.2024.101708}}


\bibitem[Gan et~al\mbox{.}(2023)]%
        {gan2023largelanguagemodelseducation}
\bibfield{author}{\bibinfo{person}{Wensheng Gan}, \bibinfo{person}{Zhenlian Qi}, \bibinfo{person}{Jiayang Wu}, {and} \bibinfo{person}{Jerry Chun-Wei Lin}.} \bibinfo{year}{2023}\natexlab{}.
\newblock \bibinfo{title}{Large Language Models in Education: Vision and Opportunities}.
\newblock
\showeprint[arxiv]{2311.13160}~[cs.AI]
\urldef\tempurl%
\url{https://arxiv.org/abs/2311.13160}
\showURL{%
\tempurl}


\bibitem[Gritta et~al\mbox{.}(2018)]%
        {gritta-etal-2018-melbourne}
\bibfield{author}{\bibinfo{person}{Milan Gritta}, \bibinfo{person}{Mohammad~Taher Pilehvar}, {and} \bibinfo{person}{Nigel Collier}.} \bibinfo{year}{2018}\natexlab{}.
\newblock \showarticletitle{Which {M}elbourne? Augmenting Geocoding with Maps}. In \bibinfo{booktitle}{\emph{Proceedings of the 56th Annual Meeting of the Association for Computational Linguistics (Volume 1: Long Papers)}}, \bibfield{editor}{\bibinfo{person}{Iryna Gurevych} {and} \bibinfo{person}{Yusuke Miyao}} (Eds.). \bibinfo{publisher}{Association for Computational Linguistics}, \bibinfo{address}{Melbourne, Australia}, \bibinfo{pages}{1285--1296}.
\newblock
\href{https://doi.org/10.18653/v1/P18-1119}{doi:\nolinkurl{10.18653/v1/P18-1119}}


\bibitem[He et~al\mbox{.}(2016)]%
        {he2016deep}
\bibfield{author}{\bibinfo{person}{Kaiming He}, \bibinfo{person}{Xiangyu Zhang}, \bibinfo{person}{Shaoqing Ren}, {and} \bibinfo{person}{Jian Sun}.} \bibinfo{year}{2016}\natexlab{}.
\newblock \showarticletitle{Deep residual learning for image recognition}. In \bibinfo{booktitle}{\emph{Proceedings of the IEEE conference on computer vision and pattern recognition}}. \bibinfo{pages}{770--778}.
\newblock


\bibitem[He et~al\mbox{.}(2018)]%
        {he2018remote}
\bibfield{author}{\bibinfo{person}{Nanjun He}, \bibinfo{person}{Leyuan Fang}, \bibinfo{person}{Shutao Li}, \bibinfo{person}{Antonio Plaza}, {and} \bibinfo{person}{Javier Plaza}.} \bibinfo{year}{2018}\natexlab{}.
\newblock \showarticletitle{Remote sensing scene classification using multilayer stacked covariance pooling}.
\newblock \bibinfo{journal}{\emph{IEEE Transactions on Geoscience and Remote Sensing}} \bibinfo{volume}{56}, \bibinfo{number}{12} (\bibinfo{year}{2018}), \bibinfo{pages}{6899--6910}.
\newblock


\bibitem[Hollmann et~al\mbox{.}(2025)]%
        {Hollmann2025}
\bibfield{author}{\bibinfo{person}{Noah Hollmann}, \bibinfo{person}{Samuel Müller}, \bibinfo{person}{Lennart Purucker}, \bibinfo{person}{Arjun Krishnakumar}, \bibinfo{person}{Max Körfer}, \bibinfo{person}{Shi~Bin Hoo}, \bibinfo{person}{Robin~Tibor Schirrmeister}, {and} \bibinfo{person}{Frank Hutter}.} \bibinfo{year}{2025}\natexlab{}.
\newblock \showarticletitle{Accurate predictions on small data with a tabular foundation model}.
\newblock \bibinfo{journal}{\emph{Nature}} \bibinfo{volume}{637}, \bibinfo{number}{8045} (\bibinfo{date}{jan} \bibinfo{year}{2025}), \bibinfo{pages}{319--326}.
\newblock
\showISSN{1476-4687}
\href{https://doi.org/10.1038/s41586-024-08328-6}{doi:\nolinkurl{10.1038/s41586-024-08328-6}}


\bibitem[Honnibal(2017)]%
        {honnibal2017spacy}
\bibfield{author}{\bibinfo{person}{Matthew Honnibal}.} \bibinfo{year}{2017}\natexlab{}.
\newblock \showarticletitle{spaCy 2: Natural language understanding with Bloom embeddings, convolutional neural networks and incremental parsing}.
\newblock \bibinfo{journal}{\emph{(No Title)}} (\bibinfo{year}{2017}).
\newblock


\bibitem[Hou et~al\mbox{.}(2024)]%
        {HOU2024216}
\bibfield{author}{\bibinfo{person}{Yujun Hou}, \bibinfo{person}{Matias Quintana}, \bibinfo{person}{Maxim Khomiakov}, \bibinfo{person}{Winston Yap}, \bibinfo{person}{Jiani Ouyang}, \bibinfo{person}{Koichi Ito}, \bibinfo{person}{Zeyu Wang}, \bibinfo{person}{Tianhong Zhao}, {and} \bibinfo{person}{Filip Biljecki}.} \bibinfo{year}{2024}\natexlab{}.
\newblock \showarticletitle{Global Streetscapes — A comprehensive dataset of 10 million street-level images across 688 cities for urban science and analytics}.
\newblock \bibinfo{journal}{\emph{ISPRS Journal of Photogrammetry and Remote Sensing}}  \bibinfo{volume}{215} (\bibinfo{year}{2024}), \bibinfo{pages}{216--238}.
\newblock
\showISSN{0924-2716}
\href{https://doi.org/10.1016/j.isprsjprs.2024.06.023}{doi:\nolinkurl{10.1016/j.isprsjprs.2024.06.023}}


\bibitem[Hu et~al\mbox{.}(2023)]%
        {HU2023103191}
\bibfield{author}{\bibinfo{person}{Xuke Hu}, \bibinfo{person}{Yeran Sun}, \bibinfo{person}{Jens Kersten}, \bibinfo{person}{Zhiyong Zhou}, \bibinfo{person}{Friederike Klan}, {and} \bibinfo{person}{Hongchao Fan}.} \bibinfo{year}{2023}\natexlab{}.
\newblock \showarticletitle{How can voting mechanisms improve the robustness and generalizability of toponym disambiguation?}
\newblock \bibinfo{journal}{\emph{International Journal of Applied Earth Observation and Geoinformation}}  \bibinfo{volume}{117} (\bibinfo{year}{2023}), \bibinfo{pages}{103191}.
\newblock
\showISSN{1569-8432}
\href{https://doi.org/10.1016/j.jag.2023.103191}{doi:\nolinkurl{10.1016/j.jag.2023.103191}}


\bibitem[Hu and Wang(2020)]%
        {unknown}
\bibfield{author}{\bibinfo{person}{Yingjie Hu} {and} \bibinfo{person}{Jimin Wang}.} \bibinfo{year}{2020}\natexlab{}.
\newblock \bibinfo{title}{How do people describe locations during a natural disaster: an analysis of tweets from Hurricane Harvey}.
\newblock
\href{https://doi.org/10.48550/arXiv.2009.12914}{doi:\nolinkurl{10.48550/arXiv.2009.12914}}


\bibitem[Huang et~al\mbox{.}(2017)]%
        {huang2017densely}
\bibfield{author}{\bibinfo{person}{Gao Huang}, \bibinfo{person}{Zhuang Liu}, \bibinfo{person}{Laurens Van Der~Maaten}, {and} \bibinfo{person}{Kilian~Q Weinberger}.} \bibinfo{year}{2017}\natexlab{}.
\newblock \showarticletitle{Densely connected convolutional networks}. In \bibinfo{booktitle}{\emph{Proceedings of the IEEE conference on computer vision and pattern recognition}}. \bibinfo{pages}{4700--4708}.
\newblock


\bibitem[Huang et~al\mbox{.}(2024)]%
        {huang2024survey}
\bibfield{author}{\bibinfo{person}{Kaiyu Huang}, \bibinfo{person}{Fengran Mo}, \bibinfo{person}{Xinyu Zhang}, \bibinfo{person}{Hongliang Li}, \bibinfo{person}{You Li}, \bibinfo{person}{Yuanchi Zhang}, \bibinfo{person}{Weijian Yi}, \bibinfo{person}{Yulong Mao}, \bibinfo{person}{Jinchen Liu}, \bibinfo{person}{Yuzhuang Xu}, {et~al\mbox{.}}} \bibinfo{year}{2024}\natexlab{}.
\newblock \showarticletitle{A survey on large language models with multilingualism: Recent advances and new frontiers}.
\newblock \bibinfo{journal}{\emph{arXiv preprint arXiv:2405.10936}} (\bibinfo{year}{2024}).
\newblock


\bibitem[Huang et~al\mbox{.}(2023)]%
        {huang2023learning}
\bibfield{author}{\bibinfo{person}{Weiming Huang}, \bibinfo{person}{Daokun Zhang}, \bibinfo{person}{Gengchen Mai}, \bibinfo{person}{Xu Guo}, {and} \bibinfo{person}{Lizhen Cui}.} \bibinfo{year}{2023}\natexlab{}.
\newblock \showarticletitle{Learning urban region representations with POIs and hierarchical graph infomax}.
\newblock \bibinfo{journal}{\emph{ISPRS Journal of Photogrammetry and Remote Sensing}}  \bibinfo{volume}{196} (\bibinfo{year}{2023}), \bibinfo{pages}{134--145}.
\newblock


\bibitem[Jan Oliver~Wallgrün and Pezanowski(2018)]%
        {Wallgrün02012018}
\bibfield{author}{\bibinfo{person}{Alan M.~MacEachren Jan Oliver~Wallgrün, Morteza~Karimzadeh} {and} \bibinfo{person}{Scott Pezanowski}.} \bibinfo{year}{2018}\natexlab{}.
\newblock \showarticletitle{GeoCorpora: building a corpus to test and train microblog geoparsers}.
\newblock \bibinfo{journal}{\emph{International Journal of Geographical Information Science}} \bibinfo{volume}{32}, \bibinfo{number}{1} (\bibinfo{year}{2018}), \bibinfo{pages}{1--29}.
\newblock
\href{https://doi.org/10.1080/13658816.2017.1368523}{doi:\nolinkurl{10.1080/13658816.2017.1368523}}
\showeprint{https://doi.org/10.1080/13658816.2017.1368523}


\bibitem[Jiang et~al\mbox{.}(2024)]%
        {jiang-etal-2024-leveraging}
\bibfield{author}{\bibinfo{person}{Hang Jiang}, \bibinfo{person}{Xiajie Zhang}, \bibinfo{person}{Robert Mahari}, \bibinfo{person}{Daniel Kessler}, \bibinfo{person}{Eric Ma}, \bibinfo{person}{Tal August}, \bibinfo{person}{Irene Li}, \bibinfo{person}{Alex Pentland}, \bibinfo{person}{Yoon Kim}, \bibinfo{person}{Deb Roy}, {and} \bibinfo{person}{Jad Kabbara}.} \bibinfo{year}{2024}\natexlab{}.
\newblock \showarticletitle{Leveraging Large Language Models for Learning Complex Legal Concepts through Storytelling}. In \bibinfo{booktitle}{\emph{Proceedings of the 62nd Annual Meeting of the Association for Computational Linguistics (Volume 1: Long Papers)}}, \bibfield{editor}{\bibinfo{person}{Lun-Wei Ku}, \bibinfo{person}{Andre Martins}, {and} \bibinfo{person}{Vivek Srikumar}} (Eds.). \bibinfo{publisher}{Association for Computational Linguistics}, \bibinfo{address}{Bangkok, Thailand}, \bibinfo{pages}{7194--7219}.
\newblock
\href{https://doi.org/10.18653/v1/2024.acl-long.388}{doi:\nolinkurl{10.18653/v1/2024.acl-long.388}}


\bibitem[Jing(2008)]%
        {jing2008remote}
\bibfield{author}{\bibinfo{person}{Yu Jing}.} \bibinfo{year}{2008}\natexlab{}.
\newblock \showarticletitle{Remote sensing semantic model for city planning}.
\newblock \bibinfo{journal}{\emph{Computer Applications, S1}} (\bibinfo{year}{2008}), \bibinfo{pages}{348--435}.
\newblock


\bibitem[Krizhevsky et~al\mbox{.}(2012)]%
        {krizhevsky2012imagenet}
\bibfield{author}{\bibinfo{person}{Alex Krizhevsky}, \bibinfo{person}{Ilya Sutskever}, {and} \bibinfo{person}{Geoffrey~E Hinton}.} \bibinfo{year}{2012}\natexlab{}.
\newblock \showarticletitle{Imagenet classification with deep convolutional neural networks}.
\newblock \bibinfo{journal}{\emph{Advances in neural information processing systems}}  \bibinfo{volume}{25} (\bibinfo{year}{2012}).
\newblock


\bibitem[Kuckreja et~al\mbox{.}(2024)]%
        {kuckreja2024geochat}
\bibfield{author}{\bibinfo{person}{Kartik Kuckreja}, \bibinfo{person}{Muhammad~Sohail Danish}, \bibinfo{person}{Muzammal Naseer}, \bibinfo{person}{Abhijit Das}, \bibinfo{person}{Salman Khan}, {and} \bibinfo{person}{Fahad~Shahbaz Khan}.} \bibinfo{year}{2024}\natexlab{}.
\newblock \showarticletitle{Geochat: Grounded large vision-language model for remote sensing}. In \bibinfo{booktitle}{\emph{Proceedings of the IEEE/CVF Conference on Computer Vision and Pattern Recognition}}. \bibinfo{pages}{27831--27840}.
\newblock


\bibitem[Lee et~al\mbox{.}(2024)]%
        {lee2024llmcxrinstructionfinetunedllmcxr}
\bibfield{author}{\bibinfo{person}{Suhyeon Lee}, \bibinfo{person}{Won~Jun Kim}, \bibinfo{person}{Jinho Chang}, {and} \bibinfo{person}{Jong~Chul Ye}.} \bibinfo{year}{2024}\natexlab{}.
\newblock \bibinfo{title}{LLM-CXR: Instruction-Finetuned LLM for CXR Image Understanding and Generation}.
\newblock
\showeprint[arxiv]{2305.11490}~[cs.CV]
\urldef\tempurl%
\url{https://arxiv.org/abs/2305.11490}
\showURL{%
\tempurl}


\bibitem[Li et~al\mbox{.}(2023b)]%
        {NEURIPS2023_5abcdf8e}
\bibfield{author}{\bibinfo{person}{Chunyuan Li}, \bibinfo{person}{Cliff Wong}, \bibinfo{person}{Sheng Zhang}, \bibinfo{person}{Naoto Usuyama}, \bibinfo{person}{Haotian Liu}, \bibinfo{person}{Jianwei Yang}, \bibinfo{person}{Tristan Naumann}, \bibinfo{person}{Hoifung Poon}, {and} \bibinfo{person}{Jianfeng Gao}.} \bibinfo{year}{2023}\natexlab{b}.
\newblock \showarticletitle{LLaVA-Med: Training a Large Language-and-Vision Assistant for Biomedicine in One Day}. In \bibinfo{booktitle}{\emph{Advances in Neural Information Processing Systems}}, \bibfield{editor}{\bibinfo{person}{A.~Oh}, \bibinfo{person}{T.~Naumann}, \bibinfo{person}{A.~Globerson}, \bibinfo{person}{K.~Saenko}, \bibinfo{person}{M.~Hardt}, {and} \bibinfo{person}{S.~Levine}} (Eds.), Vol.~\bibinfo{volume}{36}. \bibinfo{publisher}{Curran Associates, Inc.}, \bibinfo{pages}{28541--28564}.
\newblock
\urldef\tempurl%
\url{https://proceedings.neurips.cc/paper_files/paper/2023/file/5abcdf8ecdcacba028c6662789194572-Paper-Datasets_and_Benchmarks.pdf}
\showURL{%
\tempurl}


\bibitem[Li et~al\mbox{.}(2023a)]%
        {10.5555/3618408.3619222}
\bibfield{author}{\bibinfo{person}{Junnan Li}, \bibinfo{person}{Dongxu Li}, \bibinfo{person}{Silvio Savarese}, {and} \bibinfo{person}{Steven Hoi}.} \bibinfo{year}{2023}\natexlab{a}.
\newblock \showarticletitle{BLIP-2: bootstrapping language-image pre-training with frozen image encoders and large language models}. In \bibinfo{booktitle}{\emph{Proceedings of the 40th International Conference on Machine Learning}} (Honolulu, Hawaii, USA) \emph{(\bibinfo{series}{ICML'23})}. \bibinfo{publisher}{JMLR.org}, Article \bibinfo{articleno}{814}, \bibinfo{numpages}{13}~pages.
\newblock


\bibitem[Lieberman et~al\mbox{.}(2010)]%
        {5447903}
\bibfield{author}{\bibinfo{person}{Michael~D. Lieberman}, \bibinfo{person}{Hanan Samet}, {and} \bibinfo{person}{Jagan Sankaranarayanan}.} \bibinfo{year}{2010}\natexlab{}.
\newblock \showarticletitle{Geotagging with local lexicons to build indexes for textually-specified spatial data}. In \bibinfo{booktitle}{\emph{2010 IEEE 26th International Conference on Data Engineering (ICDE 2010)}}. \bibinfo{pages}{201--212}.
\newblock
\href{https://doi.org/10.1109/ICDE.2010.5447903}{doi:\nolinkurl{10.1109/ICDE.2010.5447903}}


\bibitem[Liu et~al\mbox{.}(2023)]%
        {NEURIPS2023_6dcf277e}
\bibfield{author}{\bibinfo{person}{Haotian Liu}, \bibinfo{person}{Chunyuan Li}, \bibinfo{person}{Qingyang Wu}, {and} \bibinfo{person}{Yong~Jae Lee}.} \bibinfo{year}{2023}\natexlab{}.
\newblock \showarticletitle{Visual Instruction Tuning}. In \bibinfo{booktitle}{\emph{Advances in Neural Information Processing Systems}}, \bibfield{editor}{\bibinfo{person}{A.~Oh}, \bibinfo{person}{T.~Naumann}, \bibinfo{person}{A.~Globerson}, \bibinfo{person}{K.~Saenko}, \bibinfo{person}{M.~Hardt}, {and} \bibinfo{person}{S.~Levine}} (Eds.), Vol.~\bibinfo{volume}{36}. \bibinfo{publisher}{Curran Associates, Inc.}, \bibinfo{pages}{34892--34916}.
\newblock
\urldef\tempurl%
\url{https://proceedings.neurips.cc/paper_files/paper/2023/file/6dcf277ea32ce3288914faf369fe6de0-Paper-Conference.pdf}
\showURL{%
\tempurl}


\bibitem[Liu et~al\mbox{.}(2017)]%
        {article}
\bibfield{author}{\bibinfo{person}{Kang Liu}, \bibinfo{person}{Song Gao}, \bibinfo{person}{Peiyuan Qiu}, \bibinfo{person}{Xiliang Liu}, \bibinfo{person}{Bo Yan}, {and} \bibinfo{person}{Feng Lu}.} \bibinfo{year}{2017}\natexlab{}.
\newblock \showarticletitle{Road2Vec: Measuring Traffic Interactions in Urban Road System from Massive Travel Routes}.
\newblock \bibinfo{journal}{\emph{International Journal of Geo-Information}}  \bibinfo{volume}{6} (\bibinfo{date}{10} \bibinfo{year}{2017}), \bibinfo{pages}{321}.
\newblock
\href{https://doi.org/10.3390/ijgi6110321}{doi:\nolinkurl{10.3390/ijgi6110321}}


\bibitem[Mahari et~al\mbox{.}(2023)]%
        {mahari-etal-2023-law}
\bibfield{author}{\bibinfo{person}{Robert Mahari}, \bibinfo{person}{Dominik Stammbach}, \bibinfo{person}{Elliott Ash}, {and} \bibinfo{person}{Alex Pentland}.} \bibinfo{year}{2023}\natexlab{}.
\newblock \showarticletitle{The Law and {NLP}: Bridging Disciplinary Disconnects}. In \bibinfo{booktitle}{\emph{Findings of the Association for Computational Linguistics: EMNLP 2023}}, \bibfield{editor}{\bibinfo{person}{Houda Bouamor}, \bibinfo{person}{Juan Pino}, {and} \bibinfo{person}{Kalika Bali}} (Eds.). \bibinfo{publisher}{Association for Computational Linguistics}, \bibinfo{address}{Singapore}, \bibinfo{pages}{3445--3454}.
\newblock
\href{https://doi.org/10.18653/v1/2023.findings-emnlp.224}{doi:\nolinkurl{10.18653/v1/2023.findings-emnlp.224}}


\bibitem[Mai et~al\mbox{.}(2024)]%
        {10.1145/3653070}
\bibfield{author}{\bibinfo{person}{Gengchen Mai}, \bibinfo{person}{Weiming Huang}, \bibinfo{person}{Jin Sun}, \bibinfo{person}{Suhang Song}, \bibinfo{person}{Deepak Mishra}, \bibinfo{person}{Ninghao Liu}, \bibinfo{person}{Song Gao}, \bibinfo{person}{Tianming Liu}, \bibinfo{person}{Gao Cong}, \bibinfo{person}{Yingjie Hu}, \bibinfo{person}{Chris Cundy}, \bibinfo{person}{Ziyuan Li}, \bibinfo{person}{Rui Zhu}, {and} \bibinfo{person}{Ni Lao}.} \bibinfo{year}{2024}\natexlab{}.
\newblock \showarticletitle{On the Opportunities and Challenges of Foundation Models for GeoAI (Vision Paper)}.
\newblock \bibinfo{journal}{\emph{ACM Trans. Spatial Algorithms Syst.}} \bibinfo{volume}{10}, \bibinfo{number}{2}, Article \bibinfo{articleno}{11} (\bibinfo{date}{July} \bibinfo{year}{2024}), \bibinfo{numpages}{46}~pages.
\newblock
\showISSN{2374-0353}
\href{https://doi.org/10.1145/3653070}{doi:\nolinkurl{10.1145/3653070}}


\bibitem[Mai et~al\mbox{.}(2025)]%
        {mai2025towards}
\bibfield{author}{\bibinfo{person}{Gengchen Mai}, \bibinfo{person}{Yiqun Xie}, \bibinfo{person}{Xiaowei Jia}, \bibinfo{person}{Ni Lao}, \bibinfo{person}{Jinmeng Rao}, \bibinfo{person}{Qing Zhu}, \bibinfo{person}{Zeping Liu}, \bibinfo{person}{Yao-Yi Chiang}, {and} \bibinfo{person}{Junfeng Jiao}.} \bibinfo{year}{2025}\natexlab{}.
\newblock \showarticletitle{Towards the next generation of Geospatial Artificial Intelligence}.
\newblock \bibinfo{journal}{\emph{International Journal of Applied Earth Observation and Geoinformation}}  \bibinfo{volume}{136} (\bibinfo{year}{2025}), \bibinfo{pages}{104368}.
\newblock


\bibitem[Mai et~al\mbox{.}(2020)]%
        {10.1007/978-3-030-14745-7_2}
\bibfield{author}{\bibinfo{person}{Gengchen Mai}, \bibinfo{person}{Bo Yan}, \bibinfo{person}{Krzysztof Janowicz}, {and} \bibinfo{person}{Rui Zhu}.} \bibinfo{year}{2020}\natexlab{}.
\newblock \showarticletitle{Relaxing Unanswerable Geographic Questions Using A Spatially Explicit Knowledge Graph Embedding Model}. In \bibinfo{booktitle}{\emph{Geospatial Technologies for Local and Regional Development}}, \bibfield{editor}{\bibinfo{person}{Phaedon Kyriakidis}, \bibinfo{person}{Diofantos Hadjimitsis}, \bibinfo{person}{Dimitrios Skarlatos}, {and} \bibinfo{person}{Ali Mansourian}} (Eds.). \bibinfo{publisher}{Springer International Publishing}, \bibinfo{address}{Cham}, \bibinfo{pages}{21--39}.
\newblock
\showISBNx{978-3-030-14745-7}


\bibitem[Manvi et~al\mbox{.}(2024)]%
        {10.5555/3692070.3693479}
\bibfield{author}{\bibinfo{person}{Rohin Manvi}, \bibinfo{person}{Samar Khanna}, \bibinfo{person}{Marshall Burke}, \bibinfo{person}{David Lobell}, {and} \bibinfo{person}{Stefano Ermon}.} \bibinfo{year}{2024}\natexlab{}.
\newblock \showarticletitle{Large language models are geographically biased}. In \bibinfo{booktitle}{\emph{Proceedings of the 41st International Conference on Machine Learning}} (Vienna, Austria) \emph{(\bibinfo{series}{ICML'24})}. \bibinfo{publisher}{JMLR.org}, Article \bibinfo{articleno}{1409}, \bibinfo{numpages}{16}~pages.
\newblock


\bibitem[Mo et~al\mbox{.}(2024)]%
        {mo2024survey}
\bibfield{author}{\bibinfo{person}{Fengran Mo}, \bibinfo{person}{Kelong Mao}, \bibinfo{person}{Ziliang Zhao}, \bibinfo{person}{Hongjin Qian}, \bibinfo{person}{Haonan Chen}, \bibinfo{person}{Yiruo Cheng}, \bibinfo{person}{Xiaoxi Li}, \bibinfo{person}{Yutao Zhu}, \bibinfo{person}{Zhicheng Dou}, {and} \bibinfo{person}{Jian-Yun Nie}.} \bibinfo{year}{2024}\natexlab{}.
\newblock \showarticletitle{A survey of conversational search}.
\newblock \bibinfo{journal}{\emph{arXiv preprint arXiv:2410.15576}} (\bibinfo{year}{2024}).
\newblock


\bibitem[Mo et~al\mbox{.}(2023a)]%
        {mo2023convgqr}
\bibfield{author}{\bibinfo{person}{Fengran Mo}, \bibinfo{person}{Kelong Mao}, \bibinfo{person}{Yutao Zhu}, \bibinfo{person}{Yihong Wu}, \bibinfo{person}{Kaiyu Huang}, {and} \bibinfo{person}{Jian-Yun Nie}.} \bibinfo{year}{2023}\natexlab{a}.
\newblock \showarticletitle{ConvGQR: Generative Query Reformulation for Conversational Search}. In \bibinfo{booktitle}{\emph{Proceedings of the 61st Annual Meeting of the Association for Computational Linguistics (Volume 1: Long Papers)}}. \bibinfo{pages}{4998--5012}.
\newblock


\bibitem[Mo et~al\mbox{.}(2023b)]%
        {mo2023learning}
\bibfield{author}{\bibinfo{person}{Fengran Mo}, \bibinfo{person}{Jian-Yun Nie}, \bibinfo{person}{Kaiyu Huang}, \bibinfo{person}{Kelong Mao}, \bibinfo{person}{Yutao Zhu}, \bibinfo{person}{Peng Li}, {and} \bibinfo{person}{Yang Liu}.} \bibinfo{year}{2023}\natexlab{b}.
\newblock \showarticletitle{Learning to relate to previous turns in conversational search}. In \bibinfo{booktitle}{\emph{Proceedings of the 29th ACM SIGKDD Conference on Knowledge Discovery and Data Mining}}. \bibinfo{pages}{1722--1732}.
\newblock


\bibitem[OpenAI et~al\mbox{.}(2024)]%
        {openai2024openaio1card}
\bibfield{author}{\bibinfo{person}{OpenAI}, \bibinfo{person}{:}, \bibinfo{person}{Aaron Jaech}, \bibinfo{person}{Adam Kalai}, \bibinfo{person}{Adam Lerer}, \bibinfo{person}{Adam Richardson}, \bibinfo{person}{Ahmed El-Kishky}, \bibinfo{person}{Aiden Low}, \bibinfo{person}{Alec Helyar}, \bibinfo{person}{Aleksander Madry}, \bibinfo{person}{Alex Beutel}, \bibinfo{person}{Alex Carney}, \bibinfo{person}{Alex Iftimie}, \bibinfo{person}{Alex Karpenko}, \bibinfo{person}{Alex~Tachard Passos}, \bibinfo{person}{Alexander Neitz}, \bibinfo{person}{Alexander Prokofiev}, \bibinfo{person}{Alexander Wei}, \bibinfo{person}{Allison Tam}, \bibinfo{person}{Ally Bennett}, \bibinfo{person}{Ananya Kumar}, \bibinfo{person}{Andre Saraiva}, \bibinfo{person}{Andrea Vallone}, \bibinfo{person}{Andrew Duberstein}, \bibinfo{person}{Andrew Kondrich}, \bibinfo{person}{Andrey Mishchenko}, \bibinfo{person}{Andy Applebaum}, \bibinfo{person}{Angela Jiang}, \bibinfo{person}{Ashvin Nair}, \bibinfo{person}{Barret Zoph}, \bibinfo{person}{Behrooz
  Ghorbani}, \bibinfo{person}{Ben Rossen}, \bibinfo{person}{Benjamin Sokolowsky}, \bibinfo{person}{Boaz Barak}, \bibinfo{person}{Bob McGrew}, \bibinfo{person}{Borys Minaiev}, \bibinfo{person}{Botao Hao}, \bibinfo{person}{Bowen Baker}, \bibinfo{person}{Brandon Houghton}, \bibinfo{person}{Brandon McKinzie}, \bibinfo{person}{Brydon Eastman}, \bibinfo{person}{Camillo Lugaresi}, \bibinfo{person}{Cary Bassin}, \bibinfo{person}{Cary Hudson}, \bibinfo{person}{Chak~Ming Li}, \bibinfo{person}{Charles de Bourcy}, \bibinfo{person}{Chelsea Voss}, \bibinfo{person}{Chen Shen}, \bibinfo{person}{Chong Zhang}, \bibinfo{person}{Chris Koch}, \bibinfo{person}{Chris Orsinger}, \bibinfo{person}{Christopher Hesse}, \bibinfo{person}{Claudia Fischer}, \bibinfo{person}{Clive Chan}, \bibinfo{person}{Dan Roberts}, \bibinfo{person}{Daniel Kappler}, \bibinfo{person}{Daniel Levy}, \bibinfo{person}{Daniel Selsam}, \bibinfo{person}{David Dohan}, \bibinfo{person}{David Farhi}, \bibinfo{person}{David Mely}, \bibinfo{person}{David Robinson},
  \bibinfo{person}{Dimitris Tsipras}, \bibinfo{person}{Doug Li}, \bibinfo{person}{Dragos Oprica}, \bibinfo{person}{Eben Freeman}, \bibinfo{person}{Eddie Zhang}, \bibinfo{person}{Edmund Wong}, \bibinfo{person}{Elizabeth Proehl}, \bibinfo{person}{Enoch Cheung}, \bibinfo{person}{Eric Mitchell}, \bibinfo{person}{Eric Wallace}, \bibinfo{person}{Erik Ritter}, \bibinfo{person}{Evan Mays}, \bibinfo{person}{Fan Wang}, \bibinfo{person}{Felipe~Petroski Such}, \bibinfo{person}{Filippo Raso}, \bibinfo{person}{Florencia Leoni}, \bibinfo{person}{Foivos Tsimpourlas}, \bibinfo{person}{Francis Song}, \bibinfo{person}{Fred von Lohmann}, \bibinfo{person}{Freddie Sulit}, \bibinfo{person}{Geoff Salmon}, \bibinfo{person}{Giambattista Parascandolo}, \bibinfo{person}{Gildas Chabot}, \bibinfo{person}{Grace Zhao}, \bibinfo{person}{Greg Brockman}, \bibinfo{person}{Guillaume Leclerc}, \bibinfo{person}{Hadi Salman}, \bibinfo{person}{Haiming Bao}, \bibinfo{person}{Hao Sheng}, \bibinfo{person}{Hart Andrin}, \bibinfo{person}{Hessam
  Bagherinezhad}, \bibinfo{person}{Hongyu Ren}, \bibinfo{person}{Hunter Lightman}, \bibinfo{person}{Hyung~Won Chung}, \bibinfo{person}{Ian Kivlichan}, \bibinfo{person}{Ian O'Connell}, \bibinfo{person}{Ian Osband}, \bibinfo{person}{Ignasi~Clavera Gilaberte}, \bibinfo{person}{Ilge Akkaya}, \bibinfo{person}{Ilya Kostrikov}, \bibinfo{person}{Ilya Sutskever}, \bibinfo{person}{Irina Kofman}, \bibinfo{person}{Jakub Pachocki}, \bibinfo{person}{James Lennon}, \bibinfo{person}{Jason Wei}, \bibinfo{person}{Jean Harb}, \bibinfo{person}{Jerry Twore}, \bibinfo{person}{Jiacheng Feng}, \bibinfo{person}{Jiahui Yu}, \bibinfo{person}{Jiayi Weng}, \bibinfo{person}{Jie Tang}, \bibinfo{person}{Jieqi Yu}, \bibinfo{person}{Joaquin~Quiñonero Candela}, \bibinfo{person}{Joe Palermo}, \bibinfo{person}{Joel Parish}, \bibinfo{person}{Johannes Heidecke}, \bibinfo{person}{John Hallman}, \bibinfo{person}{John Rizzo}, \bibinfo{person}{Jonathan Gordon}, \bibinfo{person}{Jonathan Uesato}, \bibinfo{person}{Jonathan Ward}, \bibinfo{person}{Joost
  Huizinga}, \bibinfo{person}{Julie Wang}, \bibinfo{person}{Kai Chen}, \bibinfo{person}{Kai Xiao}, \bibinfo{person}{Karan Singhal}, \bibinfo{person}{Karina Nguyen}, \bibinfo{person}{Karl Cobbe}, \bibinfo{person}{Katy Shi}, \bibinfo{person}{Kayla Wood}, \bibinfo{person}{Kendra Rimbach}, \bibinfo{person}{Keren Gu-Lemberg}, \bibinfo{person}{Kevin Liu}, \bibinfo{person}{Kevin Lu}, \bibinfo{person}{Kevin Stone}, \bibinfo{person}{Kevin Yu}, \bibinfo{person}{Lama Ahmad}, \bibinfo{person}{Lauren Yang}, \bibinfo{person}{Leo Liu}, \bibinfo{person}{Leon Maksin}, \bibinfo{person}{Leyton Ho}, \bibinfo{person}{Liam Fedus}, \bibinfo{person}{Lilian Weng}, \bibinfo{person}{Linden Li}, \bibinfo{person}{Lindsay McCallum}, \bibinfo{person}{Lindsey Held}, \bibinfo{person}{Lorenz Kuhn}, \bibinfo{person}{Lukas Kondraciuk}, \bibinfo{person}{Lukasz Kaiser}, \bibinfo{person}{Luke Metz}, \bibinfo{person}{Madelaine Boyd}, \bibinfo{person}{Maja Trebacz}, \bibinfo{person}{Manas Joglekar}, \bibinfo{person}{Mark Chen},
  \bibinfo{person}{Marko Tintor}, \bibinfo{person}{Mason Meyer}, \bibinfo{person}{Matt Jones}, \bibinfo{person}{Matt Kaufer}, \bibinfo{person}{Max Schwarzer}, \bibinfo{person}{Meghan Shah}, \bibinfo{person}{Mehmet Yatbaz}, \bibinfo{person}{Melody~Y. Guan}, \bibinfo{person}{Mengyuan Xu}, \bibinfo{person}{Mengyuan Yan}, \bibinfo{person}{Mia Glaese}, \bibinfo{person}{Mianna Chen}, \bibinfo{person}{Michael Lampe}, \bibinfo{person}{Michael Malek}, \bibinfo{person}{Michele Wang}, \bibinfo{person}{Michelle Fradin}, \bibinfo{person}{Mike McClay}, \bibinfo{person}{Mikhail Pavlov}, \bibinfo{person}{Miles Wang}, \bibinfo{person}{Mingxuan Wang}, \bibinfo{person}{Mira Murati}, \bibinfo{person}{Mo Bavarian}, \bibinfo{person}{Mostafa Rohaninejad}, \bibinfo{person}{Nat McAleese}, \bibinfo{person}{Neil Chowdhury}, \bibinfo{person}{Neil Chowdhury}, \bibinfo{person}{Nick Ryder}, \bibinfo{person}{Nikolas Tezak}, \bibinfo{person}{Noam Brown}, \bibinfo{person}{Ofir Nachum}, \bibinfo{person}{Oleg Boiko}, \bibinfo{person}{Oleg
  Murk}, \bibinfo{person}{Olivia Watkins}, \bibinfo{person}{Patrick Chao}, \bibinfo{person}{Paul Ashbourne}, \bibinfo{person}{Pavel Izmailov}, \bibinfo{person}{Peter Zhokhov}, \bibinfo{person}{Rachel Dias}, \bibinfo{person}{Rahul Arora}, \bibinfo{person}{Randall Lin}, \bibinfo{person}{Rapha~Gontijo Lopes}, \bibinfo{person}{Raz Gaon}, \bibinfo{person}{Reah Miyara}, \bibinfo{person}{Reimar Leike}, \bibinfo{person}{Renny Hwang}, \bibinfo{person}{Rhythm Garg}, \bibinfo{person}{Robin Brown}, \bibinfo{person}{Roshan James}, \bibinfo{person}{Rui Shu}, \bibinfo{person}{Ryan Cheu}, \bibinfo{person}{Ryan Greene}, \bibinfo{person}{Saachi Jain}, \bibinfo{person}{Sam Altman}, \bibinfo{person}{Sam Toizer}, \bibinfo{person}{Sam Toyer}, \bibinfo{person}{Samuel Miserendino}, \bibinfo{person}{Sandhini Agarwal}, \bibinfo{person}{Santiago Hernandez}, \bibinfo{person}{Sasha Baker}, \bibinfo{person}{Scott McKinney}, \bibinfo{person}{Scottie Yan}, \bibinfo{person}{Shengjia Zhao}, \bibinfo{person}{Shengli Hu},
  \bibinfo{person}{Shibani Santurkar}, \bibinfo{person}{Shraman~Ray Chaudhuri}, \bibinfo{person}{Shuyuan Zhang}, \bibinfo{person}{Siyuan Fu}, \bibinfo{person}{Spencer Papay}, \bibinfo{person}{Steph Lin}, \bibinfo{person}{Suchir Balaji}, \bibinfo{person}{Suvansh Sanjeev}, \bibinfo{person}{Szymon Sidor}, \bibinfo{person}{Tal Broda}, \bibinfo{person}{Aidan Clark}, \bibinfo{person}{Tao Wang}, \bibinfo{person}{Taylor Gordon}, \bibinfo{person}{Ted Sanders}, \bibinfo{person}{Tejal Patwardhan}, \bibinfo{person}{Thibault Sottiaux}, \bibinfo{person}{Thomas Degry}, \bibinfo{person}{Thomas Dimson}, \bibinfo{person}{Tianhao Zheng}, \bibinfo{person}{Timur Garipov}, \bibinfo{person}{Tom Stasi}, \bibinfo{person}{Trapit Bansal}, \bibinfo{person}{Trevor Creech}, \bibinfo{person}{Troy Peterson}, \bibinfo{person}{Tyna Eloundou}, \bibinfo{person}{Valerie Qi}, \bibinfo{person}{Vineet Kosaraju}, \bibinfo{person}{Vinnie Monaco}, \bibinfo{person}{Vitchyr Pong}, \bibinfo{person}{Vlad Fomenko}, \bibinfo{person}{Weiyi Zheng},
  \bibinfo{person}{Wenda Zhou}, \bibinfo{person}{Wes McCabe}, \bibinfo{person}{Wojciech Zaremba}, \bibinfo{person}{Yann Dubois}, \bibinfo{person}{Yinghai Lu}, \bibinfo{person}{Yining Chen}, \bibinfo{person}{Young Cha}, \bibinfo{person}{Yu Bai}, \bibinfo{person}{Yuchen He}, \bibinfo{person}{Yuchen Zhang}, \bibinfo{person}{Yunyun Wang}, \bibinfo{person}{Zheng Shao}, {and} \bibinfo{person}{Zhuohan Li}.} \bibinfo{year}{2024}\natexlab{}.
\newblock \bibinfo{title}{OpenAI o1 System Card}.
\newblock
\showeprint[arxiv]{2412.16720}~[cs.AI]
\urldef\tempurl%
\url{https://arxiv.org/abs/2412.16720}
\showURL{%
\tempurl}


\bibitem[Pang et~al\mbox{.}(2024)]%
        {pang2024vhmversatilehonestvision}
\bibfield{author}{\bibinfo{person}{Chao Pang}, \bibinfo{person}{Xingxing Weng}, \bibinfo{person}{Jiang Wu}, \bibinfo{person}{Jiayu Li}, \bibinfo{person}{Yi Liu}, \bibinfo{person}{Jiaxing Sun}, \bibinfo{person}{Weijia Li}, \bibinfo{person}{Shuai Wang}, \bibinfo{person}{Litong Feng}, \bibinfo{person}{Gui-Song Xia}, {and} \bibinfo{person}{Conghui He}.} \bibinfo{year}{2024}\natexlab{}.
\newblock \bibinfo{title}{VHM: Versatile and Honest Vision Language Model for Remote Sensing Image Analysis}.
\newblock
\showeprint[arxiv]{2403.20213}~[cs.CV]
\urldef\tempurl%
\url{https://arxiv.org/abs/2403.20213}
\showURL{%
\tempurl}


\bibitem[Pannone et~al\mbox{.}(2024)]%
        {Pannone2024}
\bibfield{author}{\bibinfo{person}{Andrew Pannone}, \bibinfo{person}{Aditya Raj}, \bibinfo{person}{Harikrishnan Ravichandran}, \bibinfo{person}{Sarbashis Das}, \bibinfo{person}{Ziheng Chen}, \bibinfo{person}{Collin~A. Price}, \bibinfo{person}{Mahmooda Sultana}, {and} \bibinfo{person}{Saptarshi Das}.} \bibinfo{year}{2024}\natexlab{}.
\newblock \showarticletitle{Robust chemical analysis with graphene chemosensors and machine learning}.
\newblock \bibinfo{journal}{\emph{Nature}} \bibinfo{volume}{634}, \bibinfo{number}{8034} (\bibinfo{date}{oct} \bibinfo{year}{2024}), \bibinfo{pages}{572--578}.
\newblock
\showISSN{1476-4687}
\href{https://doi.org/10.1038/s41586-024-08003-w}{doi:\nolinkurl{10.1038/s41586-024-08003-w}}


\bibitem[Qi et~al\mbox{.}(2011)]%
        {qi2011measuring}
\bibfield{author}{\bibinfo{person}{Guande Qi}, \bibinfo{person}{Xiaolong Li}, \bibinfo{person}{Shijian Li}, \bibinfo{person}{Gang Pan}, \bibinfo{person}{Zonghui Wang}, {and} \bibinfo{person}{Daqing Zhang}.} \bibinfo{year}{2011}\natexlab{}.
\newblock \showarticletitle{Measuring social functions of city regions from large-scale taxi behaviors}. In \bibinfo{booktitle}{\emph{2011 IEEE International Conference on Pervasive Computing and Communications Workshops (PERCOM Workshops)}}. IEEE, \bibinfo{pages}{384--388}.
\newblock


\bibitem[Rao et~al\mbox{.}(2023)]%
        {rao2023catsconditionaladversarialtrajectory}
\bibfield{author}{\bibinfo{person}{Jinmeng Rao}, \bibinfo{person}{Song Gao}, {and} \bibinfo{person}{Sijia Zhu}.} \bibinfo{year}{2023}\natexlab{}.
\newblock \bibinfo{title}{CATS: Conditional Adversarial Trajectory Synthesis for Privacy-Preserving Trajectory Data Publication Using Deep Learning Approaches}.
\newblock
\showeprint[arxiv]{2309.11587}~[cs.LG]
\urldef\tempurl%
\url{https://arxiv.org/abs/2309.11587}
\showURL{%
\tempurl}


\bibitem[Wang et~al\mbox{.}(2022)]%
        {wang2022advancing}
\bibfield{author}{\bibinfo{person}{Di Wang}, \bibinfo{person}{Qiming Zhang}, \bibinfo{person}{Yufei Xu}, \bibinfo{person}{Jing Zhang}, \bibinfo{person}{Bo Du}, \bibinfo{person}{Dacheng Tao}, {and} \bibinfo{person}{Liangpei Zhang}.} \bibinfo{year}{2022}\natexlab{}.
\newblock \showarticletitle{Advancing plain vision transformer toward remote sensing foundation model}.
\newblock \bibinfo{journal}{\emph{IEEE Transactions on Geoscience and Remote Sensing}}  \bibinfo{volume}{61} (\bibinfo{year}{2022}), \bibinfo{pages}{1--15}.
\newblock


\bibitem[Wang et~al\mbox{.}(2020)]%
        {wang2020neurotpr}
\bibfield{author}{\bibinfo{person}{Jimin Wang}, \bibinfo{person}{Yingjie Hu}, {and} \bibinfo{person}{Kenneth Joseph}.} \bibinfo{year}{2020}\natexlab{}.
\newblock \showarticletitle{NeuroTPR: A neuro-net toponym recognition model for extracting locations from social media messages}.
\newblock \bibinfo{journal}{\emph{Transactions in GIS}} \bibinfo{volume}{24}, \bibinfo{number}{3} (\bibinfo{year}{2020}), \bibinfo{pages}{719--735}.
\newblock


\bibitem[Wang et~al\mbox{.}(2024b)]%
        {wang2024user}
\bibfield{author}{\bibinfo{person}{Jiayin Wang}, \bibinfo{person}{Fengran Mo}, \bibinfo{person}{Weizhi Ma}, \bibinfo{person}{Peijie Sun}, \bibinfo{person}{Min Zhang}, {and} \bibinfo{person}{Jian-Yun Nie}.} \bibinfo{year}{2024}\natexlab{b}.
\newblock \showarticletitle{A User-Centric Multi-Intent Benchmark for Evaluating Large Language Models}. In \bibinfo{booktitle}{\emph{Proceedings of the 2024 Conference on Empirical Methods in Natural Language Processing}}. \bibinfo{pages}{3588--3612}.
\newblock


\bibitem[Wang et~al\mbox{.}(2024a)]%
        {Wang2024Qwen2VLEV}
\bibfield{author}{\bibinfo{person}{Peng Wang}, \bibinfo{person}{Shuai Bai}, \bibinfo{person}{Sinan Tan}, \bibinfo{person}{Shijie Wang}, \bibinfo{person}{Zhihao Fan}, \bibinfo{person}{Jinze Bai}, \bibinfo{person}{Ke-Yang Chen}, \bibinfo{person}{Xuejing Liu}, \bibinfo{person}{Jialin Wang}, \bibinfo{person}{Wenbin Ge}, \bibinfo{person}{Yang Fan}, \bibinfo{person}{Kai Dang}, \bibinfo{person}{Mengfei Du}, \bibinfo{person}{Xuancheng Ren}, \bibinfo{person}{Rui Men}, \bibinfo{person}{Dayiheng Liu}, \bibinfo{person}{Chang Zhou}, \bibinfo{person}{Jingren Zhou}, {and} \bibinfo{person}{Junyang Lin}.} \bibinfo{year}{2024}\natexlab{a}.
\newblock \showarticletitle{Qwen2-VL: Enhancing Vision-Language Model's Perception of the World at Any Resolution}.
\newblock \bibinfo{journal}{\emph{ArXiv}}  \bibinfo{volume}{abs/2409.12191} (\bibinfo{year}{2024}).
\newblock
\urldef\tempurl%
\url{https://api.semanticscholar.org/CorpusID:272704132}
\showURL{%
\tempurl}


\bibitem[Wu et~al\mbox{.}(2015)]%
        {Wu2015}
\bibfield{author}{\bibinfo{person}{Yu-Tzu Wu}, \bibinfo{person}{A.~Matthew Prina}, {and} \bibinfo{person}{Carol Brayne}.} \bibinfo{year}{2015}\natexlab{}.
\newblock \showarticletitle{The association between community environment and cognitive function: a systematic review}.
\newblock \bibinfo{journal}{\emph{Social Psychiatry and Psychiatric Epidemiology}} \bibinfo{volume}{50}, \bibinfo{number}{3} (\bibinfo{date}{mar} \bibinfo{year}{2015}), \bibinfo{pages}{351--362}.
\newblock
\showISSN{1433-9285}
\href{https://doi.org/10.1007/s00127-014-0945-6}{doi:\nolinkurl{10.1007/s00127-014-0945-6}}


\bibitem[Xia et~al\mbox{.}(2017)]%
        {7907303}
\bibfield{author}{\bibinfo{person}{Gui-Song Xia}, \bibinfo{person}{Jingwen Hu}, \bibinfo{person}{Fan Hu}, \bibinfo{person}{Baoguang Shi}, \bibinfo{person}{Xiang Bai}, \bibinfo{person}{Yanfei Zhong}, \bibinfo{person}{Liangpei Zhang}, {and} \bibinfo{person}{Xiaoqiang Lu}.} \bibinfo{year}{2017}\natexlab{}.
\newblock \showarticletitle{AID: A Benchmark Data Set for Performance Evaluation of Aerial Scene Classification}.
\newblock \bibinfo{journal}{\emph{IEEE Transactions on Geoscience and Remote Sensing}} \bibinfo{volume}{55}, \bibinfo{number}{7} (\bibinfo{year}{2017}), \bibinfo{pages}{3965--3981}.
\newblock
\href{https://doi.org/10.1109/TGRS.2017.2685945}{doi:\nolinkurl{10.1109/TGRS.2017.2685945}}


\bibitem[Xia et~al\mbox{.}(2024)]%
        {Xia_Sun_Wang_An_2024}
\bibfield{author}{\bibinfo{person}{Haochong Xia}, \bibinfo{person}{Shuo Sun}, \bibinfo{person}{Xinrun Wang}, {and} \bibinfo{person}{Bo An}.} \bibinfo{year}{2024}\natexlab{}.
\newblock \showarticletitle{Market-GAN: Adding Control to Financial Market Data Generation with Semantic Context}.
\newblock \bibinfo{journal}{\emph{Proceedings of the AAAI Conference on Artificial Intelligence}} \bibinfo{volume}{38}, \bibinfo{number}{14} (\bibinfo{date}{Mar.} \bibinfo{year}{2024}), \bibinfo{pages}{15996--16004}.
\newblock
\href{https://doi.org/10.1609/aaai.v38i14.29531}{doi:\nolinkurl{10.1609/aaai.v38i14.29531}}


\bibitem[Yang and Newsam(2010)]%
        {Nilsback08}
\bibfield{author}{\bibinfo{person}{Yi Yang} {and} \bibinfo{person}{Shawn Newsam}.} \bibinfo{year}{2010}\natexlab{}.
\newblock \showarticletitle{Bag-Of-Visual-Words and Spatial Extensions for Land-Use Classification}. In \bibinfo{booktitle}{\emph{ACM SIGSPATIAL International Conference on Advances in Geographic Information Systems (ACM GIS)}}.
\newblock


\bibitem[Yao et~al\mbox{.}(2016)]%
        {yao2016semantic}
\bibfield{author}{\bibinfo{person}{Xiwen Yao}, \bibinfo{person}{Junwei Han}, \bibinfo{person}{Gong Cheng}, \bibinfo{person}{Xueming Qian}, {and} \bibinfo{person}{Lei Guo}.} \bibinfo{year}{2016}\natexlab{}.
\newblock \showarticletitle{Semantic annotation of high-resolution satellite images via weakly supervised learning}.
\newblock \bibinfo{journal}{\emph{IEEE Transactions on Geoscience and Remote Sensing}} \bibinfo{volume}{54}, \bibinfo{number}{6} (\bibinfo{year}{2016}), \bibinfo{pages}{3660--3671}.
\newblock


\bibitem[Yao et~al\mbox{.}(2017)]%
        {yao2017sensing}
\bibfield{author}{\bibinfo{person}{Yao Yao}, \bibinfo{person}{Xia Li}, \bibinfo{person}{Xiaoping Liu}, \bibinfo{person}{Penghua Liu}, \bibinfo{person}{Zhaotang Liang}, \bibinfo{person}{Jinbao Zhang}, {and} \bibinfo{person}{Ke Mai}.} \bibinfo{year}{2017}\natexlab{}.
\newblock \showarticletitle{Sensing spatial distribution of urban land use by integrating points-of-interest and Google Word2Vec model}.
\newblock \bibinfo{journal}{\emph{International Journal of Geographical Information Science}} \bibinfo{volume}{31}, \bibinfo{number}{4} (\bibinfo{year}{2017}), \bibinfo{pages}{825--848}.
\newblock


\bibitem[Yao et~al\mbox{.}(2024)]%
        {yao2024minicpmvgpt4vlevelmllm}
\bibfield{author}{\bibinfo{person}{Yuan Yao}, \bibinfo{person}{Tianyu Yu}, \bibinfo{person}{Ao Zhang}, \bibinfo{person}{Chongyi Wang}, \bibinfo{person}{Junbo Cui}, \bibinfo{person}{Hongji Zhu}, \bibinfo{person}{Tianchi Cai}, \bibinfo{person}{Haoyu Li}, \bibinfo{person}{Weilin Zhao}, \bibinfo{person}{Zhihui He}, \bibinfo{person}{Qianyu Chen}, \bibinfo{person}{Huarong Zhou}, \bibinfo{person}{Zhensheng Zou}, \bibinfo{person}{Haoye Zhang}, \bibinfo{person}{Shengding Hu}, \bibinfo{person}{Zhi Zheng}, \bibinfo{person}{Jie Zhou}, \bibinfo{person}{Jie Cai}, \bibinfo{person}{Xu Han}, \bibinfo{person}{Guoyang Zeng}, \bibinfo{person}{Dahai Li}, \bibinfo{person}{Zhiyuan Liu}, {and} \bibinfo{person}{Maosong Sun}.} \bibinfo{year}{2024}\natexlab{}.
\newblock \bibinfo{title}{MiniCPM-V: A GPT-4V Level MLLM on Your Phone}.
\newblock
\showeprint[arxiv]{2408.01800}~[cs.CV]
\urldef\tempurl%
\url{https://arxiv.org/abs/2408.01800}
\showURL{%
\tempurl}


\bibitem[Yoo et~al\mbox{.}(2025)]%
        {Yoo2025}
\bibfield{author}{\bibinfo{person}{Seong-Keun Yoo}, \bibinfo{person}{Conall~W. Fitzgerald}, \bibinfo{person}{Byuri~Angela Cho}, \bibinfo{person}{Bailey~G. Fitzgerald}, \bibinfo{person}{Catherine Han}, \bibinfo{person}{Elizabeth~S. Koh}, \bibinfo{person}{Abhinav Pandey}, \bibinfo{person}{Hannah Sfreddo}, \bibinfo{person}{Fionnuala Crowley}, \bibinfo{person}{Michelle~Rudshteyn Korostin}, \bibinfo{person}{Neha Debnath}, \bibinfo{person}{Yan Leyfman}, \bibinfo{person}{Cristina Valero}, \bibinfo{person}{Mark Lee}, \bibinfo{person}{Joris~L. Vos}, \bibinfo{person}{Andrew~Sangho Lee}, \bibinfo{person}{Karena Zhao}, \bibinfo{person}{Stanley Lam}, \bibinfo{person}{Ezekiel Olumuyide}, \bibinfo{person}{Fengshen Kuo}, \bibinfo{person}{Eric~A. Wilson}, \bibinfo{person}{Pauline Hamon}, \bibinfo{person}{Clotilde Hennequin}, \bibinfo{person}{Miriam Saffern}, \bibinfo{person}{Lynda Vuong}, \bibinfo{person}{A.~Ari Hakimi}, \bibinfo{person}{Brian Brown}, \bibinfo{person}{Miriam Merad}, \bibinfo{person}{Sacha Gnjatic},
  \bibinfo{person}{Nina Bhardwaj}, \bibinfo{person}{Matthew~D. Galsky}, \bibinfo{person}{Eric~E. Schadt}, \bibinfo{person}{Robert~M. Samstein}, \bibinfo{person}{Thomas~U. Marron}, \bibinfo{person}{Mithat Gönen}, \bibinfo{person}{Luc G.~T. Morris}, {and} \bibinfo{person}{Diego Chowell}.} \bibinfo{year}{2025}\natexlab{}.
\newblock \showarticletitle{Prediction of checkpoint inhibitor immunotherapy efficacy for cancer using routine blood tests and clinical data}.
\newblock \bibinfo{journal}{\emph{Nature Medicine}} (\bibinfo{year}{2025}).
\newblock
\showISSN{1546-170X}
\href{https://doi.org/10.1038/s41591-024-03398-5}{doi:\nolinkurl{10.1038/s41591-024-03398-5}}
\newblock
\shownote{Published: 2025/01/06}.


\bibitem[Yuan and Raubal(2016)]%
        {yuan2016analyzing}
\bibfield{author}{\bibinfo{person}{Yihong Yuan} {and} \bibinfo{person}{Martin Raubal}.} \bibinfo{year}{2016}\natexlab{}.
\newblock \showarticletitle{Analyzing the distribution of human activity space from mobile phone usage: an individual and urban-oriented study}.
\newblock \bibinfo{journal}{\emph{International Journal of Geographical Information Science}} \bibinfo{volume}{30}, \bibinfo{number}{8} (\bibinfo{year}{2016}), \bibinfo{pages}{1594--1621}.
\newblock


\bibitem[Zhang et~al\mbox{.}(2021)]%
        {zhang2021perception}
\bibfield{author}{\bibinfo{person}{Fan Zhang}, \bibinfo{person}{Zhuangyuan Fan}, \bibinfo{person}{Yuhao Kang}, \bibinfo{person}{Yujie Hu}, {and} \bibinfo{person}{Carlo Ratti}.} \bibinfo{year}{2021}\natexlab{}.
\newblock \showarticletitle{“Perception bias”: Deciphering a mismatch between urban crime and perception of safety}.
\newblock \bibinfo{journal}{\emph{Landscape and Urban Planning}}  \bibinfo{volume}{207} (\bibinfo{year}{2021}), \bibinfo{pages}{104003}.
\newblock


\bibitem[Zhang et~al\mbox{.}(2018)]%
        {zhang2018measuring}
\bibfield{author}{\bibinfo{person}{Fan Zhang}, \bibinfo{person}{Bolei Zhou}, \bibinfo{person}{Liu Liu}, \bibinfo{person}{Yu Liu}, \bibinfo{person}{Helene~H Fung}, \bibinfo{person}{Hui Lin}, {and} \bibinfo{person}{Carlo Ratti}.} \bibinfo{year}{2018}\natexlab{}.
\newblock \showarticletitle{Measuring human perceptions of a large-scale urban region using machine learning}.
\newblock \bibinfo{journal}{\emph{Landscape and Urban Planning}}  \bibinfo{volume}{180} (\bibinfo{year}{2018}), \bibinfo{pages}{148--160}.
\newblock


\bibitem[Zhang et~al\mbox{.}(2024)]%
        {Zhang2024GeospatialAI}
\bibfield{author}{\bibinfo{person}{Ziwei Zhang}, \bibinfo{person}{Liang Wu}, \bibinfo{person}{Liufeng Tao}, \bibinfo{person}{Sheng Hu}, \bibinfo{person}{Hui Long}, \bibinfo{person}{Yongyang Xu}, \bibinfo{person}{Jinquan Li}, \bibinfo{person}{Jingjing Zhang}, \bibinfo{person}{Zhijun Zhou}, \bibinfo{person}{Jing Liu}, \bibinfo{person}{Cheng Cai}, \bibinfo{person}{Hong Zhang}, \bibinfo{person}{Dan Liu}, \bibinfo{person}{Yan Zeng}, {and} \bibinfo{person}{Wei Luo}.} \bibinfo{year}{2024}\natexlab{}.
\newblock \showarticletitle{Geospatial Applications in Alzheimer’s Disease Research and Beyond: A Systematic Review}.
\newblock \bibinfo{journal}{\emph{Annals of the American Association of Geographers}} (\bibinfo{year}{2024}).
\newblock
\urldef\tempurl%
\url{https://api.semanticscholar.org/CorpusID:271724904}
\showURL{%
\tempurl}


\bibitem[Zhao et~al\mbox{.}(2023)]%
        {2023_ceus_soundscapes}
\bibfield{author}{\bibinfo{person}{Tianhong Zhao}, \bibinfo{person}{Xiucheng Liang}, \bibinfo{person}{Wei Tu}, \bibinfo{person}{Zhengdong Huang}, {and} \bibinfo{person}{Filip Biljecki}.} \bibinfo{year}{2023}\natexlab{}.
\newblock \showarticletitle{Sensing urban soundscapes from street view imagery}.
\newblock \bibinfo{journal}{\emph{Computers, Environment and Urban Systems}}  \bibinfo{volume}{99} (\bibinfo{year}{2023}), \bibinfo{pages}{101915}.
\newblock
\href{https://doi.org/10.1016/j.compenvurbsys.2022.101915}{doi:\nolinkurl{10.1016/j.compenvurbsys.2022.101915}}


\bibitem[Zhu et~al\mbox{.}(2023)]%
        {zhu2023minigpt}
\bibfield{author}{\bibinfo{person}{Deyao Zhu}, \bibinfo{person}{Jun Chen}, \bibinfo{person}{Xiaoqian Shen}, \bibinfo{person}{Xiang Li}, {and} \bibinfo{person}{Mohamed Elhoseiny}.} \bibinfo{year}{2023}\natexlab{}.
\newblock \showarticletitle{MiniGPT-4: Enhancing Vision-Language Understanding with Advanced Large Language Models}.
\newblock \bibinfo{journal}{\emph{arXiv preprint arXiv:2304.10592}} (\bibinfo{year}{2023}).
\newblock


\end{thebibliography}

\clearpage
\appendix
\section{More Details of Experimental Setup}

\subsection{The Source of the Data and Model}

You can easily obtain the datasets or models involved in this paper from the following resource list:

\begin{enumerate}
  \item CDC Wonder: https://wonder.cdc.gov/ucd-icd10.html. The dementia mortality data are obtained from the US Centers for Disease Control and Prevention Wide-ranging Online Data for Epidemiologic Research (CDC WONDER).
  \item Gaode Maps: https://lbs.amap.com/. Obtain POI data of Beijing and Shenzhen via API.
  \item WorldView: https://worldview.earthdata.nasa.gov/.\\Through this website, you can obtain WorldView series remote sensing image data products of Beijing, Shenzhen and the United States.
  \item Google Maps : https://www.google.com/maps. The source of the street view images used in the article. Alternatively, you can directly use the Visual-Soundscapes and Global Streetscapes datasets.
  \item city government websites:https://opendata.sz.gov.cn/ and https://www.beijing.gov.cn/gongkai/guihua/wngh/cqgh/20\\1907/t20190701\_100008.html.
  \item LLaVA1.5-7B and Qwen2-VL-7B: The base model weights from the Huggingface website:https://huggingface.co/.
\end{enumerate}

\section{Additional Experimental Results}
\subsection{Results in Zero-Shot Settings}

As shown in Table~\ref{table:Zero-Shot}, we demonstrate the comparison of the generalization performance of OmniGeo and baseline models on five datasets in zero-shot settings.

\begin{table*}[!t]
  \begin{centering}
    \caption{Results of OmniGeo and the baseline models on  geospatial related tasks under the zero-shot setting.}
    \label{table:Zero-Shot}
    \vspace{-3ex}
    \resizebox{\textwidth}{!}{
  \begin{tabular}{lccccccccccccccc}
  \toprule
  {\multirow{3}{*}{Model}}  & \multicolumn{6}{c}{Semantic Analysis } & \multicolumn{3}{c}{Urban Geography} & \multicolumn{3}{c}{Urban Perception(Noise)} & \multicolumn{3}{c}{RS} \\
    \cmidrule(r){2-16}    
  & \multicolumn{3}{c}{NEEL} & \multicolumn{3}{c}{GeoCorpora} & \multicolumn{3}{c}{UG-Beijing} & \multicolumn{3}{c}{UP-Shenzhen} & \multicolumn{3}{c}{UC-Merced} \\
    \cmidrule(r){2-16}
    & P$\uparrow$ & R$\uparrow$ & F1-score$\uparrow$ & P$\uparrow$ & R$\uparrow$ & F1-score$\uparrow$ & P$\uparrow$ & R$\uparrow$ & F1$\uparrow$ & P$\uparrow$ & R$\uparrow$ & F1$\uparrow$ & P$\uparrow$ & R$\uparrow$ & F1$\uparrow$ \\
   \midrule
   Stanford NER & 0.7331 & 0.5184 & 0.6074 & 0.8118 & 0.5118 & 0.6278 & - & - & - & - & - & - & - & - & - \\
   spaCy NER & 0.5372 & 0.5488 & 0.5429 & 0.5562 & 0.5183 & 0.5366 & - & - & - & - & - & - & - & - & - \\
   \midrule
   NeuroTPR & 0.7326 & 0.7582 & 0.7452 & 0.8199 & 0.7214 & 0.7675 & - & - & - & - & - & - & - & - & - \\
   BERT & 0.5667 & 0.2213 & 0.3183 & 0.5292 & 0.3193 & 0.3983 & - & - & - & - & - & - & - & - & - \\
   \midrule
   BLIP-2 & 0.2000 & 0.0022 & 0.0043 & \textbf{0.8000} & 0.0014 & 0.0027 & 0.3790 & 0.1750 & 0.2395 & 0.3073 & 0.2724 & 0.2888 & 0.3088 & 0.2643 & 0.2848 \\
   GPT-4o & 0.6980 & \textbf{0.8221} & 0.7550 & 0.7875 & \textbf{0.8597} & \textbf{0.8220} & \textbf{0.3884} & \textbf{0.3910} & \textbf{0.3897} & 0.2858 & 0.3471 & 0.3135 & \textbf{0.6726} & \textbf{0.6405} & \textbf{0.6562} \\
   LLaVA1.5-7B & 0.4818 & 0.6876 & 0.5666 & 0.5711 & 0.6965 & 0.6276 & 0.3551 & 0.3643 & 0.3597 & 0.1704 & 0.4126 & 0.2412 & 0.4455 & 0.2738 & 0.3391 \\
   Qwen2-VL-7B & 0.5954 & 0.4534 & 0.5148 & 0.2999 & 0.0681 & 0.1110 & 0.3025 & 0.1777 & 0.2239 & 0.1415 & 0.2951 & 0.1913 & 0.5047 & 0.2381 & 0.3235 \\
    \midrule
   OmniGeo~(LLaVA) & \textbf{0.7919} & 0.8039 & \textbf{0.7585} & 0.7791 & 0.7751 & 0.7771 & 0.3756 & 0.3378 & 0.3557 & 0.4640 & \textbf{0.4726} & \textbf{0.4683} & 0.6149 & 0.5905 & 0.6025 \\
   OmniGeo~(Qwen2) & 0.7177 & 0.7115 & 0.7146 & 0.7830 & 0.6636 & 0.7184 & 0.3666 & 0.2807 & 0.3180 & \textbf{0.5513} & 0.3858 & 0.4540 & 0.3837 & 0.2500 & 0.3072 \\
  \bottomrule
  \end{tabular}}
  \end{centering}
\end{table*}

\textbf{Geospatial Semantics} task performance is shown in Table~\ref{table:Zero-Shot} where OmniGeo achieved competitive performance with GPT-4o on both the NEEL and GeoCorpora datasets.

\textbf{Urban Geography} task results are reported in Table ~\ref{table:Zero-Shot}, GPT-4o performs the best, achieving optimal results on all three metrics.OmniGeo~(LLaVA) achieved results that are competitive with GPT-4o. It is worth noting that LLaVA1.5 performs slightly better than OmniGeo~(LLaVA) on the Beijing dataset, which is understandable for two reasons. First, OmniGeo~(LLaVA) acquired substantial knowledge about the urban spatial structure of Shenzhen during fine-tuning, and transferring this knowledge directly to Beijing is challenging due to the substantial differences in urban planning and structure between Shenzhen and Beijing. Second, the Beijing dataset contains a large amount of noise. Compared to Shenzhen, which is predominantly industrial, Beijing has many regions with small sizes and unclear land-use types, or mixed land-use types.

\textbf{Urban Perception} task comparison is
presented in Table~\ref{table:Zero-Shot}. OmniGeo~(LLaVA) achieved the best performance on the Shenzhen dataset, with a particularly higher Weighted-F1 score of 0.1548 compared to GPT-4o. This suggests that OmniGeo~(LLaVA) has the capability to transfer advanced perceptual knowledge across cities for urban perception tasks, which is precisely what we aimed for.

\textbf{Remote Sensing} task results are shown in Table~\ref{table:Zero-Shot}. GPT-4o maintains a relatively high classification ability on the UC-Merced dataset, while OmniGeo~(LLaVA) consistently demonstrates competitive RS image classification ability with GPT-4o. This indicates that OmniGeo~(LLaVA) correctly understands the semantics of each scene and transfers geographical spatial knowledge across RS images with varying spatial and spectral resolutions.

\subsection{Discussion}

\begin{figure*}[!t]
  \centering
  \subfloat[ARIMA]
  {\includegraphics[width=0.4\textwidth]{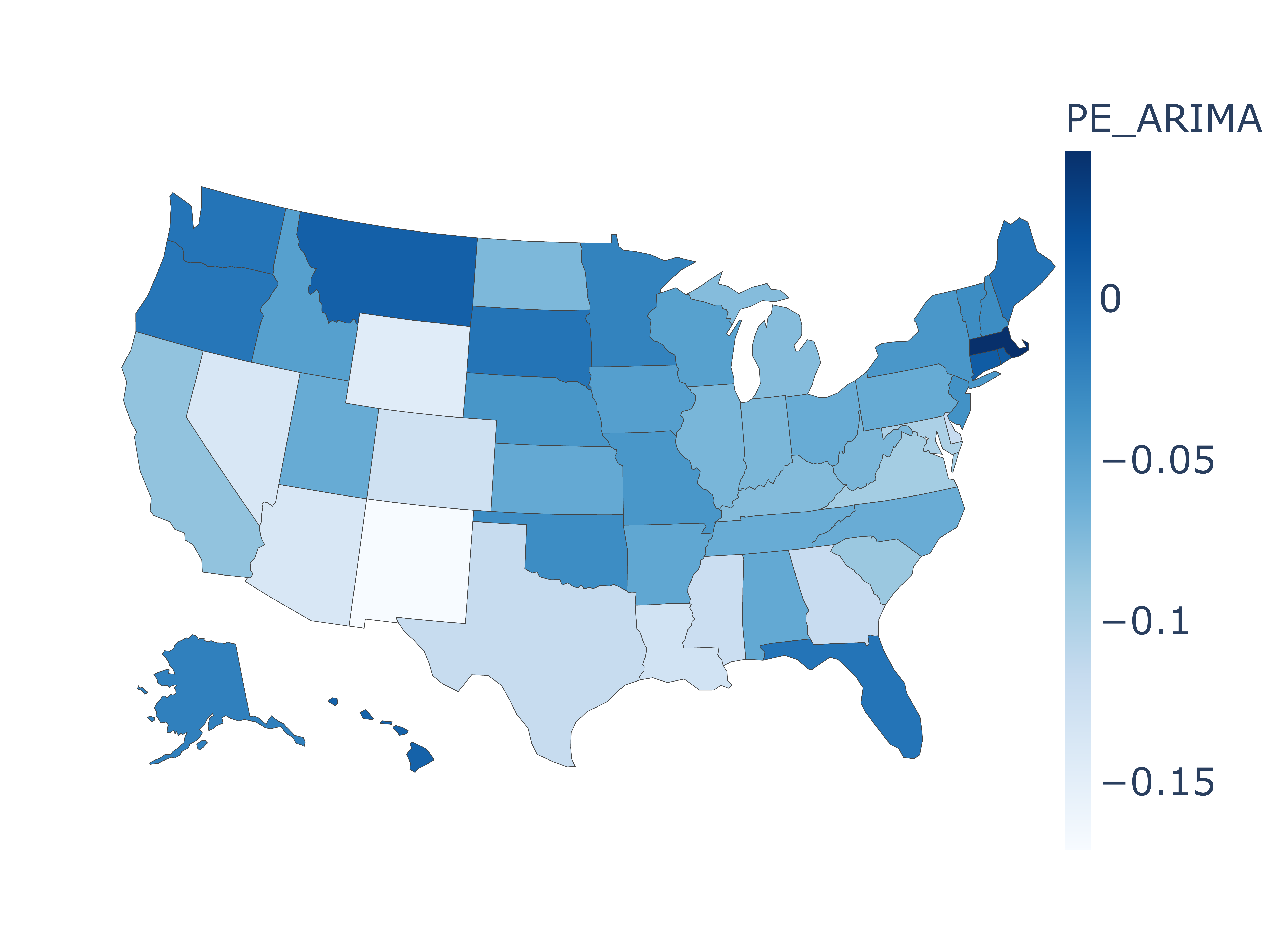}\label{fig:PE_1}}
  \hspace{3em} 
  \subfloat[GPT-4o]
  {\includegraphics[width=0.4\textwidth]{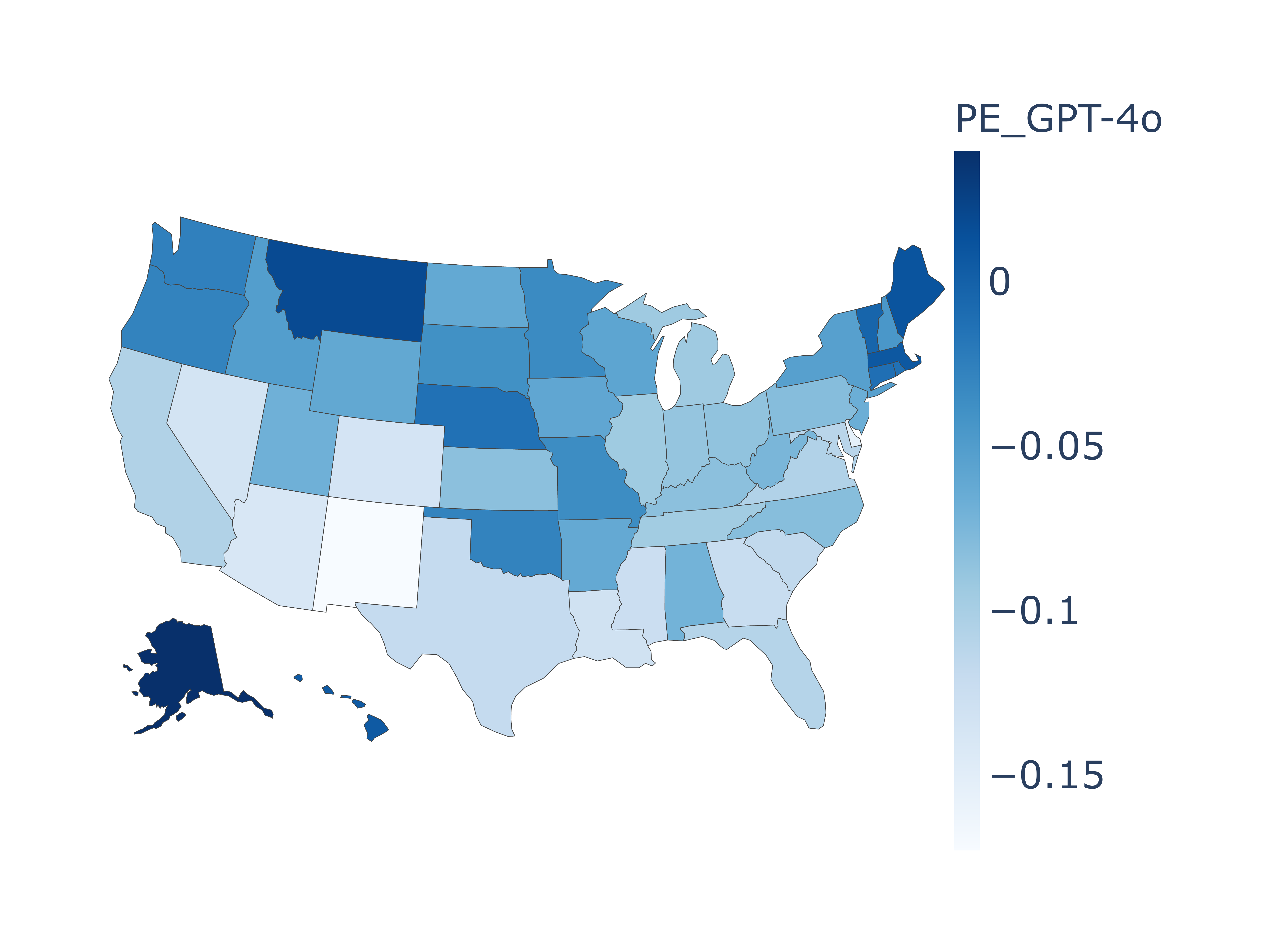}\label{fig:PE_2}}
  \hspace{3em} 
  \subfloat[LLaVA1.5]
  {\includegraphics[width=0.4\textwidth]{PE_LLaVA1.5.pdf}\label{fig:PE_3}}
  \hspace{3em} 
  \subfloat[OmniGeo~(LLaVA)]
  {\includegraphics[width=0.4\textwidth]{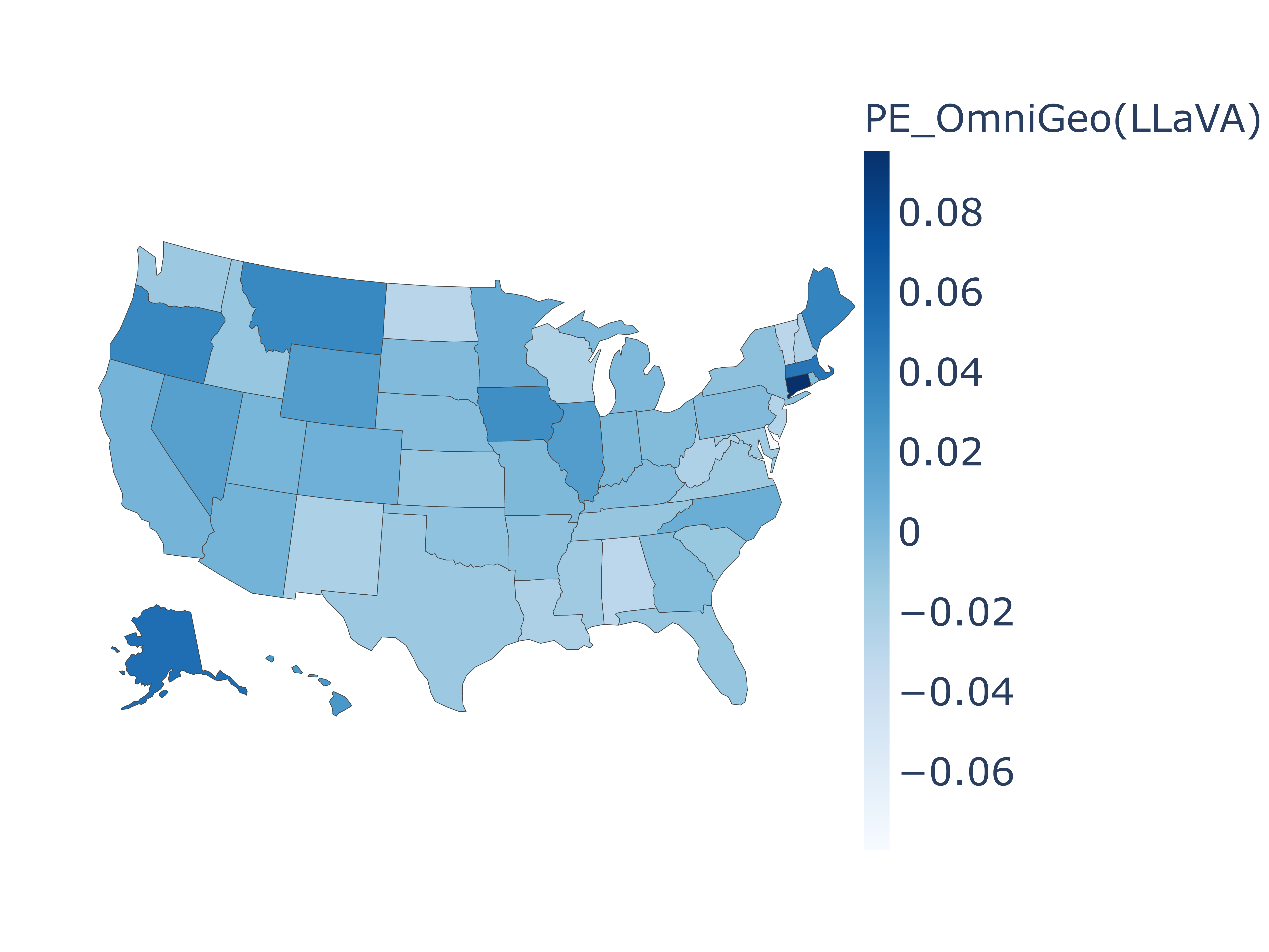}\label{fig:PE_4}}

  \caption{Prediction error plot for each baseline and OmniGeo~(LLaVA) on the dementia deaths time series forecasting task. The color of each US region indicates the percentage error of each model in predicting that state $PE = (Prediction - True)/True$.}
  \label{fig:PE}
\end{figure*}

For the dementia death counts time series fore-casting, Figure~\ref{fig:PE} visualizes the prediction capabilities of various baselines and OmniGeo~(LLaVA) on the US state-level dataset based on PE values. Significant geographic distribution differences in prediction abilities exist across different states, possibly due to "geographical bias". Based on the position 0 in the legend, it can be seen that ARIMA, GPT-4o, and LLaVA1.5 tend to overestimate the death count across the entire dataset. Regarding the degree of overestimation, LLaVA1.5 tends to overestimate more significantly, while ARIMA and GPT-4o show similar behaviors, with overestimation tendencies in Florida and Maine, respectively. In contrast, the baseline models severely underestimate the death toll in New Mexico, while OmniGeo~(LLaVA) shows balanced predictive performance. The results at the US country-level are similar, but Qwen2-VL performs poorly, which may be attributed to its poor instruction-following ability in this task and its lack of relevant geographical knowledge.

\begin{figure*}[!t]
  \centering
  
  \subfloat[Place2Vec]
  {\includegraphics[width=0.45\textwidth]{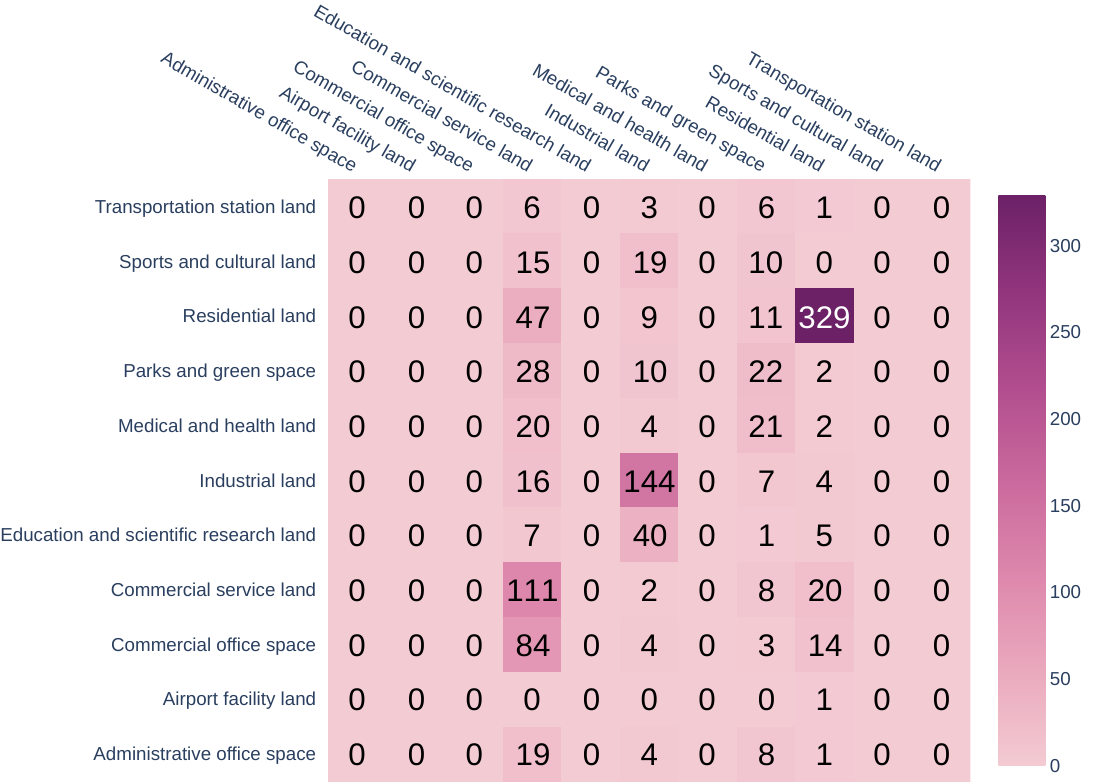}\label{fig:urban region function classification_1}}
  \quad
  \subfloat[HGI]
  {\includegraphics[width=0.45\textwidth]{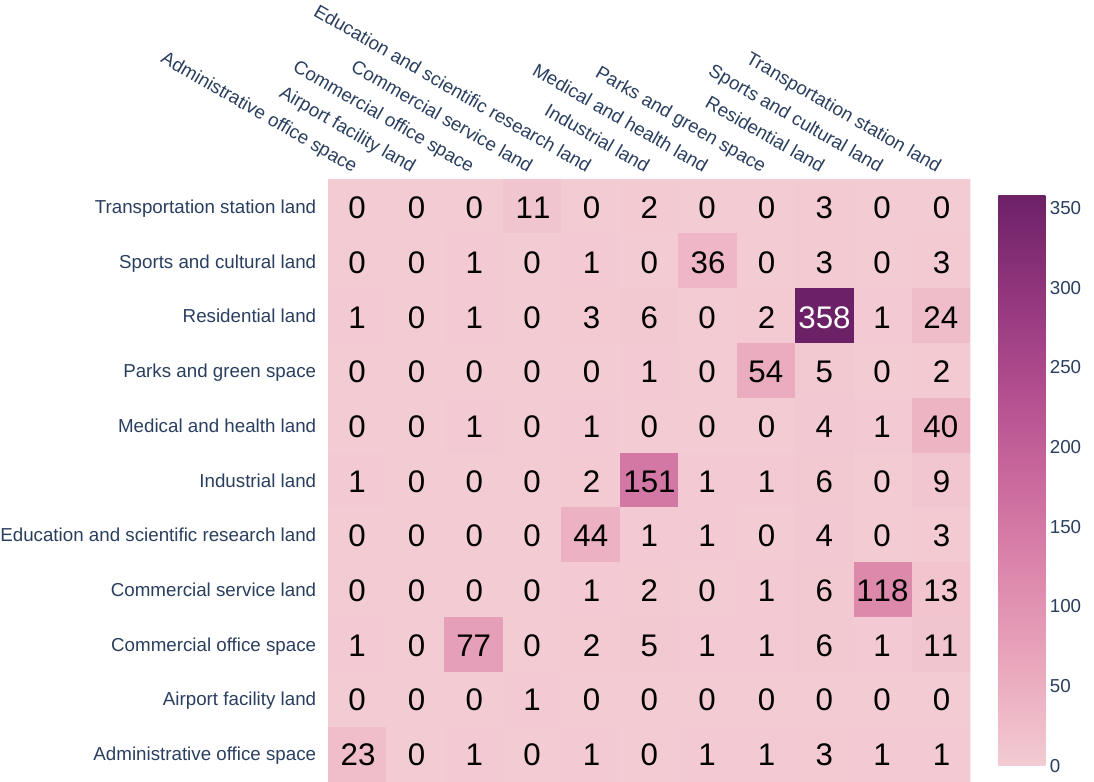}\label{fig:urban region function classification_2}}

  \subfloat[Blip2]
  {\includegraphics[width=0.3\textwidth]{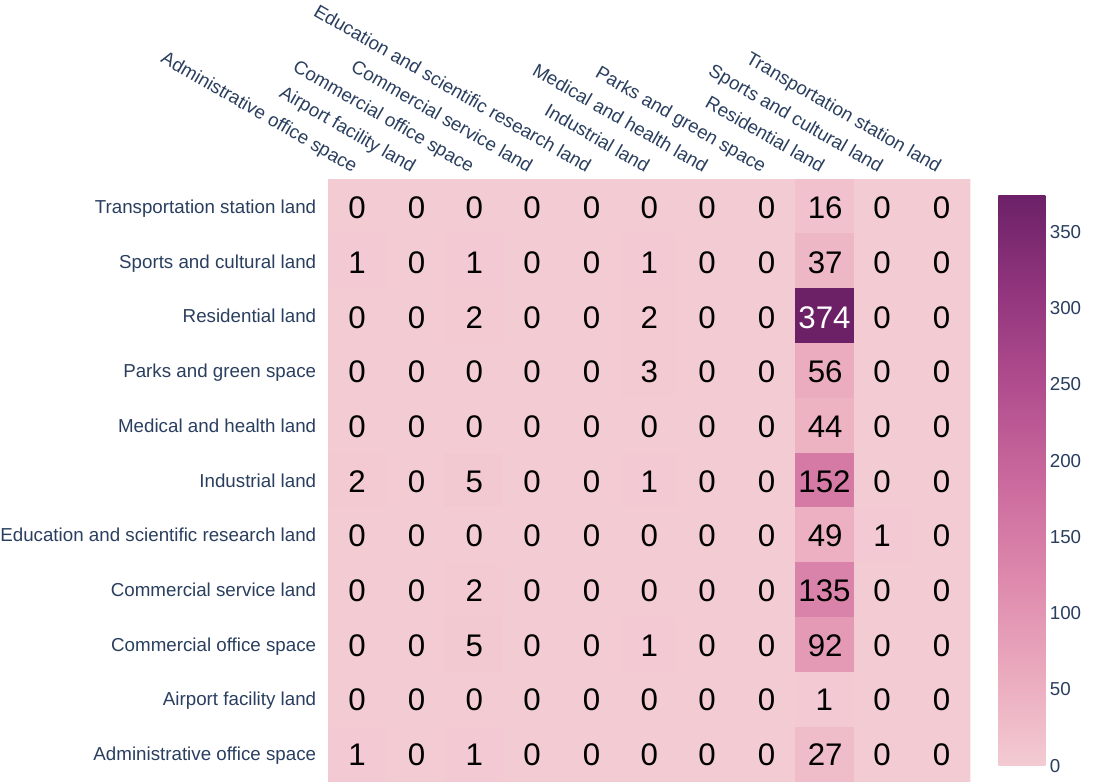}\label{fig:urban region function classification_3}}
  \quad
  \subfloat[GPT-4o]
  {\includegraphics[width=0.3\textwidth]{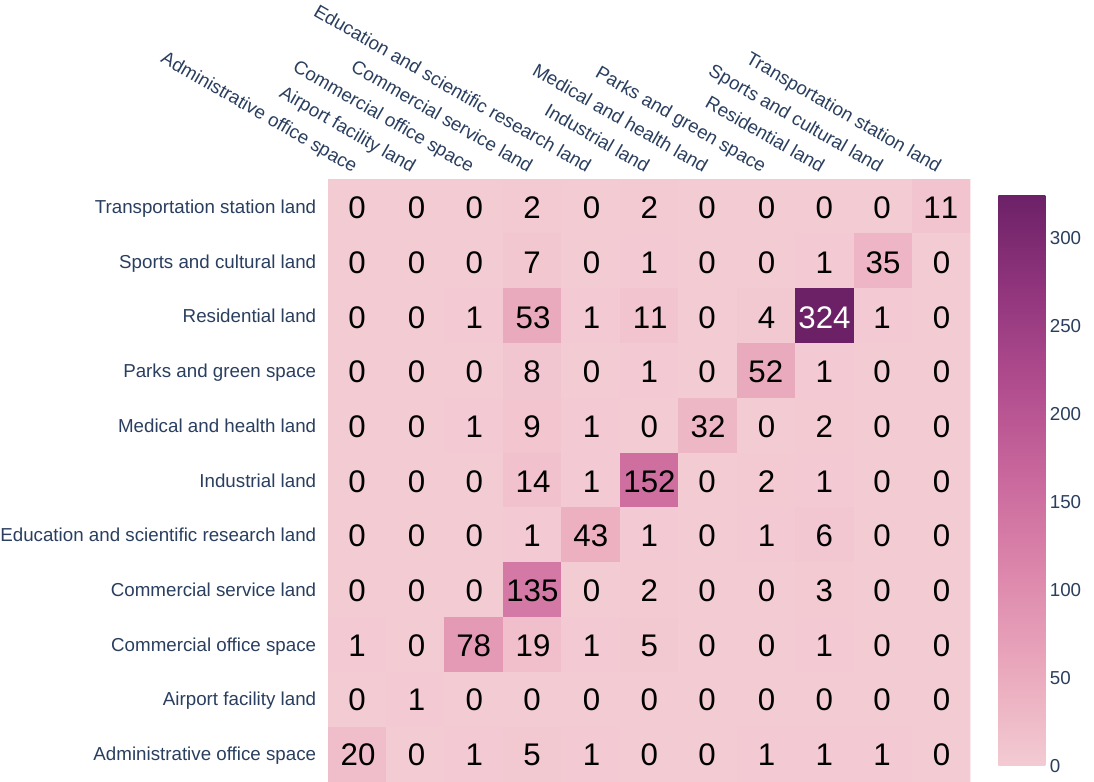}\label{fig:urban region function classification_4}}
  \quad
  \subfloat[OmniGeo~(LLaVA)]
  {\includegraphics[width=0.3\textwidth]{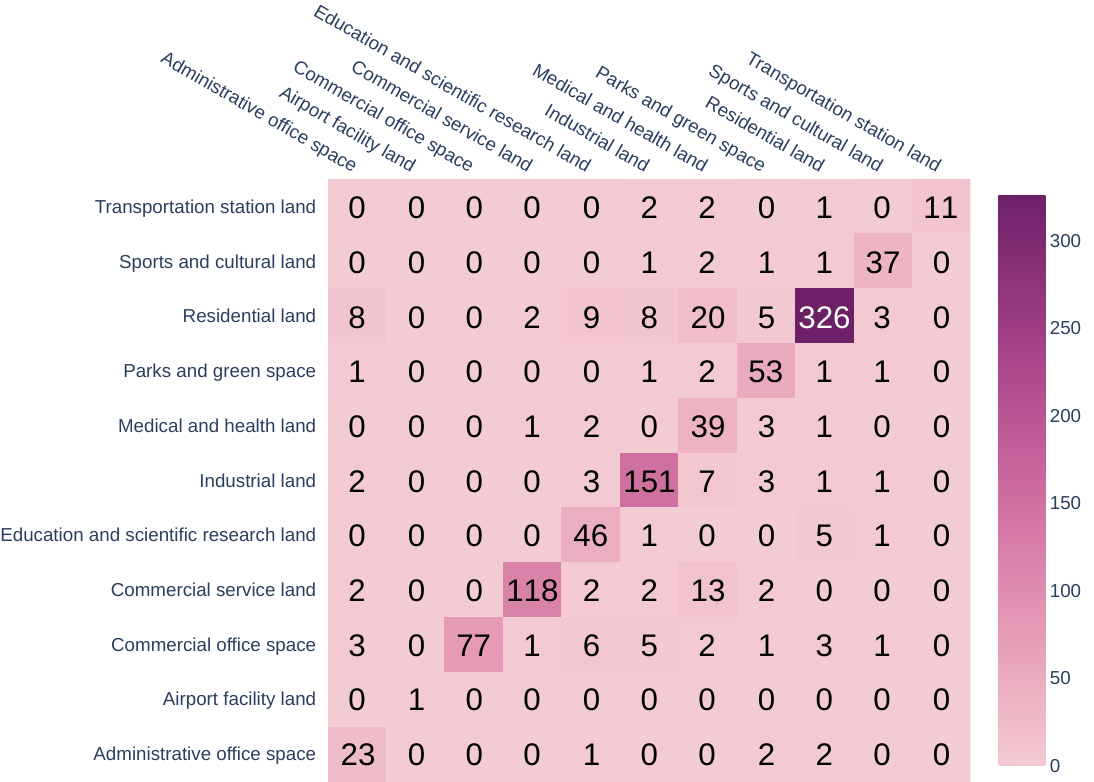}\label{fig:urban region function classification_5}}

  \caption{Confusion matrix comparison of Place2Vec, HGI, Blip2, GPT-4o and OmniGeo~(LLaVA) models on the Shenzhen urban region function classification dataset}
  \label{fig:urban region function classification}
\end{figure*}

For the urban region function classification, The embedding models Place2Vec and HGI exhibit different confusion patterns: Place2Vec tends to predict "Commercial office space" as "Commercial service land," while HGI tends to predict "Commercial service land" as "Sports and cultural land". Blip2 failed in this task, as it predicted nearly all regional functional types as "Residential land," while the GPT-4o model performed well. In contrast, OmniGeo (LLaVA) demonstrated the most stable performance. The above results suggest that "Commercial service land," "Residential land," "Residential land," and "Commercial office space" are all challenging samples to classify, and different models exhibit varying confusion patterns across different land-use types, which may be related to the models' geographic knowledge capacity and the difficulty of transferring ontology knowledge into the GeoAI domain.

\begin{figure*}[!t]
  \centering
  
  \subfloat[AlexNet]
  {\includegraphics[width=0.23\textwidth]{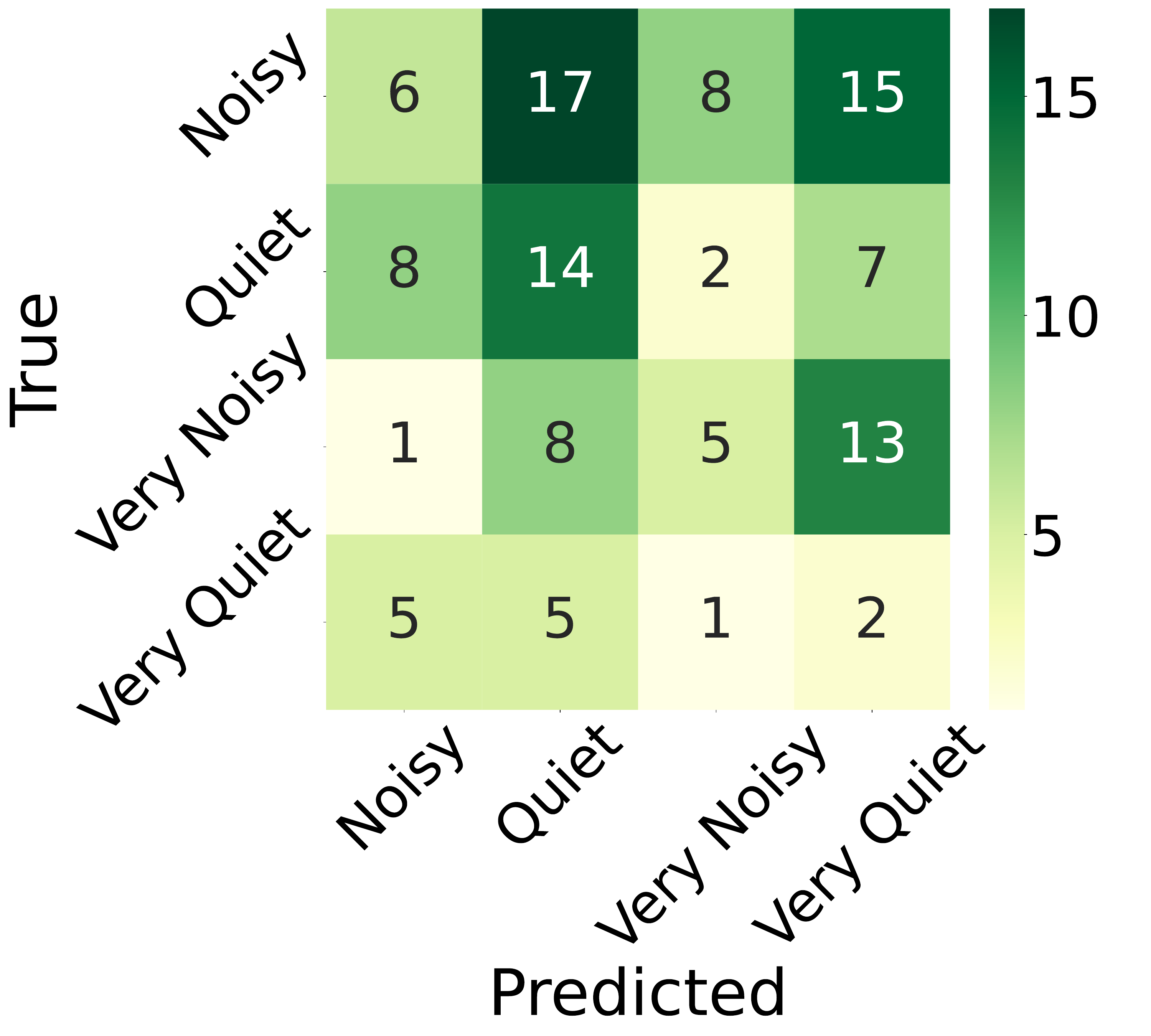}\label{fig:predicting Noise and Wealthy_1}}
  \quad
  \subfloat[GPT-4o]
  {\includegraphics[width=0.23\textwidth]{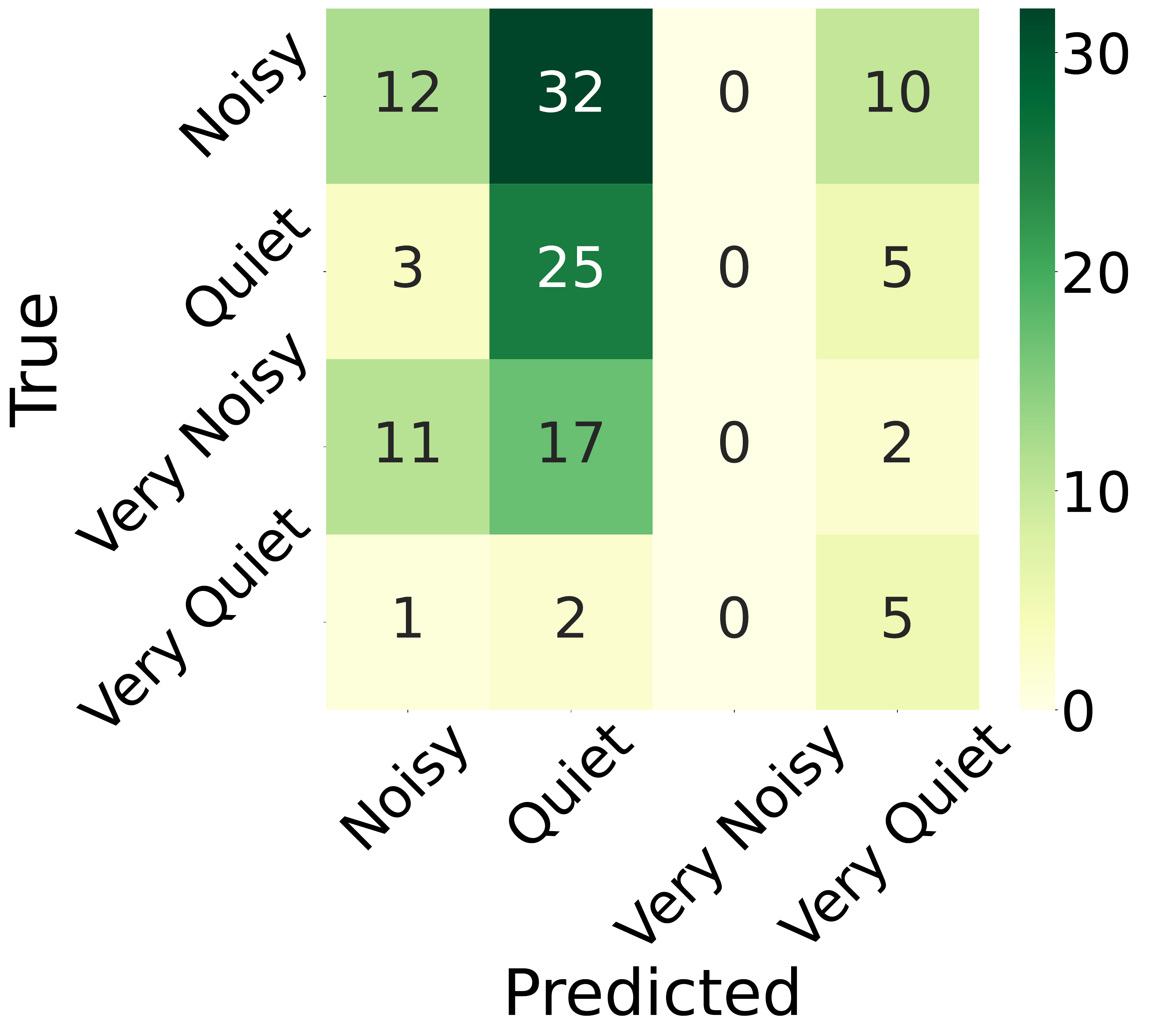}\label{fig:predicting Noise and Wealthy_2}}
  \quad
  \subfloat[InternVL2]
  {\includegraphics[width=0.23\textwidth]{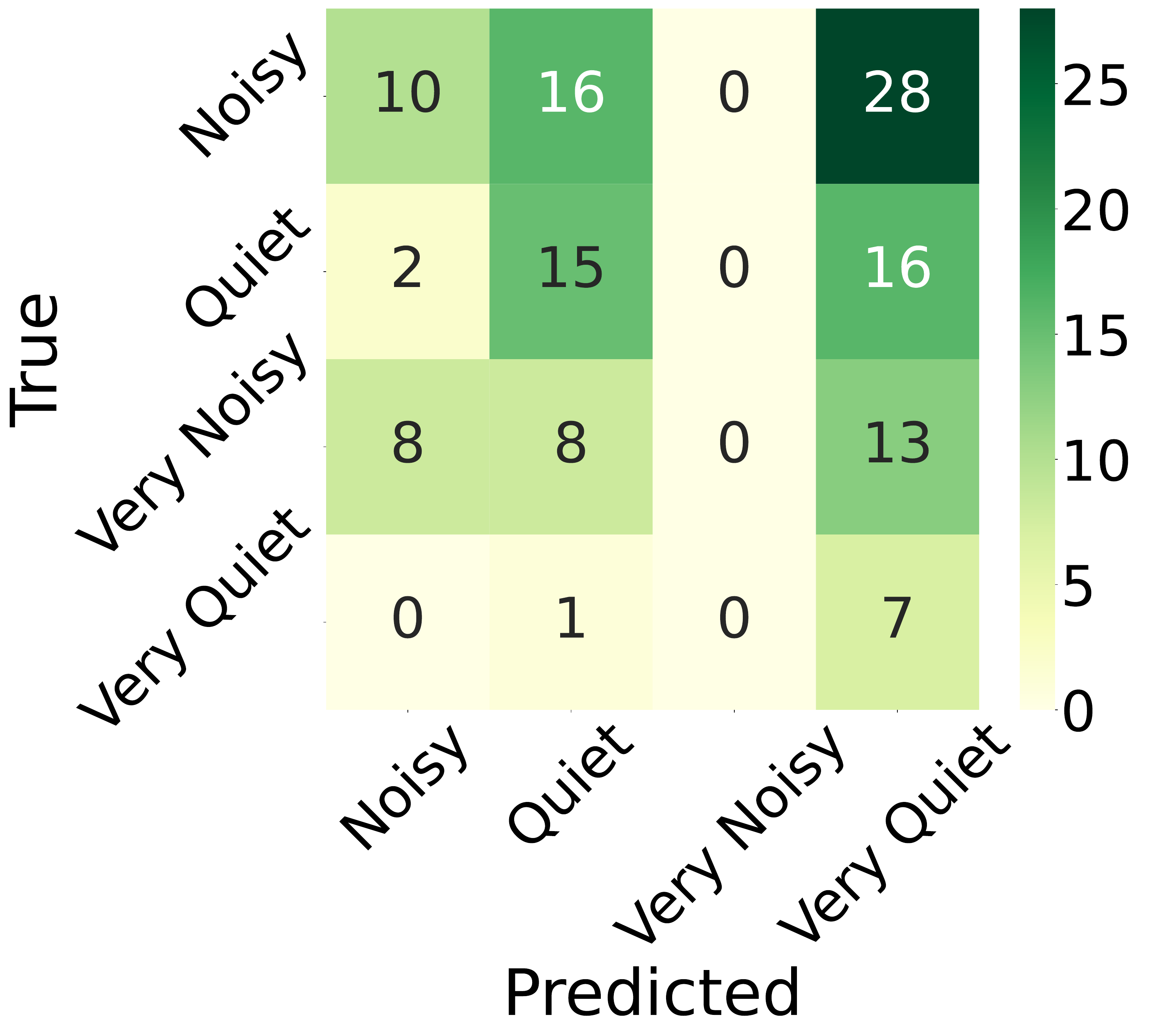}\label{fig:predicting Noise and Wealthy_3}}
  \quad
  \subfloat[OmniGeo~(LLaVA)]
  {\includegraphics[width=0.23\textwidth]{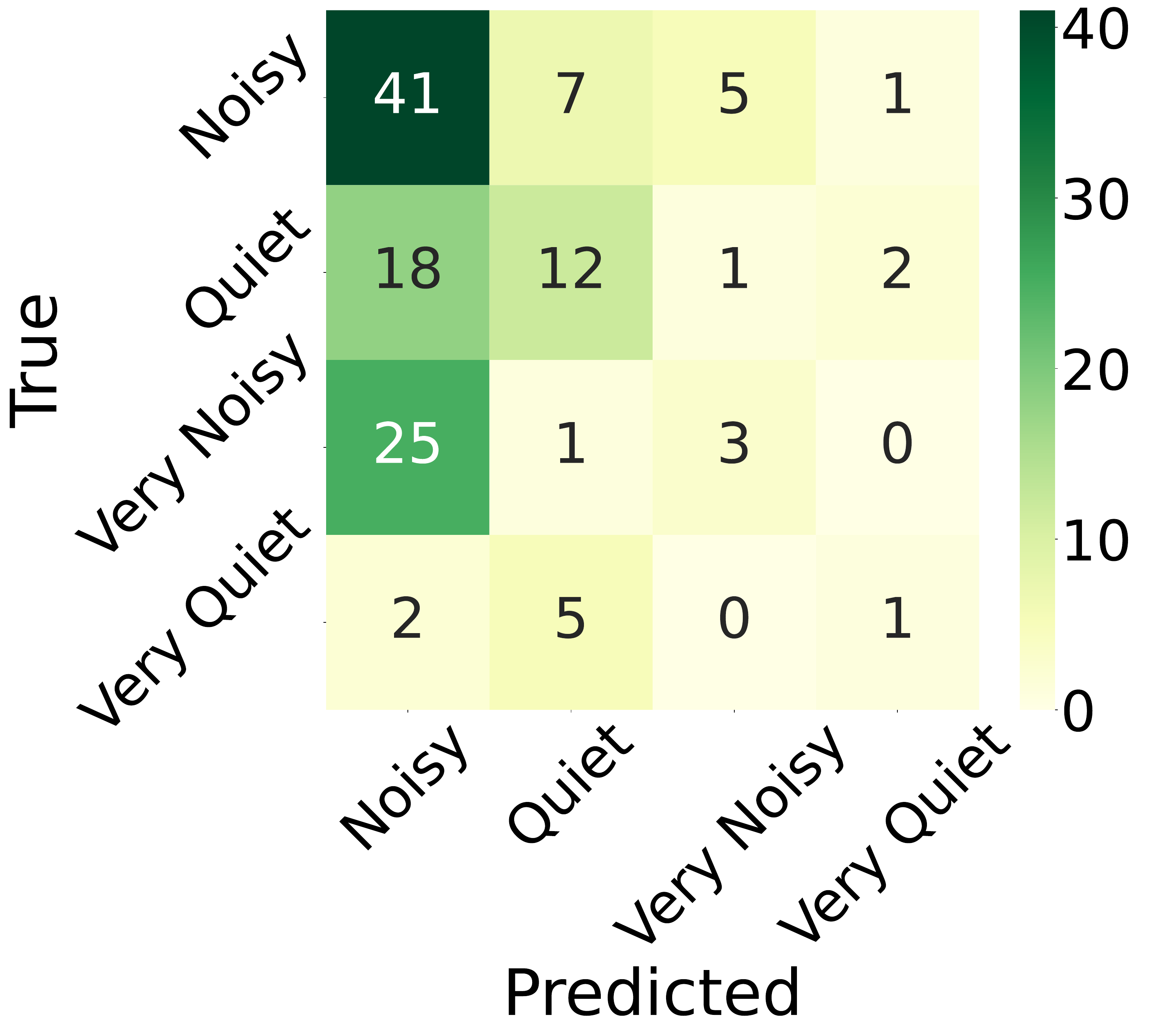}\label{fig:predicting Noise and Wealthy_4}}

  \subfloat[AlexNet]
  {\includegraphics[width=0.23\textwidth]{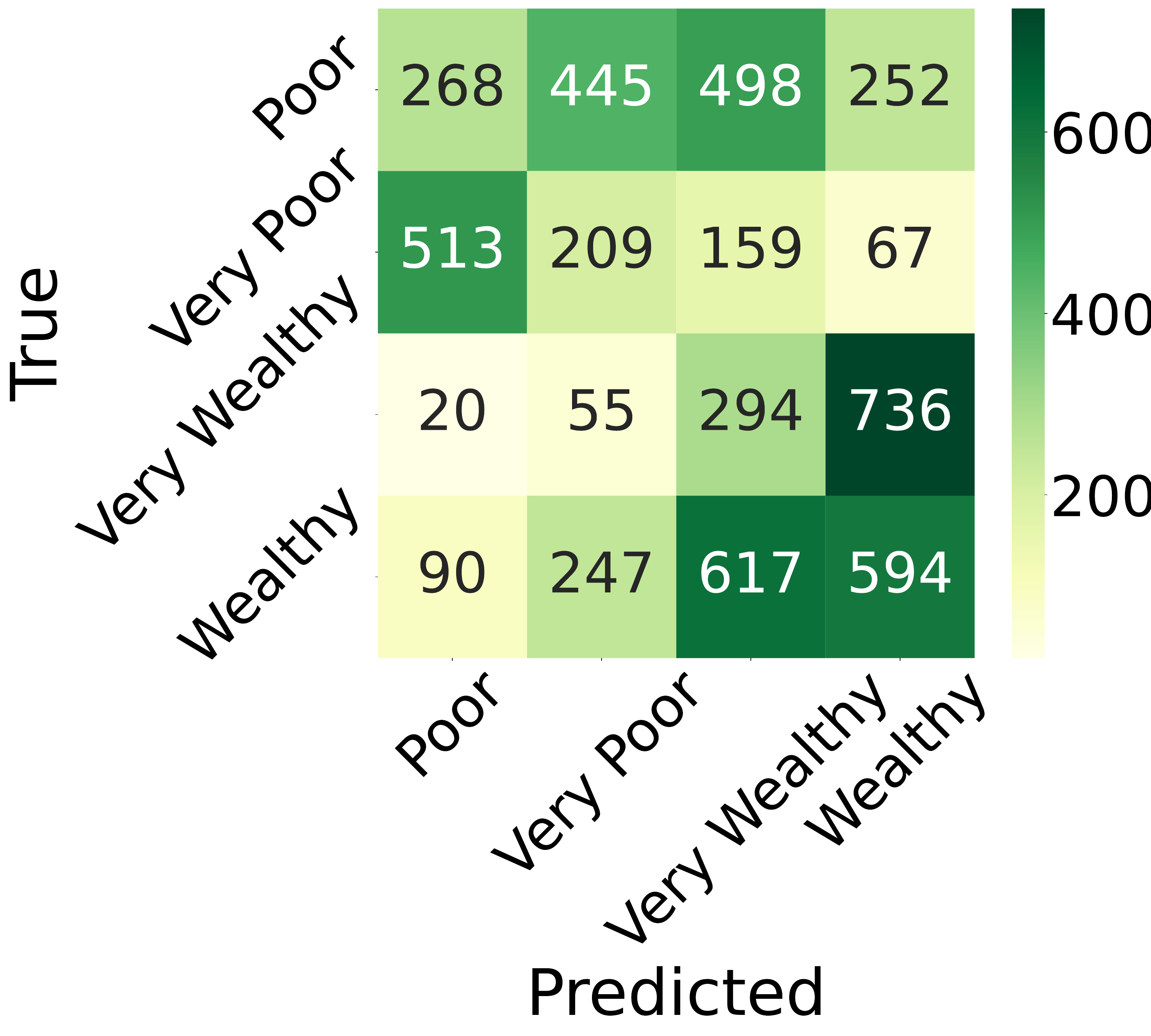}\label{fig:predicting Noise and Wealthy_5}}
  \quad
  \subfloat[GPT-4o]
  {\includegraphics[width=0.23\textwidth]{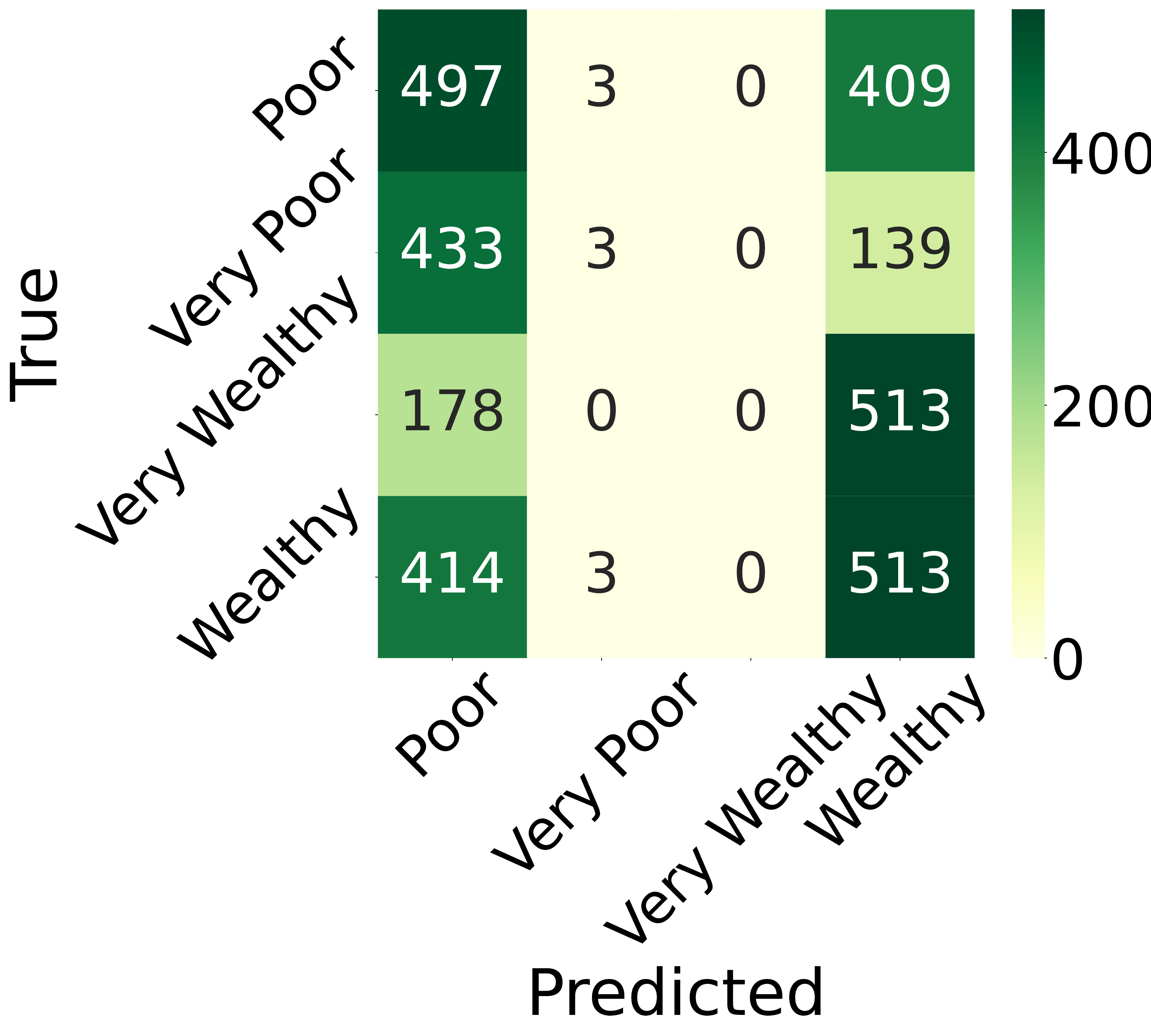}\label{fig:predicting Noise and Wealthy_6}}
  \quad
  \subfloat[InternVL2]
  {\includegraphics[width=0.23\textwidth]{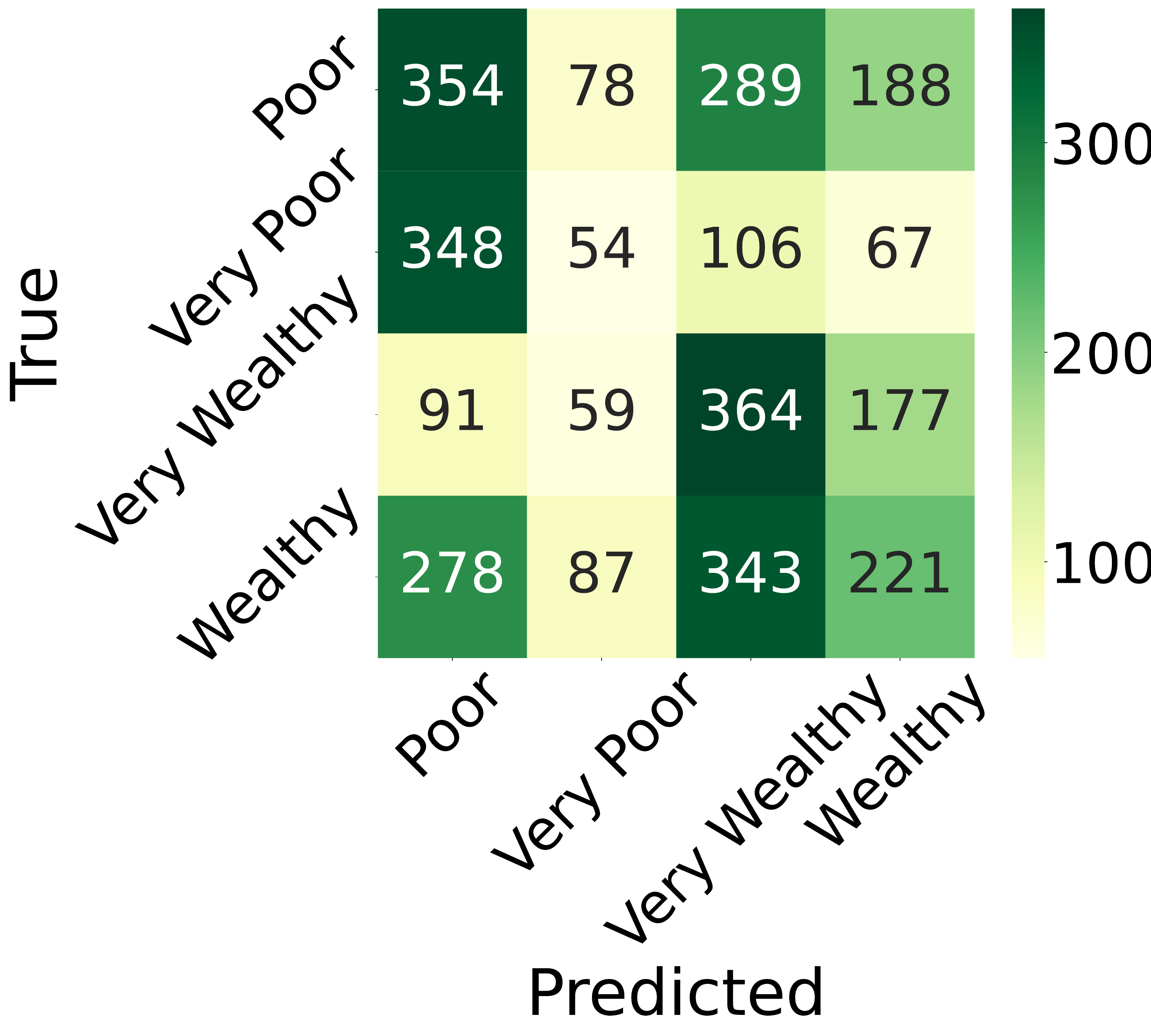}\label{fig:predicting Noise and Wealthy_7}}
  \quad
  \subfloat[OmniGeo~(LLaVA)]
  {\includegraphics[width=0.23\textwidth]{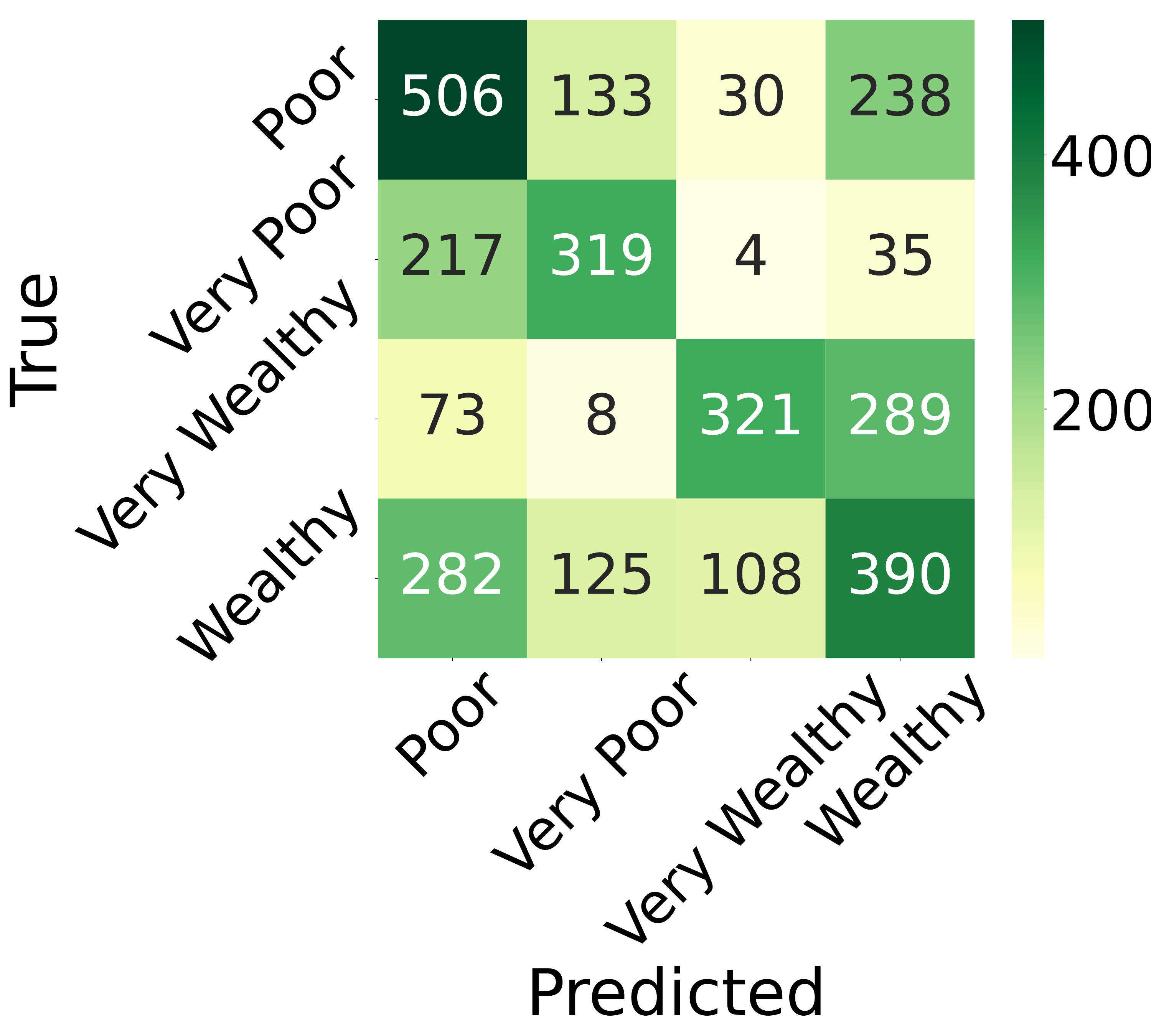}\label{fig:predicting Noise and Wealthy_8}}
  
  \caption{Confusion matrix comparison of AlexNet, InternVL2, GPT-4o and OmniGeo~(LLaVA) in predicting Noise and Wealthy indicators of urban perception}
  \label{fig:predicting Noise and Wealthy}
\end{figure*}

For the urban perception prediction, In the prediction of the Noise indicator, AlexNet tends to classify noise intensity as "Quiet" and "Very Quiet," while GPT-4o and InternVL2 exhibit similar behavior, with the only difference being that GPT-4o is more inclined to predict "Quiet," while InternVL2 tends to predict "Very Quiet". Interestingly, OmniGeo~(LLaVA) is more likely to categorize it as "Noisy." In the prediction of the Wealthy metric, the first three models still exhibit different preference patterns, while OmniGeo (LLaVA) shows relatively balanced performance.

For the geospatial semantic, we provide two cases where OmniGeo~(Qwen2) gave correct answers, but GPT-4o failed. In toponym recognition, given the text: "Romans, Sakai and the parolee died later in a shootout when the city's SWAT team stormed an apartment where the man was hiding.", GPT-4o mistakenly identifies "apartment" as a named place, which is evidently just a generic entity name. In location description recognition, given the tweet "HurricaneHarvey My mother needs help! Please send a boat to 4254 Geronimo Lake Dr. Houston, TX 77047 rescue Pleasehel", the correct answer is "4254 Geronimo Lake Dr. Houston, TX 77047", but GPT-4o erroneously splits it as "4254 Geronimo Lake Dr; Houston; TX; 77047", which is obviously an incomplete location description.

\end{document}